%% file: main.tex
\documentclass[10pt,twocolumn,letterpaper]{article}
\usepackage[accsupp]{axessibility}
 \usepackage{iccv} 

\input{preamble}

\definecolor{iccvblue}{rgb}{0.21,0.49,0.74}
\usepackage[pagebackref,breaklinks,colorlinks,allcolors=iccvblue]{hyperref}

\def\paperID{2138} 
\def\confName{ICCV}
\def\confYear{2025}
\usepackage{multirow}
\usepackage[dvipsnames]{xcolor}
\usepackage[table]{xcolor}
\usepackage{graphicx}
\DeclareGraphicsExtensions{.pdf,.png,.jpg,.jpeg}

\usepackage{listings}
\usepackage{array, booktabs, tabularx, makecell, multirow}
\usepackage{amsmath}

\definecolor{tablegreen}{RGB}{11,225,83}
\definecolor{tablered}{RGB}{230,28,100}
\definecolor{tableblue}{RGB}{60,105,225}
\definecolor{tablegray}{gray}{0.45}

\newlength\savedwidth
\newcommand\whline{\noalign{\global\savedwidth\arrayrulewidth\global\arrayrulewidth 0.8pt}
	\hline\noalign{\global\arrayrulewidth\savedwidth}}

\title{Not All Degradations Are Equal: A Targeted Feature Denoising Framework for Generalizable Image Super-Resolution
}

\author{Hongjun Wang$^{1}$,~  Jiyuan Chen$^{2}$,~ Zhengwei Yin$^{1}$, ~  Xuan Song$^{3}$\footnotemark[1],~  Yinqiang Zheng$^{1}$\thanks{Corresponding Author} \\
	$^{1}$The University of Tokyo \\
	$^{2}$The Hong Kong Polytechnic University \\
	$^{3}$Jilin University \\
	{\tt\small songxuan@jlu.edu.cn,  yqzheng@ai.u-tokyo.ac.jp}\\
}
\usepackage{pifont}
\definecolor{tablegreen}{RGB}{11,225,83}
\definecolor{tablered}{RGB}{230,28,100}
\definecolor{tableblue}{RGB}{60,105,225}
\definecolor{tablegray}{gray}{0.45}

\newtheorem{theorem}{Theorem}[section]

\newcommand{\figref}[1]{Figure~\ref{#1}} 
\usepackage{bbm}
\begin{document}
\maketitle
\begin{abstract}
	Generalizable Image Super-Resolution aims to enhance model generalization capabilities under unknown degradations. To achieve such goal, the models are expected to focus only on image content-related features instead of overfitting degradations.
	Recently, numerous approaches such as Dropout \cite{kong2022reflash} and Feature Alignment \cite{wang2024navigating} have been proposed to suppress models' natural tendency to overfitting degradations and yields promising results. Nevertheless, these works have assumed that models overfit to all degradation types (e.g., blur, noise, JPEG), while through careful investigations in this paper, we discover that models predominantly overfit to noise, largely attributable to its distinct degradation pattern compared to other degradation types. In this paper, we propose a targeted feature denoising  framework, comprising noise detection and denoising modules. Our approach presents a general solution that can be seamlessly integrated with existing super-resolution models without requiring architectural modifications.  Our framework demonstrates superior performance compared to previous regularization-based methods across five traditional benchmark and  datasets, encompassing both synthetic and real-world scenarios.	
	
\end{abstract}
\section{Introduction}
\label{sec:intro}

Image Super-Resolution (SR) has achieved remarkable progress in recent years, largely driven by the rapid evolution of deep learning techniques \cite{SRCNN, ESRGAN, VDSR, EDSR}. However, despite these advancements, deploying SR models in practical real-world applications remains highly challenging, which stems primarily from a persistent domain gap between synthetic training pipelines and the complex, diverse degradations encountered in real-world imagery.

\begin{figure}[t]
	\centering
	\includegraphics[width=1\linewidth, height=0.35\textheight]{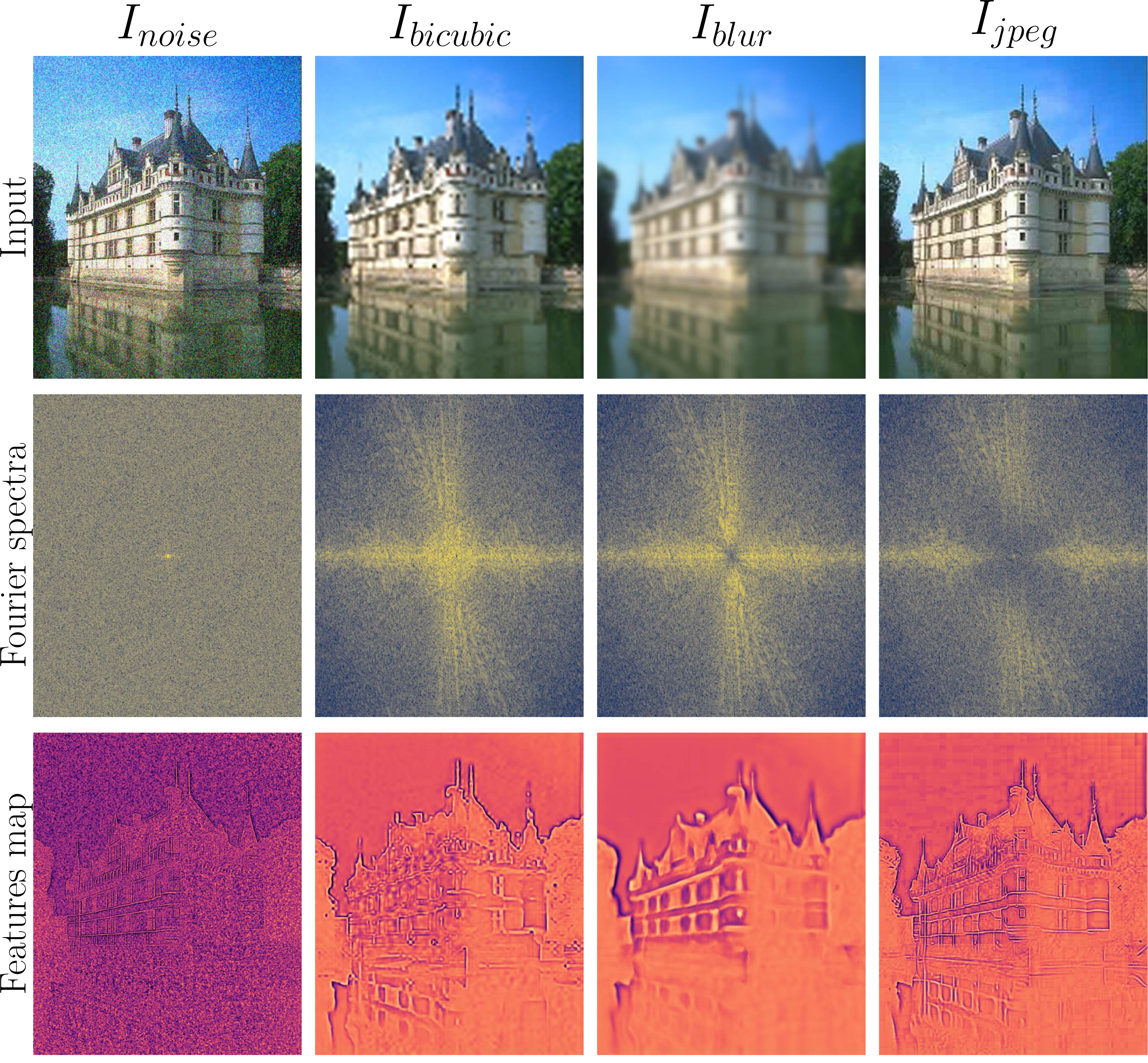}
	\caption{\textbf{Visualization of different degradation types, including input images (top), their Fourier spectra of residual images (middle), and SRResNet feature maps (bottom).} Noise (first columns) shows distinct characteristics compared to others.}
	\label{fig:motivation}
\end{figure}

Conventional SR models are typically trained on synthetic low-resolution (LR) images, which are generated using predefined degradations (e.g., bicubic downsampling) \cite{zhang2018learning, timofte2017ntire}. This simplified degradation process, while computationally convenient, is fundamentally misaligned with the intricate degradation patterns observed in real photographs. This misalignment arises from three key factors: (1) Real-world paired LR-HR data collection is non-trivial, requiring precisely aligned captures across different camera systems, lighting conditions, and imaging pipelines \cite{chen2019camera, zhang2019zoom}; (2) Real-world degradations involve complex interactions between sensor noise, optical aberrations, compression artifacts, and environmental factors, which are difficult to fully characterize using simple analytical models \cite{liu2022blind, zhang2021designing}; and (3) Existing real-world SR datasets often have limited coverage, capturing only a narrow subset of real degradations, thus lacking the diversity needed for robust generalization \cite{cai2019toward, wei2021unsupervised}.

\begin{figure}[t]
	\centering
	\subfloat[DIV2K]{\includegraphics[width=0.5\linewidth]{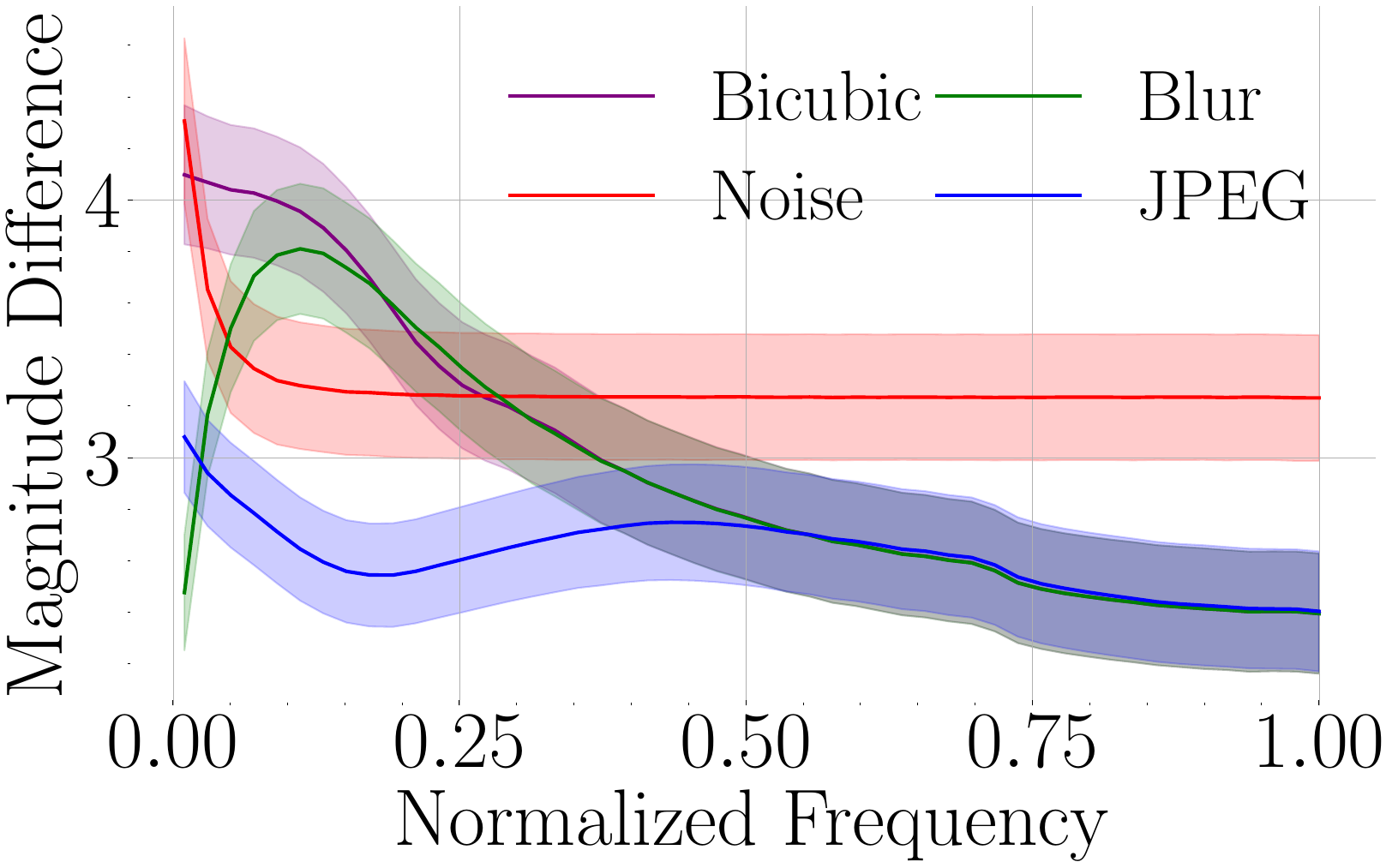}}
	\hfill
	\subfloat[BSD100]{\includegraphics[width=0.5\linewidth]{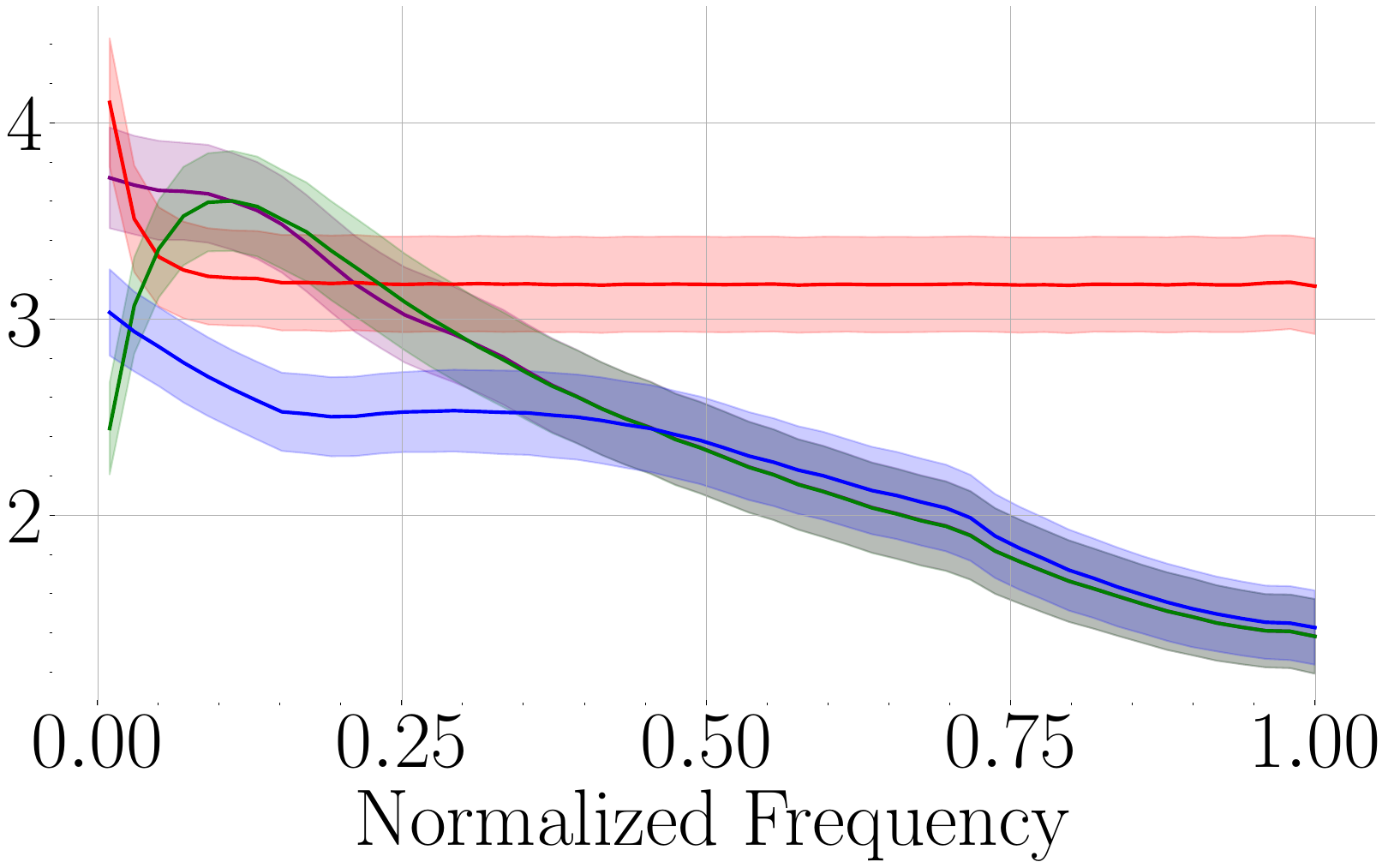}}
	\caption{\textbf{Frequency-domain analysis of image degradation showing average magnitude differences across normalized frequencies for DIV2K training set (a) and BSD100 test set (b).} The distinctive spectral characteristics of noise degradation contribute to model overfitting issues.}
	\label{fig:freq}
\end{figure}

To bridge this gap, recent works have introduced more realistic, high-order degradation modeling pipelines \cite{wang2021realesrgan,zhang2021designing,zhang2022closer}, designed to synthesize complex degradation mixtures that better mimic real-world scenarios by mixing several basic degradations, like blur, noise, JPEG. While these efforts represent significant progress, \citet{kong2022reflash} observed that even with improved degradation modeling, existing SR models still tend to overfit to specific degradation patterns, ultimately limiting their generalization potential.

In response, regularization-based strategies such as dropout \cite{kong2022reflash} and feature alignment \cite{wang2024navigating} have been proposed to mitigate degradation overfitting. However, these methods often implicitly assume uniform overfitting across all degradation types, applying generic regularization regardless of the degradation source (e.g., blur, noise, or compression artifacts). Such uniform treatment overlooks the degradation-specific characteristics that drive overfitting. In particular, through detailed analysis, we uncover a critical observation: \textit{noise degradations exhibit uniquely disruptive behaviors that make them the dominant factor contributing to overfitting}. 

To systematically analyze degradation impacts, we decompose high-order degradations \cite{wang2021realesrgan} into individual components and examine their isolated effects on model performance.  As shown in \figref{fig:motivation}, noise exhibits random, unstructured spatial patterns, unlike the localized or texture-aware degradations (e.g., blur, JPEG). This irregularity propagates into the feature space, corrupting structural consistency far more severely than other degradations. In \figref{fig:freq},  noise further amplifies mid-to-high frequency magnitudes \cite{chen2024comparative}, while other degradations follow smoother spectral decay.  But, in low-frequency regions, due to the contamination of various degradations, it more challenging to overfit to specific degradation.
This discrepancy makes noise prone to overfitting, severely impairing the model’s ability to learn reliable content features.

\begin{figure}[t]
	\centering
	\subfloat[SRResNet]{\includegraphics[width=0.24\linewidth]{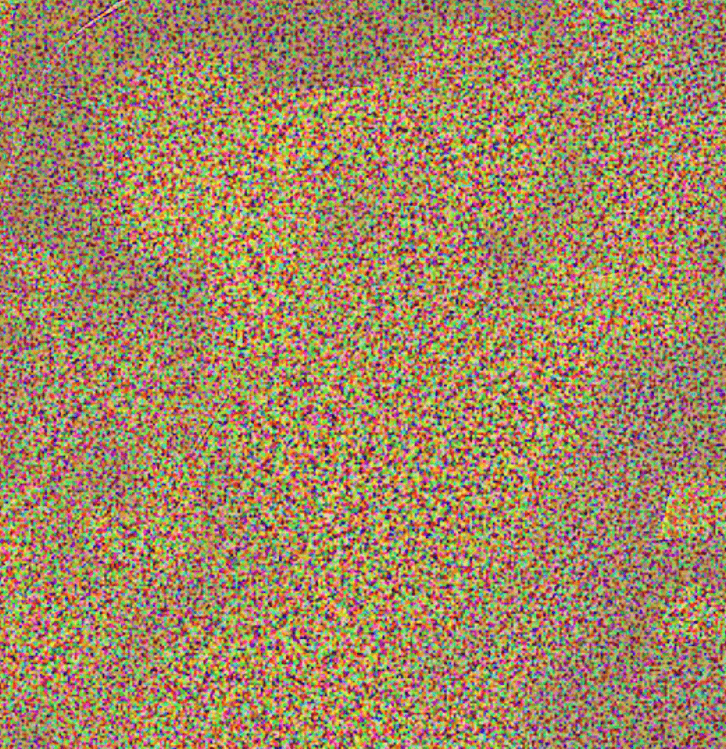}}
	\hfill
	\subfloat[+Dropout]{\includegraphics[width=0.24\linewidth]{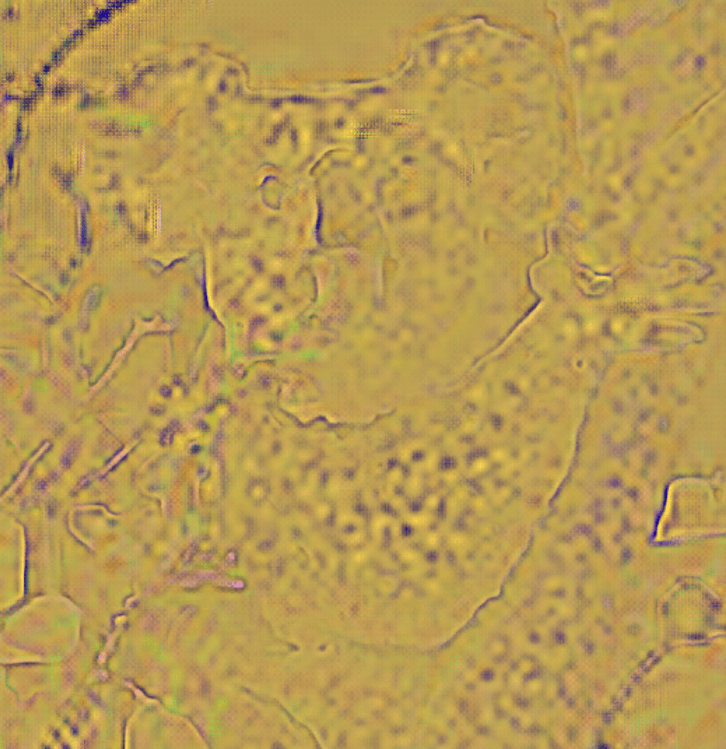}}
	\hfill
	\subfloat[+Alignment]{\includegraphics[width=0.24\linewidth]{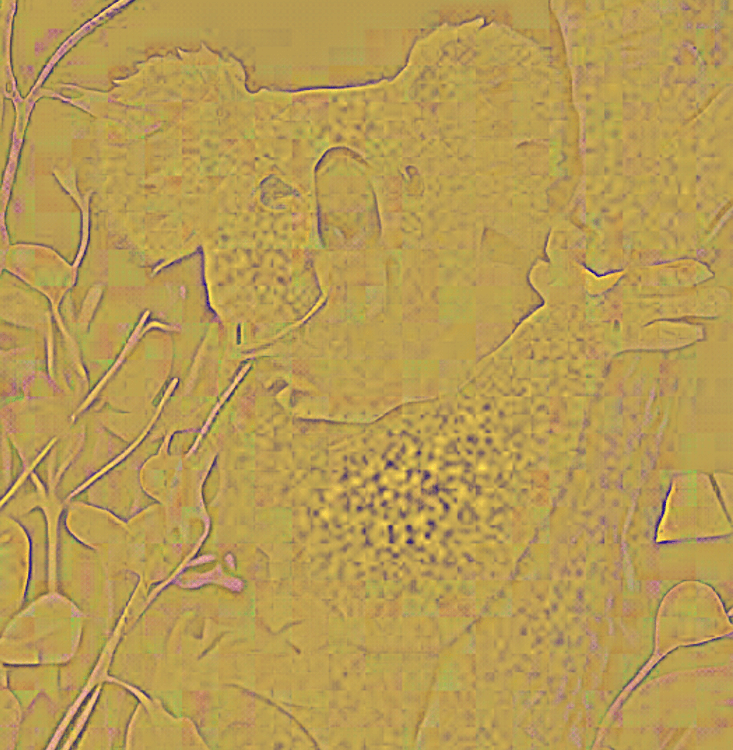}}
	\hfill
	\subfloat[+TFD (ours)]{\includegraphics[width=0.24\linewidth]{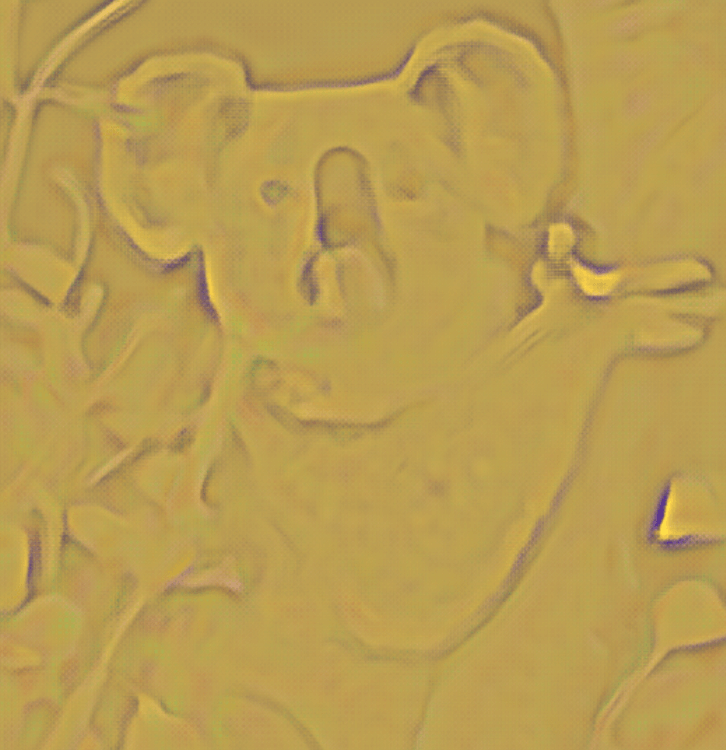}}
	\caption{\textbf{Visualization of feature representations across different methods. The baseline model (a) shows significant noise corruption.} While dropout  \cite{kong2022reflash} and feature alignment \cite{wang2024navigating} reduce noise to some extent, they both retain noticeable artifacts. Our denoising approach  effectively preserves structural details while thoroughly suppressing noise artifacts.}\label{fig:compare}
\end{figure}

Analysis of SRResNet's internal feature maps (\figref{fig:compare}) reveals that conventional regularization techniques inadequately suppress noise, resulting in persistent artifacts. While the baseline SRResNet exhibits significant noise amplification, established strategies like dropout and feature alignment offer only marginal improvement. This highlights a critical limitation: conventional regularization fails to effectively disentangle noise from content features, ultimately compromising both image fidelity and robustness.

To explicitly address this issue, we propose a  targeted feature denoising (TFD) framework, which directly tackles noise-induced overfitting in SR models. Instead of applying uniform constraints across all features, it dynamically detects noise-corrupted features and applies adaptive denoising, preserving content-related structures while suppressing noise artifacts. Our design is: \textbf{1)} \emph{noise-aware}: unlike generic regularization, TFD  on noise—the primary source of overfitting—and applies degradation-specific treatment.
\textbf{2)} \emph{model-agnostic}: feature denoising framework is a plug-and-play module compatible with diverse SR architectures \cite{SRResNet, chen2021attention}.
\textbf{3)} \emph{lightweight}: The selective denoising mechanism introduces minimal computational overhead. Comprehensive experiments on ten synthetic and real-world benchmarks demonstrate that TFD consistently enhances generalization across diverse degradation scenarios, significantly outperforming state-of-the-art regularization techniques.

\section{Related Work}

\noindent\textbf{Single Image Super-Resolution.}
Deep learning has revolutionized single image super-resolution, beginning with SRCNN's pioneering CNN application \cite{SRCNN}. Subsequent architectural innovations—including residual learning \cite{VDSR}, dense connectivity \cite{RDN}, and attention mechanisms \cite{RCAN}—have progressively improved reconstruction quality. However, these methods primarily excel on synthetic benchmarks with well-defined degradation models, limiting their real-world applicability.
Blind super-resolution has emerged to address this limitation by enhancing images without prior knowledge of degradation processes. Recent approaches pursue two complementary strategies: (1) \textit{Practical degradation modeling}, where Zhang et al. \cite{zhang2021designing} proposed comprehensive degradation pipelines, Wang et al. \cite{wang2021realesrgan} introduced Real-ESRGAN with carefully designed training data synthesis, and Liang et al. \cite{liang2022efficient} developed degradation-adaptive networks with improved computational efficiency. (2) \textit{Explicit degradation estimation}, exemplified by the IKC framework's iterative kernel refinement \cite{IKC}, Zheng et al.'s unfolded deep learning approach with domain adaptation \cite{zheng2022unfolded}, and Bell-Kligler et al.'s GAN-based kernel estimation for unknown degradations \cite{bell2019blind}.

\noindent\textbf{Generalizable Image Super-Resolution.}
Achieving robust generalization across diverse degradation conditions remains a fundamental challenge in SISR. Recent advances have explored complementary strategies to address this issue. Component-based approaches include Wei et al.'s degradation-specific subcomponent decomposition \cite{wei2020component} and Li et al.'s knowledge transfer from face-specific models \cite{li2022face}. Data-centric methods encompass Chen et al.'s human-guided ground-truth generation \cite{chen2023human} and Sahak et al.'s denoising diffusion probabilistic models for severe degradations \cite{sahak2023denoising}. 
Regularization strategies have also proven effective, with Kong et al. \cite{kong2022reflash} introducing dropout as stochastic regularization to prevent degradation-specific overfitting, and Wang et al. \cite{wang2024navigating} developing feature alignment techniques for domain-invariant representations. Recent complementary work has explored random noise injection on feature statistics, data efficiency considerations, and flexible manipulable restoration approaches \cite{yinrandom, yin2024exploring, yin2024flexir}. Our proposed method differs from existing general regularization strategies by explicitly addressing noise overfitting, which we identify as a primary bottleneck in model generalization.



\section{Analysis on Feature Representation}

We investigate noise overfitting from both empirical and theoretical perspectives, revealing its disruptive effect on content preservation. 

\begin{figure}[t]
	\centering
	\subfloat[SwinIR]{\includegraphics[width=0.49\linewidth]{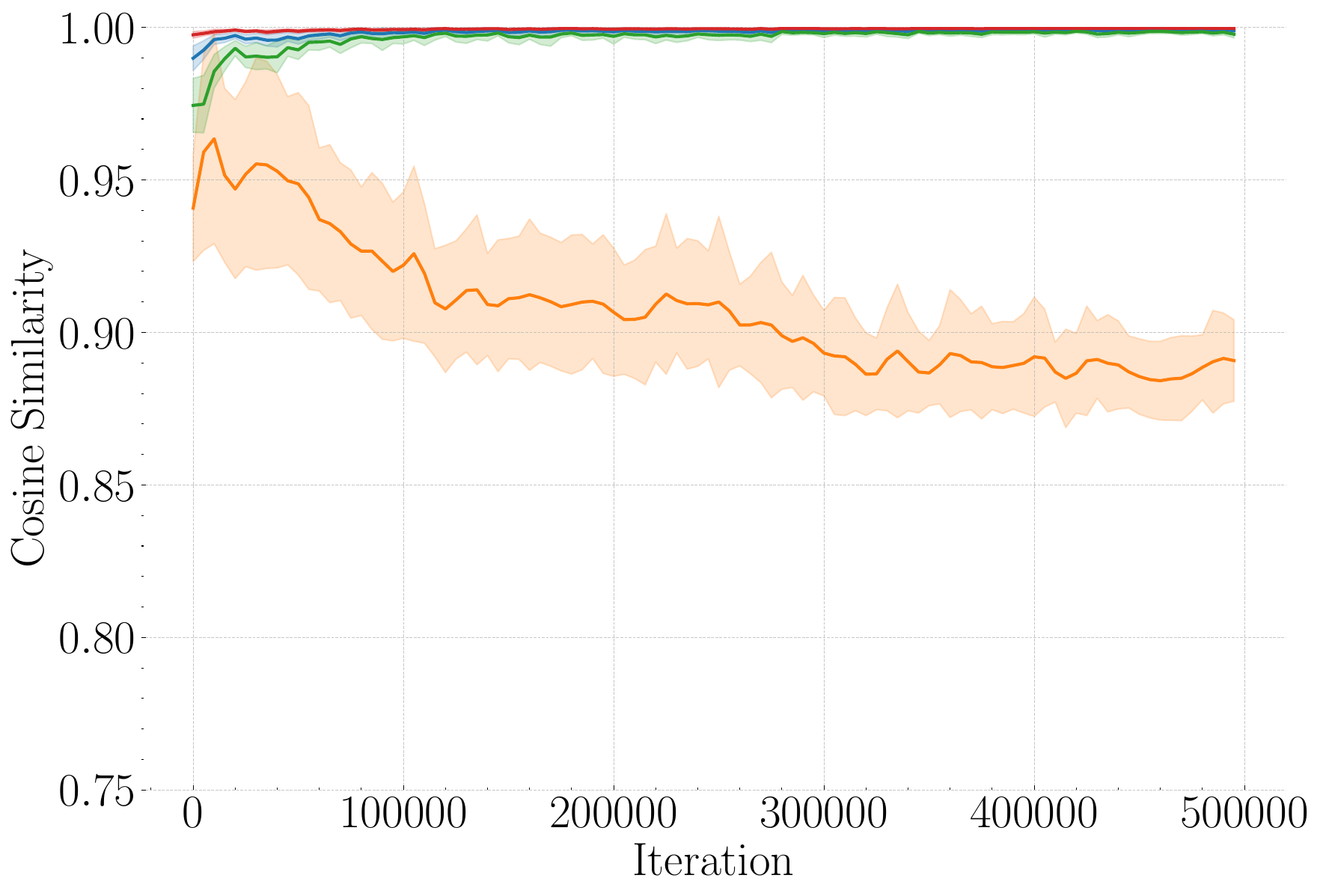}}
	\subfloat[SRResNet]{\includegraphics[width=0.475\linewidth]{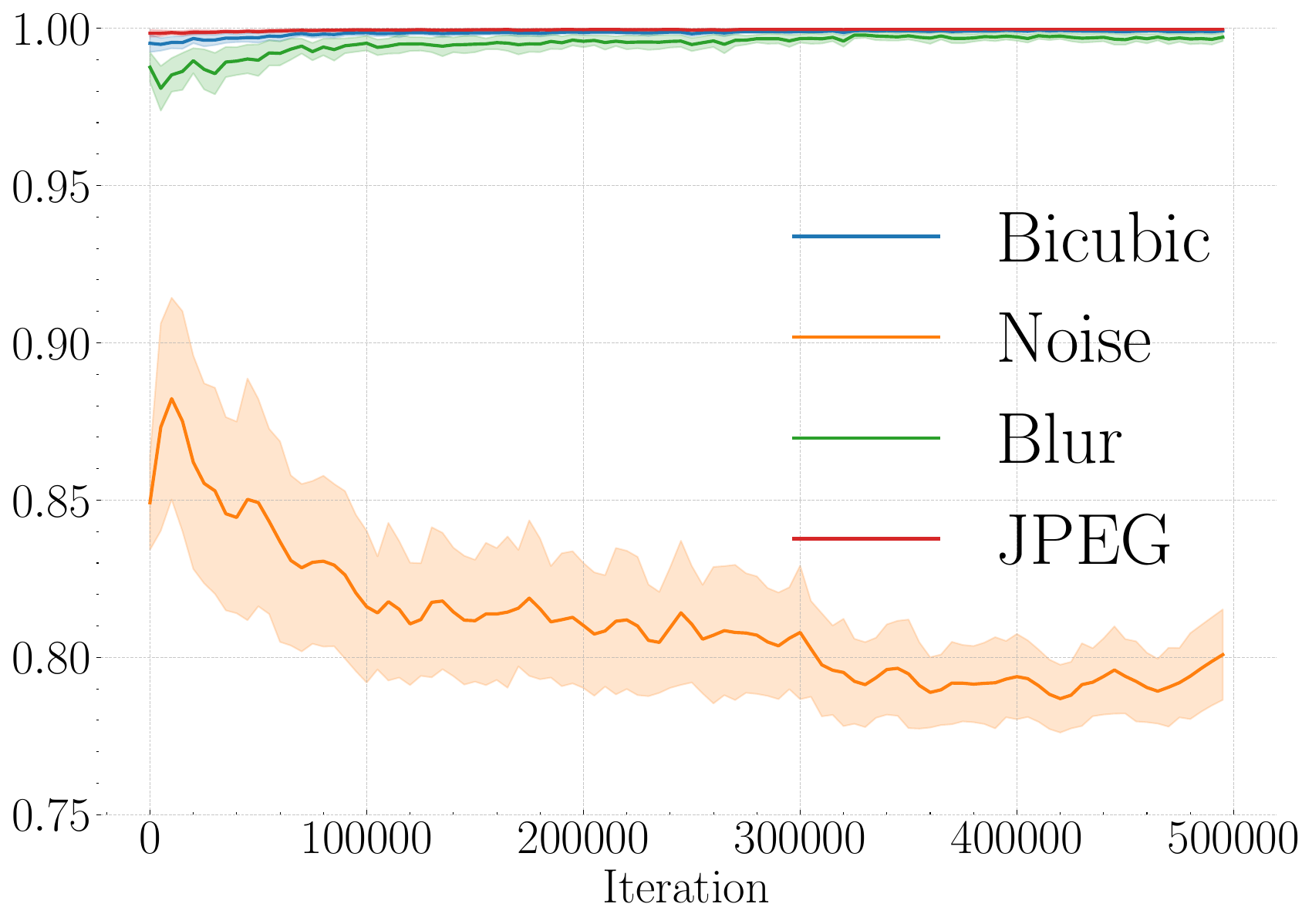}}
	\caption{\textbf{Feature similarity between degraded and clean images across training on BSD100.} Noise leads to sharper content drift than other degradations, indicating overfitting.}
	\label{fig:similar2gt}
\end{figure}

\noindent\emph{Empirical Analysis on Noise Overfitting.}
To quantify the effect of different degradations on feature retention, we analyze feature similarity throughout training. Specifically, we train SwinIR and SRResNet using RealESRGAN \cite{wang2021realesrgan} on DIV2K \cite{DIV2K} and evaluate on BSD100 \cite{BSD100} under isolated degradations. Given intermediate feature representations of ground-truth and degraded images at training step $t$, denoted as $h_{{gt}}^{(t)}$ and $h_{{deg}}^{(t)}$, we compute cosine similarity:
\begin{align}
	\text{CosSim}(h_{{gt}}^{(t)}, h_{{deg}}^{(t)}) = \frac{h_{{gt}}^t \cdot h_{{deg}}^{(t)}}{\|h_{{gt}}^{(t)}\| \cdot \|h_{{deg}}^{(t)}\|}.
\end{align}
As shown in \figref{fig:similar2gt}, noise-induced degradation leads to a sharper and more severe drop in feature similarity compared to blur and JPEG compression. This suggests that noise significantly disrupts content preservation, as noise randomness fundamentally interferes with coherent feature learning. These findings motivate our targeted noise-aware strategy to explicitly mitigate noise-related overfitting.

\noindent\emph{Theoretical Analysis on Noise Overfitting.}
To explain noise overfitting, we conduct a frequency analysis via 2D Fast Fourier Transform (FFT) denoted as $\mathcal{F}$. By leveraging the linearity property of the Fourier transform \cite{gaskill1978linear} and noise additivity \cite{lehtinen2018noise2noise}, we can decompose degraded images as: $\mathcal{F}(I_{deg}) = \mathcal{F}(I_{\text{content}}) + \mathcal{F}(I_{\text{noise}}).$
Unlike other degradations that concentrate in low frequencies, noise spans all frequencies, as shown in \figref{fig:freq}. This spectral spread amplifies overfitting, as explained by the Frequency Principle~\cite{xu2019frequency}:
\begin{theorem}[Frequency Principle]\label{thm}
	Let $I_{SR}(t)$ be the network output at training step $t$ and $I_{HR}$ the target high-resolution image. Define $S^k(t) = \mathcal{F}(I_{SR}(t))_{\mathbf{k}}$ and $H^k = \mathcal{F}(I_{HR})_{\mathbf{k}}$ as their respective Fourier coefficients at frequency $\mathbf{k}$. For the relative error $\Delta F(\mathbf{k},t) = |S^k(t) - H^k| / |H^k|$ and frequencies $\mathbf{k}_1, \mathbf{k}_2$ with $|\mathbf{k}_1| < |\mathbf{k}_2|$
	\begin{equation}
		\begin{aligned}
			\Delta F(\mathbf{k}_1,t) &< \Delta F(\mathbf{k}_2,t),\\
			\left|\frac{d}{dt}\Delta F(\mathbf{k}_1,t)\right| &> \left|\frac{d}{dt}\Delta F(\mathbf{k}_2,t)\right|.
		\end{aligned}
	\end{equation}
\end{theorem}
This theorem provides crucial insights into the behavior of neural networks during training, which shows that:
\begin{itemize}
	\item \emph{Networks learn low-frequency components first}: The error for low frequencies ($\Delta F(\mathbf{k}_1,t)$) is consistently smaller than for high frequencies ($\Delta F(\mathbf{k}_2,t)$).
	\item  \emph{Low-frequency learning occurs faster}: The rate of error reduction $|\frac{d}{dt}\Delta F(\mathbf{k}_1,t)|$ is greater for low frequencies than high frequencies.
\end{itemize}
Theorem \ref{thm} behavior is reflected in the gradient update: $\frac{d \mathcal{L}}{d\theta} \propto \sum_{\omega} w(\omega, t) \mathcal{F}(I_{deg}, \omega),$
where $w(\omega, t)$ encodes the frequency-dependent learning dynamics over training time.  As training progresses, the effective signal-to-noise ratio (SNR) \cite{box1988signal} in gradient updates monotonically decreases:
\begin{align}
	\text{SNR}(t) = \frac{\sum_{\omega} w(\omega, t)|\mathcal{F}(I_{content}, \omega)|^2}{\sum_{\omega} w(\omega, t)|\mathcal{F}(I_{noise}, \omega)|^2}.
\end{align}
Since $w(\omega, t)$ shifts emphasis from low to high frequencies with increasing $t$, and content energy diminishes at higher frequencies while noise remains uniform across the spectrum, we obtain:
$\text{SNR}(t+1) < \text{SNR}(t), \ \forall t \geq 0$.
This temporal degradation of SNR explains the increasing vulnerability to noise overfitting in later training stages.

However, unlike degradations such as blur or JPEG that selectively impair certain frequency bands, noise spans the full frequency spectrum (see \figref{fig:freq}).  Since noise is semantically meaningless and inherently random, this overfitting directly compromises content fidelity, explaining the severe drop in feature similarity observed in \figref{fig:similar2gt}.
%

This spectral entanglement between high-frequency image content and noise poses a unique challenge for conventional learning pipelines. 
It highlights the necessity for noise-aware training strategies that explicitly disentangle meaningful content from noise across all frequency bands.

\section{Targeted Feature Denoising Framework}  
\begin{figure}[t]
	\centering
	\includegraphics[width=1\linewidth]{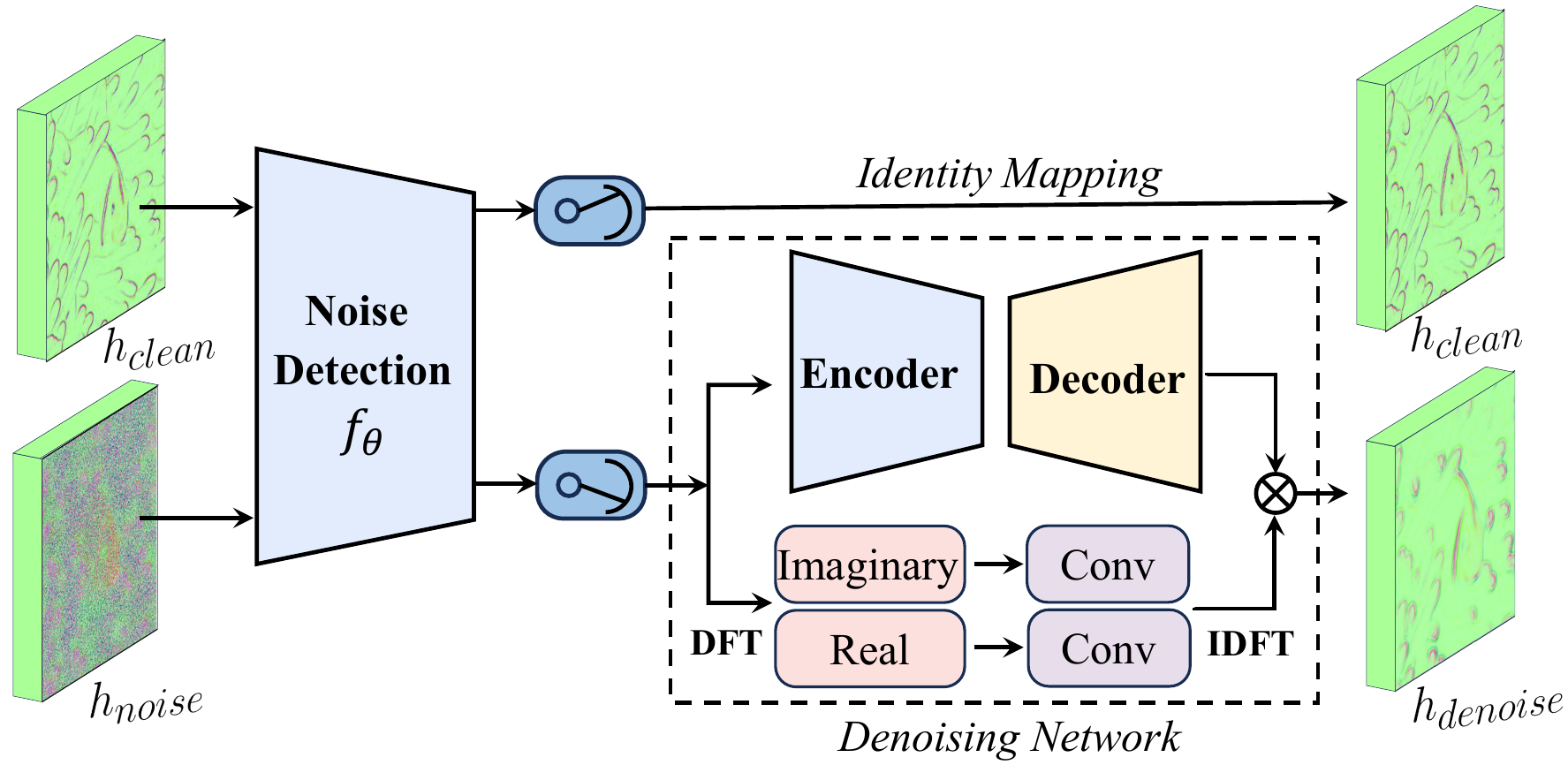}
	\caption{\textbf{Overview of the proposed denoising framework.} The noise estimation module ($f_\theta$) first decides whether denoising is required. If not, the features are directly passed through an identity shortcut. Otherwise, the features are refined by a denoising network (dashed box), which integrates spatial processing (encoder-decoder) and frequency processing (DFT-IDFT). $\otimes$ denotes feature fusion.}
	\label{fig:framework}
\end{figure}

In this section, we introduce our novel targeted feature denoising framework, designed to explicitly identify and suppress noise-contaminated features while preserving semantically-relevant content representations. The overall pipeline, illustrated in \figref{fig:framework}, consists of two essential components: 1) a noise detection module, which predicts the likelihood of noise corruption at the feature level, and 2) a {frequency-spatial denoising module}, which performs selective feature refinement by jointly leveraging complementary frequency and spatial clues.

\noindent\textbf{Noise Detection Module.}
Our noise detection module builds upon the key observation from \citet{dosselmann2012no} that \textit{noise contamination significantly amplifies high-frequency spectral components, caused by abrupt pixel intensity variations.} To leverage this property, we design a lightweight and adaptable noise detection module $f_\theta$ that explicitly captures these spectral signatures. This module is designed to be \emph{plug-and-play}, seamlessly integrating with spatial-domain backbones, while maintaining strong generalizability to unseen noise types due to its spectral grounding.
Given an intermediate feature map $h \in \mathbb{R}^{C \times H \times W}$ extracted from the backbone encoder, we first transform $h$ into the frequency domain via the Fourier transform:
$\mathcal{F}(h) = F_r + iF_i$
where $F_r$ and $F_i$ represent the real and imaginary components, respectively. To emphasize noise-specific spectral responses, we apply two independent learnable filters $\mathcal{W}_r$ and $\mathcal{W}_i$ to these frequency components:
\begin{equation}
	F'_r = \phi(\mathcal{W}_r \circledast F_r), \quad F'_i = \phi(\mathcal{W}_i \circledast F_i)
\end{equation}
where $\phi$ denotes the ReLU activation and $\circledast$ represents convolution, which adaptively amplifies noise-specific spectral signatures while preserving essential structural information.

The frequency domain contains both amplitude and phase information, while direct classification in the frequency domain may focus too much on amplitude and ignore phase. Through Inverse FFT $\mathcal{F}^{-1}$, we allow these two types of information to be reintegrated to provide a more comprehensive feature representation.
The  feature map is then processed by Inverse FFT, global average pooling $\Psi$, flattened into a feature vector, and passed through a lightweight convolutional classifier to predict whether the feature is clean or noisy:
\[
y = \text{ConvClassifier}(\text{Flatten}(\Psi(\mathcal{F}^{-1}(F'_r + iF'_i)))) \in \mathbb{R}^2,
\]
where $y$ denotes binary logits indicating the noise state.



\noindent\textbf{Frequency-Spatial Denoising Module.}  
To effectively exploit the complementary strengths of frequency-domain analysis and spatial-domain processing, we propose a dual-path denoising architecture. As illustrated in \figref{fig:motivation}, this design is motivated by the observation that noise exhibits distinct characteristics from other degradations across frequency and spatial domains. By jointly leveraging these two perspectives, our module achieves enhanced noise suppression while preserving semantic content.

\noindent\emph{Frequency Domain Branch.}   To improve computational efficiency, this branch reuses the Fourier transform results $\mathcal{F}(h_{n})$ already obtained from the noise detection module $f_{\theta}$. Given these pre-computed frequency components, we introduce two specialized learnable filters $\mathcal{W}_{r}^{n}$ and $\mathcal{W}_{i}^{n}$, which are applied independently to the real and imaginary parts of the spectral representation. The denoised frequency representation is then transformed back to the spatial domain, generating a noise attention mask:
\begin{equation}
	h_{\text{freq}} = \sigma \left( \mathcal{F}^{-1} \left( \mathcal{W}_r^n \circledast \mathcal{F}_r + \mathcal{W}_i^n \circledast \mathcal{F}_i \right) \right),
\end{equation}
where $\circledast$ denotes Hadamard product, and $\sigma$ is the sigmoid function. The result $h_{\text{freq}} \in [0,1]^{C \times H \times W}$ encodes per-pixel noise likelihood, guiding spatial refinement.

\noindent\emph{2) Spatial Domain Branch.}  
This branch performs hierarchical reconstruction via an encoder-bottleneck-decoder pipeline: 
\begin{equation}
	h_{\text{up}} = \mathcal{D}(\Gamma(\mathcal{E}(h_n))),
\end{equation}
where $\mathcal{E}$ and $\mathcal{D}$ denote the encoder and decoder, and $\Gamma$ is a non-linear bottleneck operator. Each block contains a residual attention unit defined as:
\begin{equation}
	h_{\text{res}} = \mathcal{W}_{1\times1} \left( \text{CA} \left( \mathcal{W}_{3\times3}^{\text{dw}} \left( \mathcal{W}_{1\times1} (\text{LN}(h_n)) \right) \right) \right) + h_n,
\end{equation}
where LN is layer normalization, $\mathcal{W}_{3\times3}^{\text{dw}}$ is a depthwise convolution, and $\text{CA}$ is a channel attention module. Skip connections between $\mathcal{E}$ and $\mathcal{D}$ enable signal preservation. The output is fused with input features via:
\begin{equation}
	h_{\text{spatial}} = \Lambda([h_{\text{up}}, h_n]),
\end{equation}
where $\Lambda$ is a lightweight $1\times1$ convolution that consolidates scale-aware residuals.

\noindent\emph{3) Cross-Domain Fusion.}  
We apply the frequency-guided mask to spatial features:
\begin{equation}
	h_{\text{denoised}} = h_{\text{freq}} \odot h_{\text{spatial}},
\end{equation}
where $\odot$ denotes element-wise multiplication. This operation adaptively scales each channel and pixel according to its noise probability, enhancing restoration accuracy in contaminated regions.

\vspace{0.8em}
\noindent\textbf{Training Strategy.}  
We use a tri-objective loss to balance fidelity, noise detection, and feature consistency.

\noindent\emph{1) Reconstruction Loss.}  
The main objective supervises pixel-wise recovery:

\begin{equation}
	\mathcal{L}_{\text{rec}} = \|I_{\text{SR}} - I_{\text{HR}}\|_1,
\end{equation}
where $I_{\text{SR}}$ is the super-resolved output and $I_{\text{HR}}$ is the clean reference.

\noindent\emph{2) Noise Classification Loss.}  
The noise classifier $f_\theta$ is supervised using cross-entropy:

\begin{equation}
	\mathcal{L}_{\text{cls}} = -\frac{1}{N}\sum_{i=1}^N \sum_{c=0}^1 y_{i,c} \log(\hat{y}_{i,c}),
\end{equation}
with $y_{i,c} \in \{0,1\}$ the ground-truth label and $\hat{y}_{i,c}$ the predicted softmax probability.

\noindent\emph{3) Feature Consistency Loss.}  
To preserve high-level semantics, we enforce that denoised features remain close to clean references:
\begin{equation}
	\mathcal{L}_{\text{feat}} = \|h_{\text{denoised}} - h_{\text{ref}}\|_1,
\end{equation}
where $h_{\text{ref}}$ is extracted from the noise-free image via the same encoder.

\noindent\emph{Total Loss.}  
The final objective is:
\begin{equation}
	\mathcal{L} = \mathcal{L}_{\text{rec}} + \lambda_{\text{cls}} \mathcal{L}_{\text{cls}} + \lambda_{\text{feat}} \mathcal{L}_{\text{feat}}.
\end{equation}
To prevent early overfitting to uncertain noise estimates, we activate denoising only when the predicted noise confidence surpasses a 0.75 threshold. This dynamic scheduling allows structure-preserving training in early stages, with aggressive suppression deferred to later epochs.

\begin{table*}[t]
	\centering
	\caption{\textbf{Average PSNR and SSIM of different methods in $\times$4 blind SR on five benchmarks with eight types of degradations.}}
	\label{table:degradations1}
	\renewcommand{\arraystretch}{1.05}
	\resizebox{1\linewidth}{!}{
		\begin{tabular}{cccccccccccccccccccccc}
			\hline
			\multirow{2}{*}{Data} & \multirow{2}{*}{Method} & \multicolumn{2}{c}{Clean} & \multicolumn{2}{c}{Blur} & \multicolumn{2}{c}{Noise} & \multicolumn{2}{c}{JPEG} & \multicolumn{2}{c}{Blur+Noise} & \multicolumn{2}{c}{Blur+JPEG} & \multicolumn{2}{c}{Noise+JPEG} & \multicolumn{2}{c|}{Blur+Noise+JPEG} & \multicolumn{2}{c}{Average} \\ \cline{3-20}
			&  & \multicolumn{2}{c}{PSNR \ SSIM} & \multicolumn{2}{c}{PSNR \ SSIM} & \multicolumn{2}{c}{PSNR \ SSIM} & \multicolumn{2}{c}{PSNR \ SSIM} & \multicolumn{2}{c}{PSNR \ SSIM} & \multicolumn{2}{c}{PSNR \ SSIM} & \multicolumn{2}{c}{PSNR \ SSIM} & \multicolumn{2}{c|}{PSNR \ SSIM} & \multicolumn{2}{c}{PSNR SSIM}  \\ \toprule
			\multirow{8}{*}{\rotatebox{90}{Set5}}
			& SRResNet \cite{SRResNet} & \multicolumn{2}{c}{24.85 \ 0.7295} & \multicolumn{2}{c}{24.73 \ 0.7113} & \multicolumn{2}{c}{23.69 \ 0.6422} & \multicolumn{2}{c}{23.69 \ 0.6714} & \multicolumn{2}{c}{23.25 \ 0.6142} & \multicolumn{2}{c}{23.41 \ 0.6534} & \multicolumn{2}{c}{23.10 \ 0.6508} & \multicolumn{2}{c|}{22.68 \ 0.6249} & \multicolumn{2}{c}{23.68 \ 0.6622} \\
			& +TFD & \multicolumn{2}{c}{ \bf 26.71 \ 0.7295} & \multicolumn{2}{c}{ \bf 26.20 \ 0.7113} & \multicolumn{2}{c}{ \bf 24.31 \ 0.6422} & \multicolumn{2}{c}{ \bf 24.49 \ 0.6714} & \multicolumn{2}{c}{ \bf 23.39 \ 0.6142} & \multicolumn{2}{c}{ \bf 23.92 \ 0.6534} & \multicolumn{2}{c}{ \bf 23.67 \ 0.6508} & \multicolumn{2}{c|}{\bf 22.99 \ 0.6249} & \multicolumn{2}{c}{ \bf 24.46 \ 0.6622} \\ \cline{2-20}
			& RRDB \cite{ESRGAN} & \multicolumn{2}{c}{25.18 \ 0.7344} & \multicolumn{2}{c}{25.12 \ 0.7176} & \multicolumn{2}{c}{22.92 \ 0.6317} & \multicolumn{2}{c}{23.82 \ 0.6739} & \multicolumn{2}{c}{23.44 \ 0.6183} & \multicolumn{2}{c}{23.45 \ 0.6542} & \multicolumn{2}{c}{23.32 \ 0.6548} & \multicolumn{2}{c|}{22.81 \ 0.6279} & \multicolumn{2}{c}{23.76 \ 0.6641} \\
			& +TFD & \multicolumn{2}{c}{ \bf 26.83 \ 0.7773} & \multicolumn{2}{c}{ \bf 26.59 \ 0.7632} & \multicolumn{2}{c}{ \bf 24.96 \ 0.7068} & \multicolumn{2}{c}{ \bf 24.56 \ 0.6974} & \multicolumn{2}{c}{ \bf 24.07 \ 0.6681} & \multicolumn{2}{c}{ \bf 24.04 \ 0.6741} & \multicolumn{2}{c}{ \bf 23.81 \ 0.6753} & \multicolumn{2}{c|}{\bf 23.14 \ 0.6432} & \multicolumn{2}{c}{ \bf 24.75 \ 0.7700} \\ \cline{2-20}
			& HAT  \cite{chen2023activating} & \multicolumn{2}{c}{26.40 \ 0.7608} & \multicolumn{2}{c}{25.84 \ 0.7367} & \multicolumn{2}{c}{24.86 \ 0.7071} & \multicolumn{2}{c}{24.54 \ 0.6843} & \multicolumn{2}{c}{23.93 \ 0.6676} & \multicolumn{2}{c}{23.95 \ 0.6590} & \multicolumn{2}{c}{23.84 \ 0.6728} & \multicolumn{2}{c|}{23.15 \ 0.6417} & \multicolumn{2}{c}{24.56 \ 0.6913} \\
			& +TFD & \multicolumn{2}{c}{ \bf 26.84 \ 0.7923} & \multicolumn{2}{c}{ \bf 26.38 \ 0.7753} & \multicolumn{2}{c}{ \bf 24.98 \ 0.7040} & \multicolumn{2}{c}{ \bf 24.64 \ 0.7034} & \multicolumn{2}{c}{ \bf 24.03 \ 0.6680} & \multicolumn{2}{c}{ \bf 24.05 \ 0.6820} & \multicolumn{2}{c}{ \bf 23.94 \ 0.6728} & \multicolumn{2}{c|}{\bf 23.25 \ 0.6417} & \multicolumn{2}{c}{ \bf 24.76 \ 0.7049} \\ \cline{2-20}
			& SwinIR \cite{liang2021swinir} & \multicolumn{2}{c}{26.25 \ 0.7498} & \multicolumn{2}{c}{26.03 \ 0.7317} & \multicolumn{2}{c}{24.15 \ 0.6484} & \multicolumn{2}{c}{24.37 \ 0.6816} & \multicolumn{2}{c}{23.80 \ 0.6249} & \multicolumn{2}{c}{23.84 \ 0.6601} & \multicolumn{2}{c}{23.67 \ 0.6609} & \multicolumn{2}{c|}{22.99 \ 0.6315} & \multicolumn{2}{c}{24.39 \ 0.6736} \\
			& +TFD & \multicolumn{2}{c}{ \bf 26.92 \ 0.7989} & \multicolumn{2}{c}{ \bf 26.43 \ 0.7863} & \multicolumn{2}{c}{ \bf 24.76 \ 0.7037} & \multicolumn{2}{c}{ \bf 24.56 \ 0.7095} & \multicolumn{2}{c}{ \bf 23.90 \ 0.6712} & \multicolumn{2}{c}{ \bf 24.03 \ 0.6894} & \multicolumn{2}{c}{ \bf 23.77 \ 0.6744} & \multicolumn{2}{c|}{\bf 23.09 \ 0.6443} & \multicolumn{2}{c}{ \bf 24.68 \ 0.7097} \\
			\midrule
			
			\multirow{8}{*}{\rotatebox{90}{Set14}}
			& SRResNet \cite{SRResNet} & \multicolumn{2}{c}{23.25 \ 0.6286} & \multicolumn{2}{c}{23.05 \ 0.6085} & \multicolumn{2}{c}{22.50 \ 0.5563} & \multicolumn{2}{c}{22.36 \ 0.5817} & \multicolumn{2}{c}{22.23 \ 0.5304} & \multicolumn{2}{c}{22.10 \ 0.5637} & \multicolumn{2}{c}{22.06 \ 0.5637} & \multicolumn{2}{c|}{21.77 \ 0.5441} & \multicolumn{2}{c}{22.41 \ 0.5721} \\
			& +TFD & \multicolumn{2}{c}{ \bf 24.54 \ 0.6286} & \multicolumn{2}{c}{ \bf 24.15 \ 0.6085} & \multicolumn{2}{c}{ \bf 23.02 \ 0.5563} & \multicolumn{2}{c}{ \bf 23.11 \ 0.5817} & \multicolumn{2}{c}{ \bf 22.43 \ 0.5304} & \multicolumn{2}{c}{ \bf 22.79 \ 0.5637} & \multicolumn{2}{c}{ \bf 22.60 \ 0.5637} & \multicolumn{2}{c|}{\bf 22.13 \ 0.5441} & \multicolumn{2}{c}{ \bf 23.10 \ 0.5721} \\ \cline{2-20}
			& RRDB \cite{ESRGAN} & \multicolumn{2}{c}{23.74 \ 0.6352} & \multicolumn{2}{c}{23.36 \ 0.6129} & \multicolumn{2}{c}{22.33 \ 0.5542} & \multicolumn{2}{c}{22.59 \ 0.5853} & \multicolumn{2}{c}{22.47 \ 0.5341} & \multicolumn{2}{c}{22.17 \ 0.5648} & \multicolumn{2}{c}{22.29 \ 0.5675} & \multicolumn{2}{c|}{21.95 \ 0.5467} & \multicolumn{2}{c}{22.61 \ 0.5751} \\
			& +TFD & \multicolumn{2}{c}{ \bf 24.72 \ 0.6625} & \multicolumn{2}{c}{ \bf 24.37 \ 0.6436} & \multicolumn{2}{c}{ \bf 23.49 \ 0.6044} & \multicolumn{2}{c}{ \bf 23.27 \ 0.5974} & \multicolumn{2}{c}{ \bf 22.85 \ 0.5720} & \multicolumn{2}{c}{ \bf 22.89 \ 0.5765} & \multicolumn{2}{c}{ \bf 22.75 \ 0.5756} & \multicolumn{2}{c|}{\bf 22.25 \ 0.5506} & \multicolumn{2}{c}{ \bf 23.32 \ 0.5978} \\ \cline{2-20}
			& HAT  \cite{chen2023activating} & \multicolumn{2}{c}{24.30 \ 0.6390} & \multicolumn{2}{c}{23.85 \ 0.6145} & \multicolumn{2}{c}{23.33 \ 0.5959} & \multicolumn{2}{c}{23.15 \ 0.5885} & \multicolumn{2}{c}{22.71 \ 0.5640} & \multicolumn{2}{c}{22.70 \ 0.5662} & \multicolumn{2}{c}{22.67 \ 0.5697} & \multicolumn{2}{c|}{22.17 \ 0.5450} & \multicolumn{2}{c}{23.11 \ 0.5854} \\
			& +TFD & \multicolumn{2}{c}{ \bf 24.83 \ 0.6777} & \multicolumn{2}{c}{ \bf 24.44 \ 0.6558} & \multicolumn{2}{c}{ \bf 23.60 \ 0.6102} & \multicolumn{2}{c}{ \bf 23.30 \ 0.6014} & \multicolumn{2}{c}{ \bf 22.83 \ 0.5766} & \multicolumn{2}{c}{ \bf 22.84 \ 0.5787} & \multicolumn{2}{c}{ \bf 22.75 \ 0.5789} & \multicolumn{2}{c|}{\bf 22.23 \ 0.5529} & \multicolumn{2}{c}{ \bf 23.35 \ 0.6040} \\ \cline{2-20}
			& SwinIR \cite{liang2021swinir} & \multicolumn{2}{c}{24.53 \ 0.6453} & \multicolumn{2}{c}{24.25 \ 0.6241} & \multicolumn{2}{c}{23.46 \ 0.5680} & \multicolumn{2}{c}{23.14 \ 0.5935} & \multicolumn{2}{c}{22.53 \ 0.5350} & \multicolumn{2}{c}{22.73 \ 0.5738} & \multicolumn{2}{c}{22.59 \ 0.5732} & \multicolumn{2}{c|}{22.20 \ 0.5507} & \multicolumn{2}{c}{23.18 \ 0.5830} \\
			& +TFD & \multicolumn{2}{c}{ \bf 24.96 \ 0.6936} & \multicolumn{2}{c}{ \bf 24.60 \ 0.6735} & \multicolumn{2}{c}{ \bf 23.56 \ 0.6044} & \multicolumn{2}{c}{ \bf 23.23 \ 0.6063} & \multicolumn{2}{c}{ \bf 22.70 \ 0.5738} & \multicolumn{2}{c}{ \bf 22.83 \ 0.5833} & \multicolumn{2}{c}{ \bf 22.69 \ 0.5775} & \multicolumn{2}{c|}{\bf 22.30 \ 0.5533} & \multicolumn{2}{c}{ \bf 23.36 \ 0.6082} \\ 
			\midrule
			\multirow{8}{*}{\rotatebox{90}{BSD100}}
			& SRResNet \cite{SRResNet} & \multicolumn{2}{c}{23.06 \ 0.5943} & \multicolumn{2}{c}{22.99 \ 0.5755} & \multicolumn{2}{c}{22.45 \ 0.5160} & \multicolumn{2}{c}{22.48 \ 0.5526} & \multicolumn{2}{c}{22.26 \ 0.4905} & \multicolumn{2}{c}{22.34 \ 0.5365} & \multicolumn{2}{c}{22.22 \ 0.5343} & \multicolumn{2}{c|}{22.05 \ 0.5158} & \multicolumn{2}{c}{22.48 \ 0.5394} \\
			& +TFD & \multicolumn{2}{c}{ \bf 23.87 \ 0.5943} & \multicolumn{2}{c}{ \bf 23.71 \ 0.5755} & \multicolumn{2}{c}{ \bf 22.67 \ 0.5160} & \multicolumn{2}{c}{ \bf 22.96 \ 0.5526} & \multicolumn{2}{c}{ \bf 22.36 \ 0.4905} & \multicolumn{2}{c}{ \bf 22.78 \ 0.5365} & \multicolumn{2}{c}{ \bf 22.52 \ 0.5343} & \multicolumn{2}{c|}{\bf 22.29 \ 0.5158} & \multicolumn{2}{c}{ \bf 22.89 \ 0.5394} \\ \cline{2-20}
			& RRDB \cite{ESRGAN} & \multicolumn{2}{c}{23.38 \ 0.5994} & \multicolumn{2}{c}{23.32 \ 0.5803} & \multicolumn{2}{c}{22.09 \ 0.5119} & \multicolumn{2}{c}{22.73 \ 0.5563} & \multicolumn{2}{c}{22.39 \ 0.4926} & \multicolumn{2}{c}{22.47 \ 0.5382} & \multicolumn{2}{c}{22.42 \ 0.5371} & \multicolumn{2}{c|}{22.15 \ 0.5175} & \multicolumn{2}{c}{22.62 \ 0.5417} \\
			& +TFD & \multicolumn{2}{c}{ \bf 24.11 \ 0.6191} & \multicolumn{2}{c}{ \bf 24.05 \ 0.6058} & \multicolumn{2}{c}{ \bf 23.13 \ 0.5628} & \multicolumn{2}{c}{ \bf 23.19 \ 0.5637} & \multicolumn{2}{c}{ \bf 22.74 \ 0.5356} & \multicolumn{2}{c}{ \bf 22.95 \ 0.5458} & \multicolumn{2}{c}{ \bf 22.69 \ 0.5436} & \multicolumn{2}{c|}{\bf 22.41 \ 0.5234} & \multicolumn{2}{c}{ \bf 23.15 \ 0.5625} \\ \cline{2-20}
			& HAT  \cite{chen2023activating} & \multicolumn{2}{c}{23.78 \ 0.5976} & \multicolumn{2}{c}{23.55 \ 0.5769} & \multicolumn{2}{c}{23.03 \ 0.5586} & \multicolumn{2}{c}{23.07 \ 0.5571} & \multicolumn{2}{c}{22.65 \ 0.5306} & \multicolumn{2}{c}{22.82 \ 0.5385} & \multicolumn{2}{c}{22.65 \ 0.5406} & \multicolumn{2}{c|}{22.35 \ 0.5196} & \multicolumn{2}{c}{22.99 \ 0.5524} \\
			& +TFD & \multicolumn{2}{c}{ \bf 24.11 \ 0.6287} & \multicolumn{2}{c}{ \bf 23.97 \ 0.6137} & \multicolumn{2}{c}{ \bf 23.15 \ 0.5640} & \multicolumn{2}{c}{ \bf 23.17 \ 0.5631} & \multicolumn{2}{c}{ \bf 22.85 \ 0.5368} & \multicolumn{2}{c}{ \bf 22.91 \ 0.5449} & \multicolumn{2}{c}{ \bf 22.75 \ 0.5430} & \multicolumn{2}{c|}{\bf 22.45 \ 0.5218} & \multicolumn{2}{c}{ \bf 23.17 \ 0.5645} \\ \cline{2-20}
			& SwinIR \cite{liang2021swinir} & \multicolumn{2}{c}{23.91 \ 0.6062} & \multicolumn{2}{c}{23.83 \ 0.5870} & \multicolumn{2}{c}{23.27 \ 0.5253} & \multicolumn{2}{c}{23.04 \ 0.5610} & \multicolumn{2}{c}{22.61 \ 0.4950} & \multicolumn{2}{c}{22.82 \ 0.5432} & \multicolumn{2}{c}{22.61 \ 0.5397} & \multicolumn{2}{c|}{22.34 \ 0.5207} & \multicolumn{2}{c}{23.05 \ 0.5473} \\
			& +TFD & \multicolumn{2}{c}{ \bf 24.14 \ 0.6415} & \multicolumn{2}{c}{ \bf 24.06 \ 0.6263} & \multicolumn{2}{c}{ \bf 23.50 \ 0.5595} & \multicolumn{2}{c}{ \bf 23.27 \ 0.5684} & \multicolumn{2}{c}{ \bf 22.84 \ 0.5331} & \multicolumn{2}{c}{ \bf 23.05 \ 0.5505} & \multicolumn{2}{c}{ \bf 22.84 \ 0.5420} & \multicolumn{2}{c|}{\bf 22.57 \ 0.5218} & \multicolumn{2}{c}{ \bf 23.28 \ 0.5679} \\

			\midrule
			
			\multirow{8}{*}{\rotatebox{90}{Urban100}}
			& SRResNet \cite{SRResNet} & \multicolumn{2}{c}{21.24 \ 0.6351} & \multicolumn{2}{c}{21.06 \ 0.6090} & \multicolumn{2}{c}{20.82 \ 0.5656} & \multicolumn{2}{c}{20.60 \ 0.5949} & \multicolumn{2}{c}{20.46 \ 0.5277} & \multicolumn{2}{c}{20.30 \ 0.5652} & \multicolumn{2}{c}{20.43 \ 0.5761} & \multicolumn{2}{c|}{20.10 \ 0.5436} & \multicolumn{2}{c}{20.63 \ 0.5771} \\
			& +TFD & \multicolumn{2}{c}{ \bf 22.28 \ 0.6351} & \multicolumn{2}{c}{ \bf 21.89 \ 0.6090} & \multicolumn{2}{c}{ \bf 21.20 \ 0.5656} & \multicolumn{2}{c}{ \bf 21.30 \ 0.5949} & \multicolumn{2}{c}{ \bf 20.52 \ 0.5277} & \multicolumn{2}{c}{ \bf 20.84 \ 0.5652} & \multicolumn{2}{c}{ \bf 20.87 \ 0.5761} & \multicolumn{2}{c|}{\bf 20.33 \ 0.5436} & \multicolumn{2}{c}{ \bf 21.15 \ 0.5771} \\ \cline{2-20}
			& RRDB \cite{ESRGAN} & \multicolumn{2}{c}{21.57 \ 0.6404} & \multicolumn{2}{c}{21.18 \ 0.6106} & \multicolumn{2}{c}{19.61 \ 0.5487} & \multicolumn{2}{c}{20.93 \ 0.5996} & \multicolumn{2}{c}{20.57 \ 0.5297} & \multicolumn{2}{c}{20.40 \ 0.5667} & \multicolumn{2}{c}{20.74 \ 0.5807} & \multicolumn{2}{c|}{20.24 \ 0.5458} & \multicolumn{2}{c}{20.66 \ 0.5778} \\
			& +TFD & \multicolumn{2}{c}{ \bf 22.44 \ 0.6654} & \multicolumn{2}{c}{ \bf 22.13 \ 0.6441} & \multicolumn{2}{c}{ \bf 21.66 \ 0.6166} & \multicolumn{2}{c}{ \bf 21.45 \ 0.6108} & \multicolumn{2}{c}{ \bf 20.99 \ 0.5755} & \multicolumn{2}{c}{ \bf 20.93 \ 0.5764} & \multicolumn{2}{c}{ \bf 21.09 \ 0.5887} & \multicolumn{2}{c|}{\bf 20.53 \ 0.5521} & \multicolumn{2}{c}{ \bf 21.40 \ 0.6037} \\ \cline{2-20}
			& HAT  \cite{chen2023activating} & \multicolumn{2}{c}{22.05 \ 0.6412} & \multicolumn{2}{c}{21.56 \ 0.6094} & \multicolumn{2}{c}{21.39 \ 0.6028} & \multicolumn{2}{c}{21.28 \ 0.5999} & \multicolumn{2}{c}{20.70 \ 0.5596} & \multicolumn{2}{c}{20.72 \ 0.5642} & \multicolumn{2}{c}{20.93 \ 0.5799} & \multicolumn{2}{c|}{20.34 \ 0.5427} & \multicolumn{2}{c}{21.12 \ 0.5875} \\
			& +TFD & \multicolumn{2}{c}{ \bf 22.58 \ 0.6782} & \multicolumn{2}{c}{ \bf 22.23 \ 0.6568} & \multicolumn{2}{c}{ \bf 21.79 \ 0.6284} & \multicolumn{2}{c}{ \bf 21.55 \ 0.6174} & \multicolumn{2}{c}{ \bf 21.10 \ 0.5874} & \multicolumn{2}{c}{ \bf 20.98 \ 0.5837} & \multicolumn{2}{c}{ \bf 21.18 \ 0.5986} & \multicolumn{2}{c|}{\bf 20.58 \ 0.5610} & \multicolumn{2}{c}{ \bf 21.50 \ 0.6139} \\ \cline{2-20}
			& SwinIR \cite{liang2021swinir} & \multicolumn{2}{c}{22.18 \ 0.6489} & \multicolumn{2}{c}{21.90 \ 0.6204} & \multicolumn{2}{c}{20.56 \ 0.5614} & \multicolumn{2}{c}{21.32 \ 0.6050} & \multicolumn{2}{c}{20.89 \ 0.5350} & \multicolumn{2}{c}{20.79 \ 0.5724} & \multicolumn{2}{c}{20.98 \ 0.5841} & \multicolumn{2}{c|}{20.45 \ 0.5498} & \multicolumn{2}{c}{21.13 \ 0.5846} \\
			& +TFD & \multicolumn{2}{c}{ \bf 22.63 \ 0.6923} & \multicolumn{2}{c}{ \bf 22.31 \ 0.6725} & \multicolumn{2}{c}{ \bf 21.61 \ 0.6243} & \multicolumn{2}{c}{ \bf 21.47 \ 0.6228} & \multicolumn{2}{c}{ \bf 20.95 \ 0.5829} & \multicolumn{2}{c}{ \bf 20.93 \ 0.5894} & \multicolumn{2}{c}{ \bf 21.08 \ 0.5982} & \multicolumn{2}{c|}{\bf 20.55 \ 0.5618} & \multicolumn{2}{c}{ \bf 21.44 \ 0.6180} \\
			\midrule
			
			\multirow{8}{*}{\rotatebox{90}{Manga109}}
			& SRResNet \cite{SRResNet} & \multicolumn{2}{c}{18.42 \ 0.6467} & \multicolumn{2}{c}{18.75 \ 0.6453} & \multicolumn{2}{c}{18.32 \ 0.5903} & \multicolumn{2}{c}{18.30 \ 0.6266} & \multicolumn{2}{c}{18.60 \ 0.5851} & \multicolumn{2}{c}{18.53 \ 0.6226} & \multicolumn{2}{c}{18.25 \ 0.6142} & \multicolumn{2}{c|}{18.43 \ 0.6091} & \multicolumn{2}{c}{18.45 \ 0.6175} \\
			& +TFD & \multicolumn{2}{c}{ \bf 19.22 \ 0.6467} & \multicolumn{2}{c}{ \bf 19.52 \ 0.6453} & \multicolumn{2}{c}{ \bf 18.98 \ 0.5903} & \multicolumn{2}{c}{ \bf 18.96 \ 0.6266} & \multicolumn{2}{c}{ \bf 19.14 \ 0.5851} & \multicolumn{2}{c}{ \bf 19.11 \ 0.6226} & \multicolumn{2}{c}{ \bf 18.83 \ 0.6142} & \multicolumn{2}{c|}{\bf 18.92 \ 0.6091} & \multicolumn{2}{c}{ \bf 19.09 \ 0.6175} \\ \cline{2-20}
			& RRDB \cite{ESRGAN} & \multicolumn{2}{c}{18.59 \ 0.6498} & \multicolumn{2}{c}{18.64 \ 0.6437} & \multicolumn{2}{c}{18.30 \ 0.5900} & \multicolumn{2}{c}{18.41 \ 0.6285} & \multicolumn{2}{c}{18.83 \ 0.5886} & \multicolumn{2}{c}{18.43 \ 0.6208} & \multicolumn{2}{c}{18.38 \ 0.6167} & \multicolumn{2}{c|}{18.41 \ 0.6088} & \multicolumn{2}{c}{18.50 \ 0.6183} \\
			& +TFD & \multicolumn{2}{c}{ \bf 19.28 \ 0.6632} & \multicolumn{2}{c}{ \bf 18.64 \ 0.6437} & \multicolumn{2}{c}{ \bf 19.09 \ 0.6352} & \multicolumn{2}{c}{ \bf 19.05 \ 0.6375} & \multicolumn{2}{c}{ \bf 19.21 \ 0.6270} & \multicolumn{2}{c}{ \bf 19.09 \ 0.6322} & \multicolumn{2}{c}{ \bf 18.84 \ 0.6197} & \multicolumn{2}{c|}{\bf 18.91 \ 0.6133} & \multicolumn{2}{c}{ \bf 19.01 \ 0.6340} \\ \cline{2-20}
			& HAT  \cite{chen2023activating} & \multicolumn{2}{c}{19.49 \ 0.6666} & \multicolumn{2}{c}{19.66 \ 0.6608} & \multicolumn{2}{c}{19.35 \ 0.6481} & \multicolumn{2}{c}{19.26 \ 0.6444} & \multicolumn{2}{c}{19.40 \ 0.6367} & \multicolumn{2}{c}{19.29 \ 0.6348} & \multicolumn{2}{c}{19.12 \ 0.6337} & \multicolumn{2}{c|}{19.10 \ 0.6237} & \multicolumn{2}{c}{19.33 \ 0.6436} \\
			& +TFD & \multicolumn{2}{c}{ \bf 19.59 \ 0.6778} & \multicolumn{2}{c}{ \bf 19.76 \ 0.6759} & \multicolumn{2}{c}{ \bf 19.34 \ 0.6579} & \multicolumn{2}{c}{ \bf 19.26 \ 0.6524} & \multicolumn{2}{c}{ \bf 19.50 \ 0.6481} & \multicolumn{2}{c}{ \bf 19.31 \ 0.6444} & \multicolumn{2}{c}{ \bf 19.12 \ 0.6399} & \multicolumn{2}{c|}{\bf 19.10 \ 0.6300} & \multicolumn{2}{c}{ \bf 19.37 \ 0.6533} \\ \cline{2-20}
			& SwinIR \cite{liang2021swinir} & \multicolumn{2}{c}{19.10 \ 0.6583} & \multicolumn{2}{c}{19.27 \ 0.6523} & \multicolumn{2}{c}{18.71 \ 0.5964} & \multicolumn{2}{c}{18.95 \ 0.6372} & \multicolumn{2}{c}{19.07 \ 0.5924} & \multicolumn{2}{c}{19.02 \ 0.6308} & \multicolumn{2}{c}{18.79 \ 0.6242} & \multicolumn{2}{c|}{18.80 \ 0.6153} & \multicolumn{2}{c}{18.96 \ 0.6259} \\
			& +TFD & \multicolumn{2}{c}{ \bf 19.20 \ 0.6739} & \multicolumn{2}{c}{ \bf 19.37 \ 0.6709} & \multicolumn{2}{c}{ \bf 19.34 \ 0.6690} & \multicolumn{2}{c}{ \bf 18.91 \ 0.6489} & \multicolumn{2}{c}{ \bf 19.17 \ 0.6417} & \multicolumn{2}{c}{ \bf 19.12 \ 0.6414} & \multicolumn{2}{c}{ \bf 18.89 \ 0.6358} & \multicolumn{2}{c|}{\bf 18.90 \ 0.6264} & \multicolumn{2}{c}{ \bf 19.34 \ 0.6690} \\ 
			\bottomrule
		\end{tabular}
	}
\end{table*}

\section{Experiments}

\noindent\textbf{Datasets.} For training, we use the DIV2K dataset \cite{DIV2K}, which contains 800 high-quality images with multi-degradations settings used in Blind SR  \cite{wang2021realesrgan}. For evaluation, we utilize five standard SR benchmark datasets: Set5~\cite{Set5}, Set14~\cite{Set14}, BSD100~\cite{BSD100}, Manga109~\cite{Manga109}, and Urban100~\cite{Urban100}. We also evaluate on real-world datasets including DIV2K validation tracks (Difficult, Wild, and Mild) and the real-world DSLR dataset \cite{cai2019toward} with Canon and Nikon subsets to demonstrate generalization capabilities.

\noindent\textbf{Degradation Modeling and Experimental Protocol.}  
Following previous works \cite{wang2024navigating, kong2022reflash, wang2021realesrgan}, we implemented a comprehensive degradation pipeline simulating real-world scenarios. We adopted ``second-order'' degradation settings with eight configurations: (1) Clean (bicubic downsampling), (2) Blur (Gaussian blur + downsampling), (3) Noise (additive Gaussian noise, $\sigma=20$, post-downsampling), (4) JPEG (compression quality=30, post-downsampling), and (5)-(8) combinations thereof. This diverse set enables thorough evaluation of model robustness and generalization across realistic distortions.

\noindent\textbf{Quantitative Results.} 
We evaluate TFD through rigorous benchmarking and real-world generalization testing.

\noindent\emph{Performance on Traditional Benchmark Datasets.}
Table~\ref{table:degradations1} highlights the performance of TFD when applied to multiple SR models. Notably, on SRResNet~\cite{SRResNet}, TFD achieves significant PSNR gains of $1.86$dB, $0.85$dB, $0.41$dB, and $0.77$dB on Set5, Set14, BSD100, and Urban100, respectively. This trend becomes even more pronounced under noise-corrupted settings, where TFD consistently improves performance across all tested architectures, averaging a $0.89$dB gain. Although our method specifically targets noise, these results suggest that by preventing the model from overfitting to noise patterns, we free up model capacity that can be redirected toward better handling all degradation types. This confirms our hypothesis that \textit{noise overfitting is a critical bottleneck for generalization in Blind SR - even when noise isn't present in the test image. The consistent improvements across different architectures demonstrate that addressing noise overfitting benefits general SR performance regardless of the underlying model structure.}

\begin{figure}[t]
	\centering
	\subfloat[SwinIR]{\includegraphics[width=0.49\linewidth]{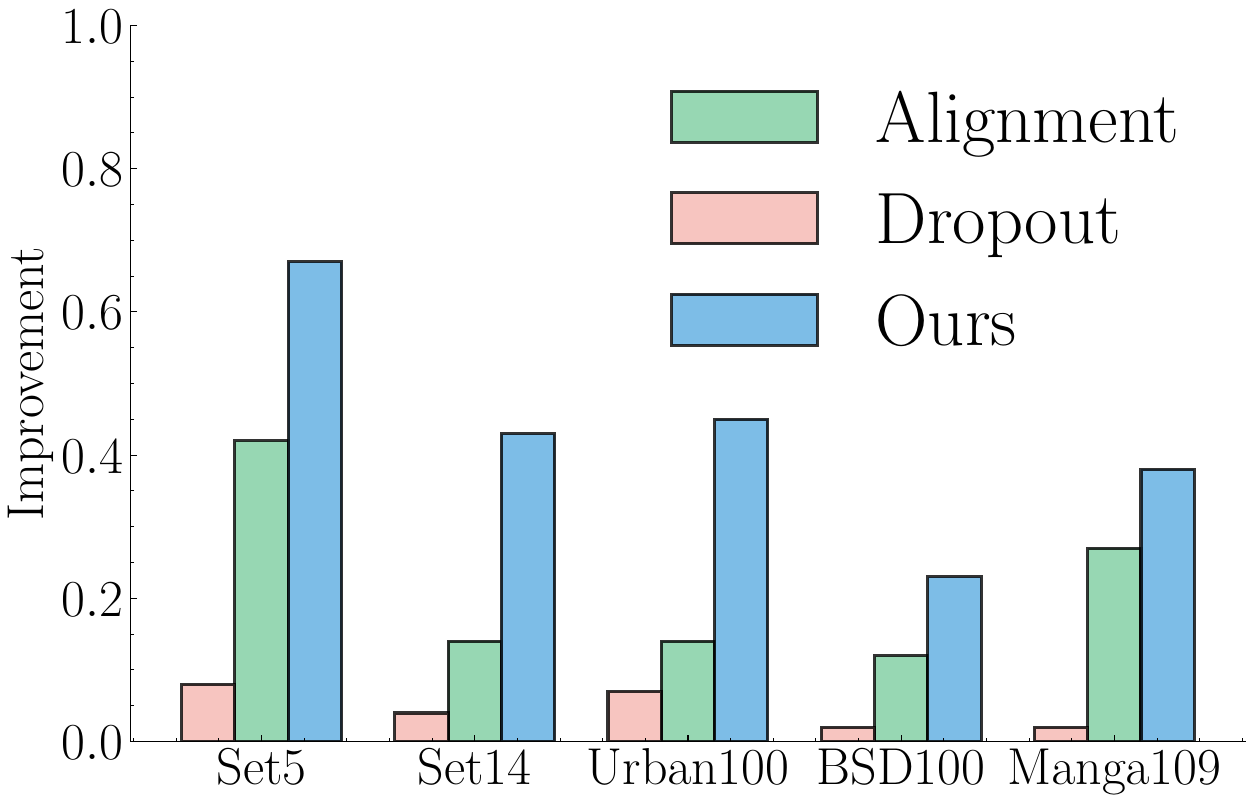}}
	\subfloat[SRResNet]{\includegraphics[width=0.49\linewidth]{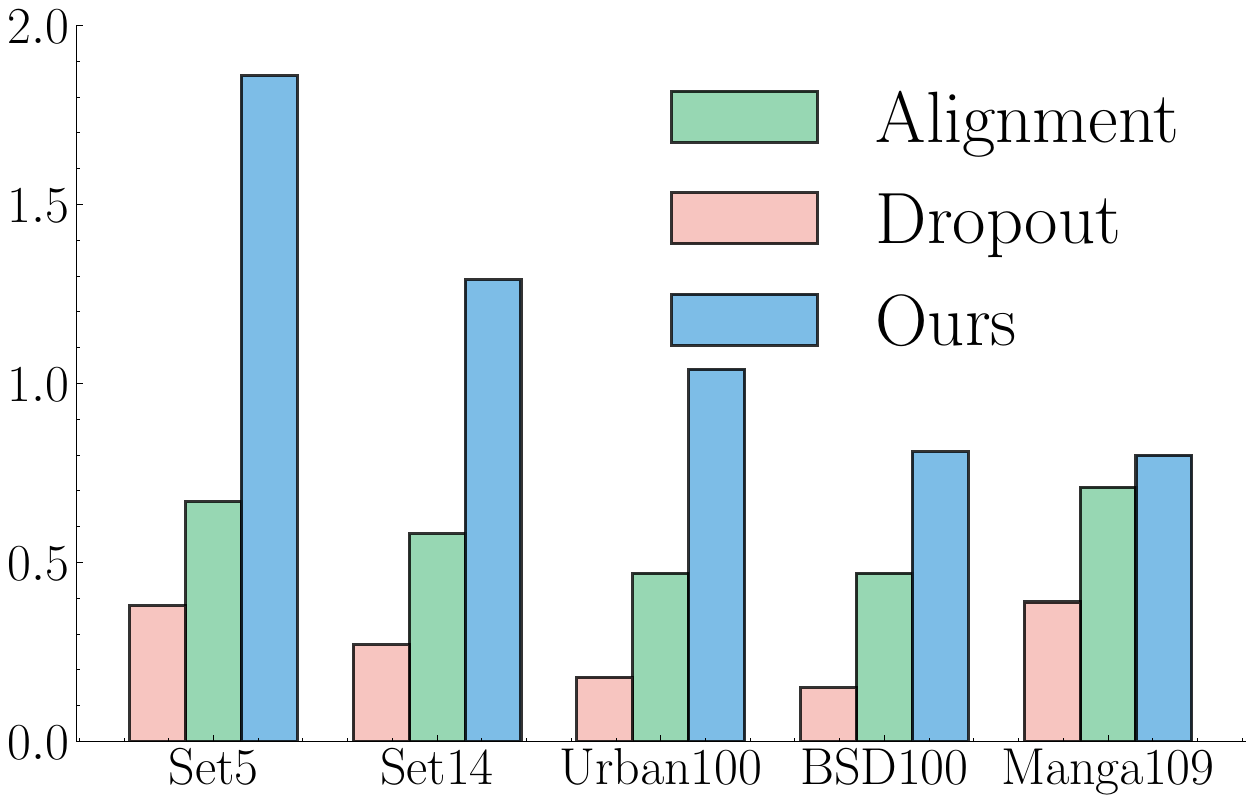}}
	\caption{\textbf{PSNR improvement comparison of different enhancement methods across benchmark datasets.}}
	\label{fig:improve}
\end{figure}

Figure~\ref{fig:improve} further reveals that TFD consistently outperforms existing regularization techniques.  \textit{The detailed data of \figref{fig:improve} is in supplementary material}. 
On SwinIR, TFD improves PSNR by $0.67$dB on Set5 and $0.45$dB on Urban100, with even larger gains observed for SRResNet (up to $1.86$dB). Compared to general-purpose regularization such as Dropout or feature alignment, which apply uniform constraints across all features, TFD explicitly disentangles noise from content features. This targeted noise suppression not only preserves fine details but also enhances robustness under compound degradations. Importantly, TFD’s architecture-agnostic design consistently benefits both transformer-based models (SwinIR, HAT) and CNN-based models (SRResNet, RRDB), demonstrating its broad applicability.

\noindent\emph{Generalization to Real-world Degradations.}
We further evaluate robustness on real-world datasets (Table~\ref{tab:real}), where TFD consistently outperforms existing methods. On DIV2K variants, TFD improves PSNR by $0.62$dB, $0.60$dB, and $0.52$dB. On Canon and Nikon datasets, TFD achieves remarkable gains of $1.50$dB and $1.46$dB. Notably, TFD reaches $25.72$dB PSNR on Canon, surpassing Dropout ($24.86$dB) and feature alignment ($25.13$dB). TFD also improves LPIPS by up to $0.020$, demonstrating superior perceptual quality. These results validate TFD’s ability to selectively suppress noise, addressing a key bottleneck in generalizable super-resolution.

\begin{table}[t]
	\centering
	\caption{\textbf{Comparison of different strategies in real setting.}}\label{tab:real}
	\resizebox{1\linewidth}{!}{
		\begin{tabular}{c|l|ccccc}
			\hline
			Dataset & Method & PSNR $\uparrow$
			& SSIM $\uparrow$
			& LPIPS $\downarrow$
			& MAD $\downarrow$
			& NLPD $\downarrow$
			\\
			\hline
			\multirowcell{4}{DIV2K \\ Difficult \\ \cite{DIV2K}}
			& SRResNet & 17.83 & 0.584 & 0.565 & 0.954 & 0.914 \\
			& +Dropout & 18.02 & 0.591 & 0.576 & 0.933 & 0.920 \\
			& +Alignment & 18.07 & 0.597 & 0.570 & 0.933 & 0.923 \\
			&  +TFD & \textbf{18.45} & \textbf{0.615} & \textbf{0.560} & \textbf{0.909} & \textbf{0.910} \\
			\hline
			\multirowcell{4}{DIV2K \\ Wild \\ \cite{DIV2K}}
			& SRResNet & 17.60 & 0.582 & 0.531 & 0.954 & 0.903 \\
			& +Dropout & 17.77 & 0.590 & 0.545 & 0.938 & 0.908 \\
			& +Alignment & 17.83 & 0.595 & 0.532 & 0.936 & 0.911 \\
			& +TFD & \textbf{18.20} & \textbf{0.605} & \textbf{0.525} & \textbf{0.927} & \textbf{0.900} \\
			\hline
			\multirowcell{4}{DIV2K \\ Mild \\ \cite{DIV2K}}
			& SRResNet & 16.98 & 0.567 & 0.522 & 0.965 & 0.895 \\
			& +Dropout & 17.14 & 0.574 & 0.536 & 0.953 & 0.900 \\
			& +Alignment & 17.21 & 0.580 & 0.522 & 0.952 & 0.903 \\
			& +TFD & \textbf{17.50} & \textbf{0.590} & \textbf{0.515} & \textbf{0.942} & \textbf{0.890} \\
			\hline
			\multirowcell{4}{Canon \\ \cite{cai2019toward}}
			& SRResNet & 24.22 & 0.787 & 0.327 & 0.629 & 0.702 \\
			& +Dropout & 24.86 & 0.792 & 0.332 & 0.580 & 0.711 \\
			& +Alignment & 25.13 & 0.795 & 0.328 & 0.566 & 0.709 \\
			& +TFD & \textbf{25.72} & \textbf{0.802} & \textbf{0.315} & \textbf{0.559} & \textbf{0.700} \\
			\hline
			\multirowcell{4}{Nikon \\ \cite{cai2019toward}}
			& SRResNet & 23.85 & 0.754 & 0.368 & 0.684 & 0.712 \\
			& +Dropout & 24.37 & 0.759 & 0.375 & 0.650 & 0.715 \\
			& +Alignment & 24.74 & 0.762 & 0.370 & 0.629 & 0.720 \\
			& +TFD & \textbf{25.31} & \textbf{0.773} & \textbf{0.348} & \textbf{0.622} & \textbf{0.700} \\
			\hline
		\end{tabular}
	}
\end{table}

\noindent\textbf{Ablation Study.}
We conduct systematic ablations to evaluate the effectiveness, efficiency, and robustness of our framework across three key dimensions.

\noindent\emph{Component Analysis.}
Table~\ref{table:ablation} presents our architectural investigation across diverse network backbones. The full TFD model delivers impressive gains across all degradation scenarios, improving PSNR by $1.86$dB (SRResNet) and $0.67$dB (SwinIR) on Set5, with particularly strong performance on noise-corrupted Urban100 images (+$1.04$dB). When analyzing individual components, we observe that both spectral-domain (FD) and spatial-domain (SD) denoising pathways provide complementary benefits, supporting our dual-path design rationale. The noise detection module proves crucial—its removal causes significant performance drops ($1.30$dB for SRResNet, $0.50$dB for SwinIR on Set5). \textit{This substantial decline occurs because the ND module enables adaptive processing through selective feature routing—clean features bypass denoising via identity mapping while corrupted features undergo restoration. Without this discriminative capability, the model applies unnecessary transformations to clean features, compromising generalization to out-of-distribution degradation patterns.}

\begin{table}[t]
	\centering
	\fontsize{8.5pt}{9.5pt}\selectfont
	\tabcolsep=6pt
	\caption{\textbf{Ablation study on Noise Detection (ND), Spatial Denoising (SD), and Frequency Domain (FD) processing. }}
	\label{table:ablation}
	\begin{tabular}{l|ccc|cc|cc}
		\whline
		\multirow{2}{*}{Model} & \multicolumn{3}{c|}{TFD Components} & \multicolumn{2}{c|}{Set5} & \multicolumn{2}{c}{Urban100} \\
		\cline{2-8}
		& ND & SD & FD & Clean & Noise & Clean & Noise \\
		\whline
		\multirow{5}{*}{SRResNet} & \ding{51} & \ding{51} & \ding{51} & \textbf{26.71} & \textbf{24.31} & \textbf{22.28} & \textbf{21.20} \\
		& {\color{gray}\ding{55}} & \ding{51} & \ding{51}  & 26.32 & 24.15 & 21.89 & 20.96 \\
		& \ding{51} & {\color{gray}\ding{55}} & \ding{51} & 26.28 & 24.08 & 21.85 & 20.92 \\
		&  \ding{51} & \ding{51} & {\color{gray}\ding{55}}& 25.41 & 23.92 & 21.43 & 20.87 \\
		& {\color{gray}\ding{55}} & {\color{gray}\ding{55}} & {\color{gray}\ding{55}} & 24.85 & 23.69 & 21.24 & 20.82 \\
		\hline
		\multirow{5}{*}{SwinIR} & \ding{51} & \ding{51} & \ding{51} & \textbf{26.92} & \textbf{24.76} & \textbf{22.63} & \textbf{21.61} \\
		& {\color{gray}\ding{55}} & \ding{51} & \ding{51} & 26.58 & 24.52 & 22.35 & 21.38 \\
		& \ding{51} & {\color{gray}\ding{55}} & \ding{51} & 26.54 & 24.48 & 22.31 & 21.35 \\
		&  \ding{51} & \ding{51} & {\color{gray}\ding{55}}  & 26.42 & 24.32 & 22.25 & 21.15 \\
		& {\color{gray}\ding{55}} & {\color{gray}\ding{55}} & {\color{gray}\ding{55}} & 26.25 & 24.15 & 22.18 & 20.56 \\
		\whline
	\end{tabular}
\end{table}

\begin{table}[t]
	\centering
	\fontsize{8.5pt}{9.5pt}\selectfont
	\tabcolsep=4.5pt
	\renewcommand{\arraystretch}{1.1}
	\caption{\textbf{Model complexity vs. performance.}}\label{tab:flops}
	\resizebox{1\linewidth}{!}{\begin{tabular}{c|ccc|ccc}
			\whline
			\multirow{2}{*}{Model} & \multicolumn{3}{c|}{Parameters \& Complexity} & \multicolumn{3}{c}{PSNR on Test Sets} \\
			\cline{2-7}
			& Params (M) & MACs & Increase & BSD100 & Urban100 & Manga109 \\
			\whline
			SwinIR & 10.72 & 571.77 & - & 23.05 & 22.18 & 19.10 \\
			+ Light  & 11.12 & 588.91 & +3.00\% & 23.21 & 22.38 & 19.23 \\
			+ Medium  & 11.68 & 606.08 & +6.00\% & 23.24 & 22.51 & 19.28 \\
			+ Full  & 12.21 & 623.37 & +9.02\% & \textbf{23.28} & \textbf{22.63} & \textbf{19.34} \\
			+ Heavy  & 12.83 & 640.71 & +12.06\% & 23.26 & 22.60 & 19.31 \\
			\hline
			HAT & 12.01 & 599.78 & - & 22.99 & 22.05 & 19.49 \\
			+ Light  & 12.42 & 616.92 & +2.86\% & 23.10 & 22.25 & 19.53 \\
			+ Medium  & 12.98 & 634.08 & +5.72\% & 23.14 & 22.41 & 19.55 \\
			+ Full  & 13.52 & 651.38 & +8.60\% & \textbf{23.17} & \textbf{22.58} & \textbf{19.59} \\
			+ Heavy  & 14.14 & 668.71 & +11.49\% & 23.16 & 22.56 & 19.57 \\
			\whline
	\end{tabular}}
\end{table}

\begin{figure*}[t]
	\centering
	\scriptsize
	\tabcolsep=2pt
	\begin{tabular}{cccccccc}
		\multirow{-7.5}{*}{\includegraphics[width=0.37\linewidth]{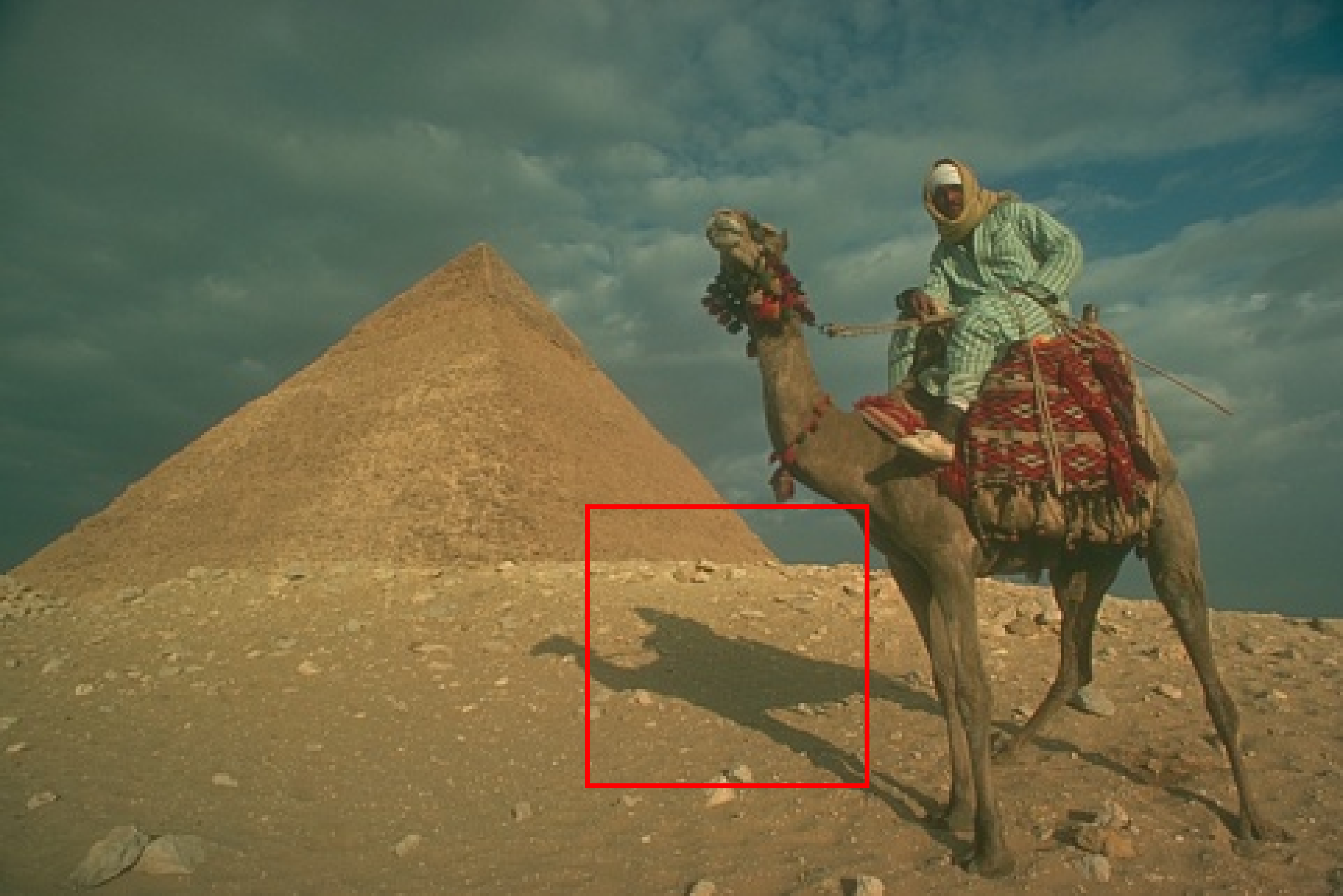}}
		& \includegraphics[width=0.115\linewidth]{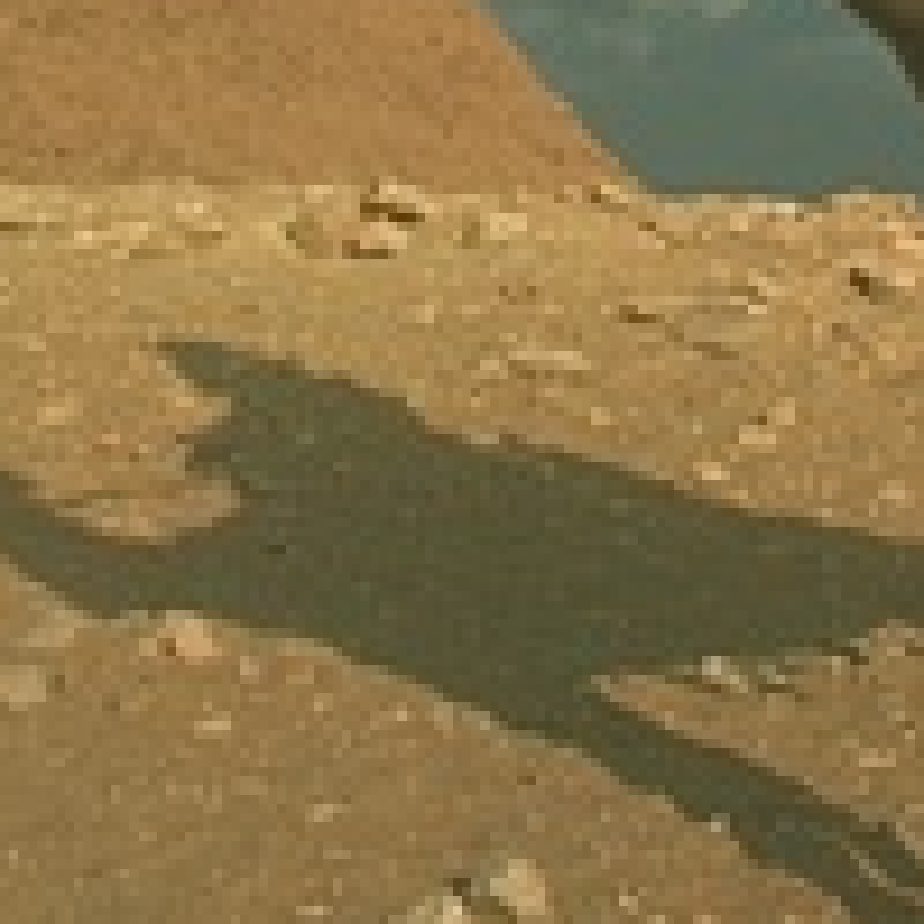}
		& \includegraphics[width=0.115\linewidth]{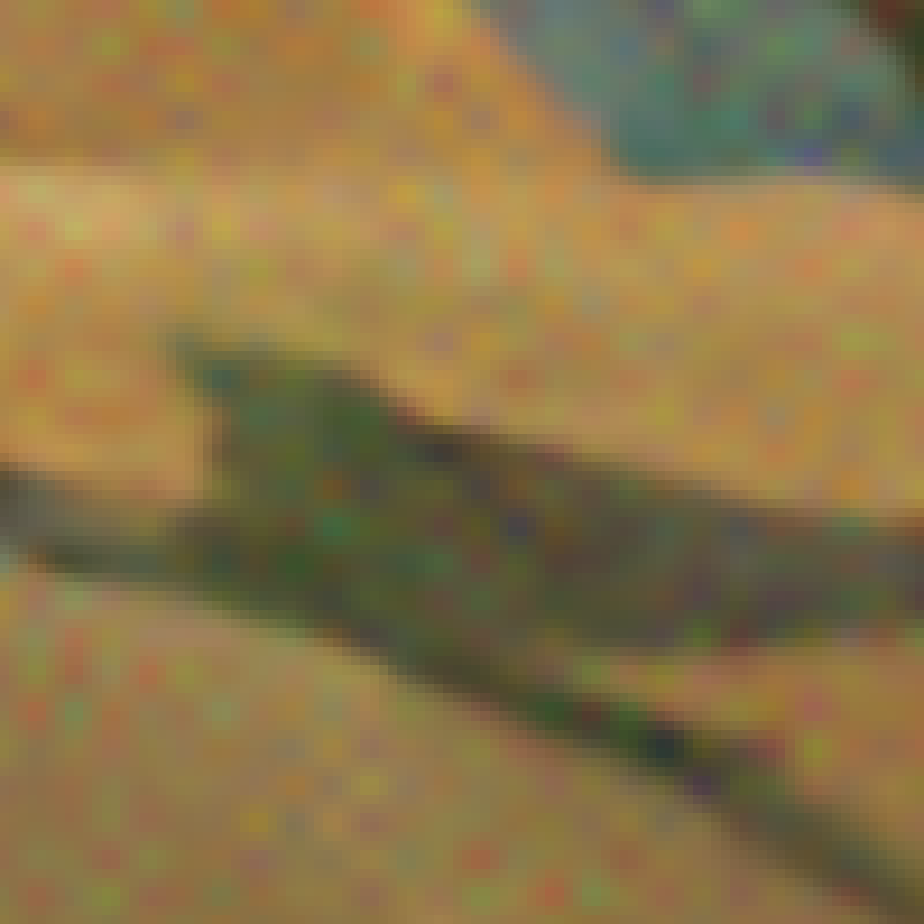}
		& \includegraphics[width=0.115\linewidth]{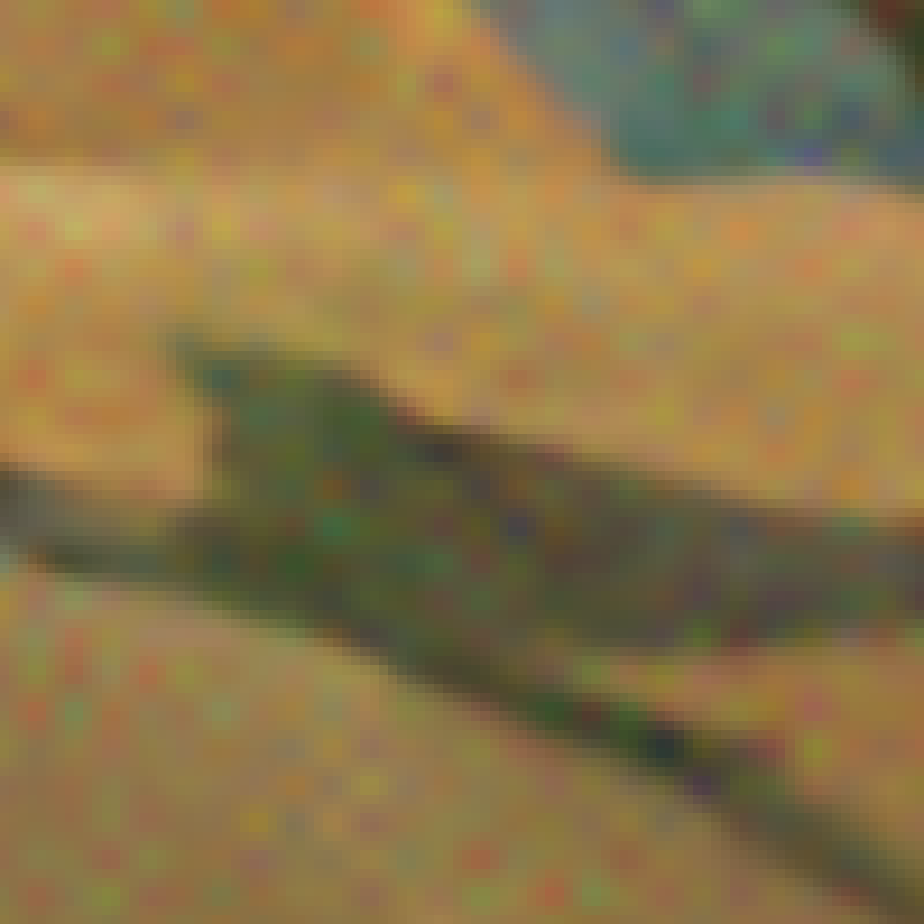}
		& \includegraphics[width=0.115\linewidth]{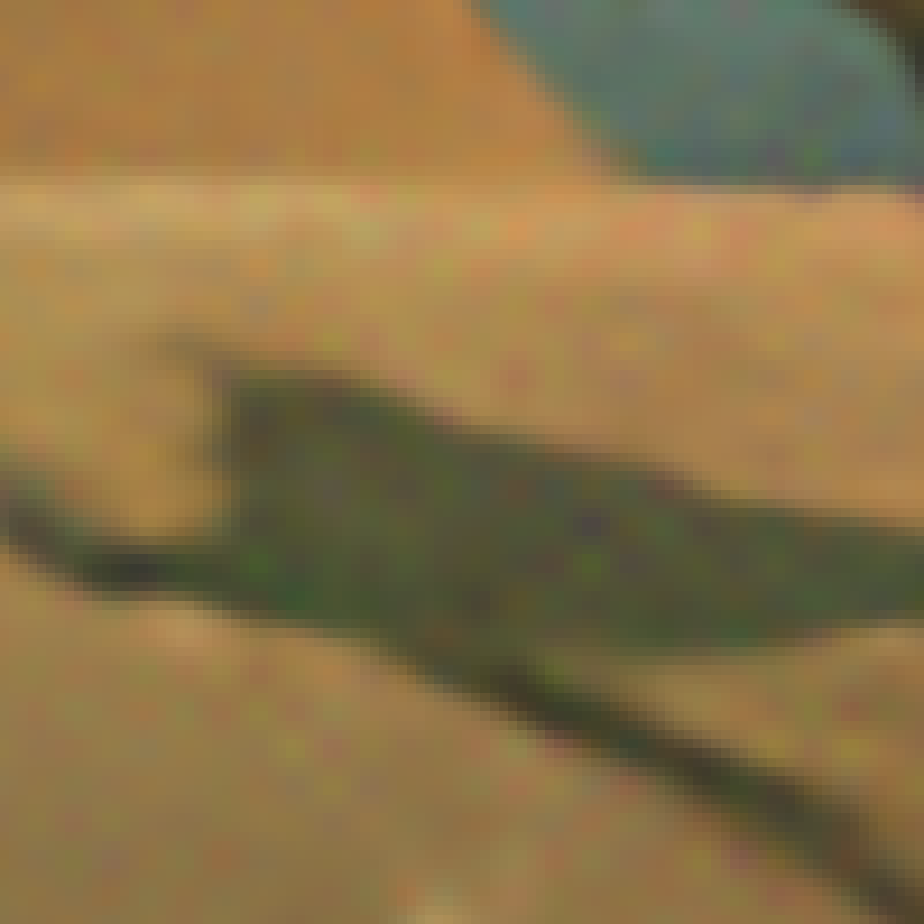}
		& \includegraphics[width=0.115\linewidth]{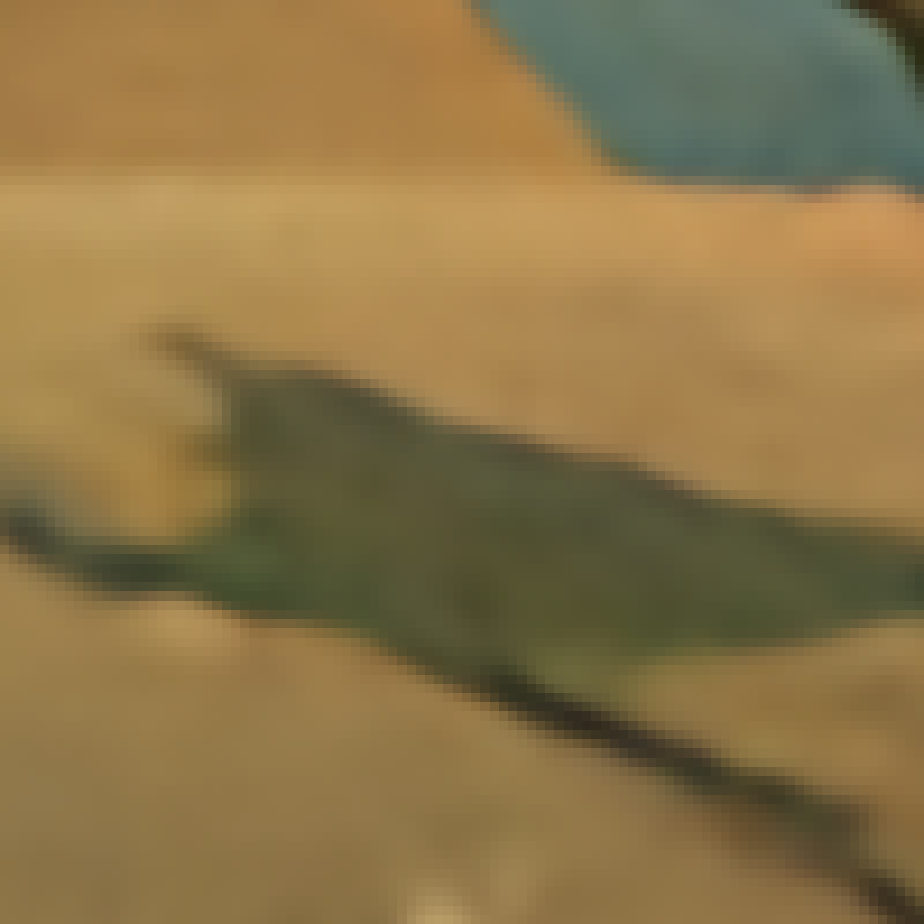}
		\\
		& GT Patch  & SRResNet & +Dropout \cite{kong2022reflash} & RRDBNet & +Dropout \cite{kong2022reflash}
		\\
		& \includegraphics[width=0.115\linewidth]{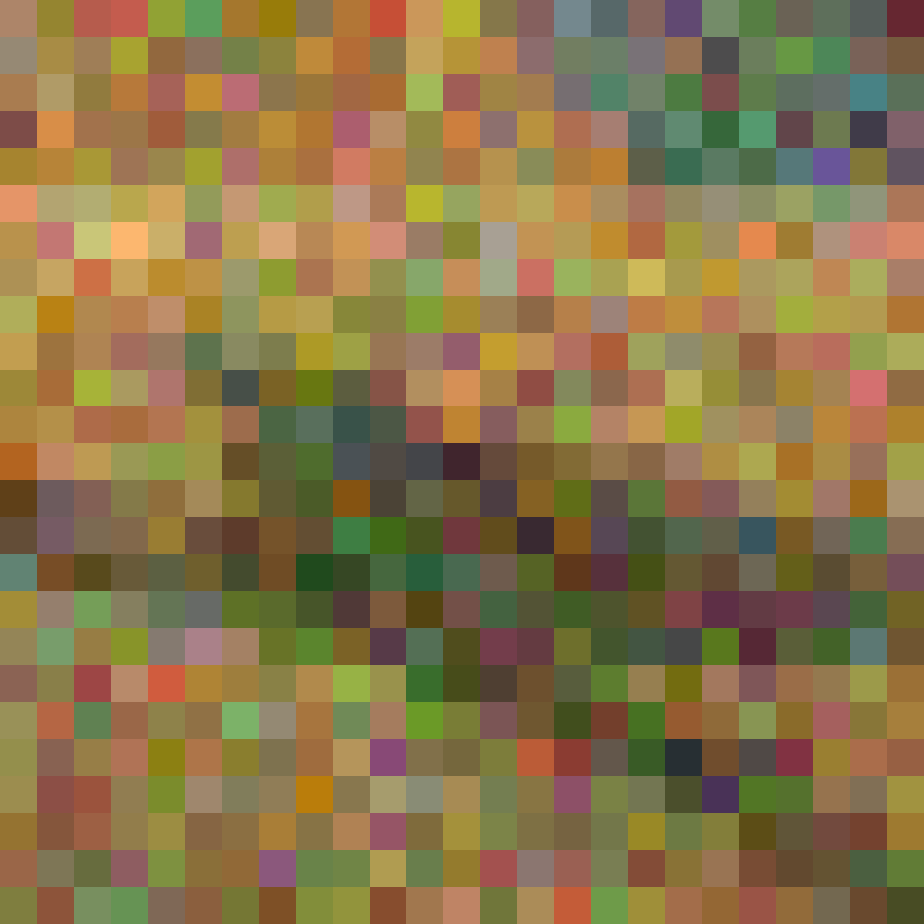}
		& \includegraphics[width=0.115\linewidth]{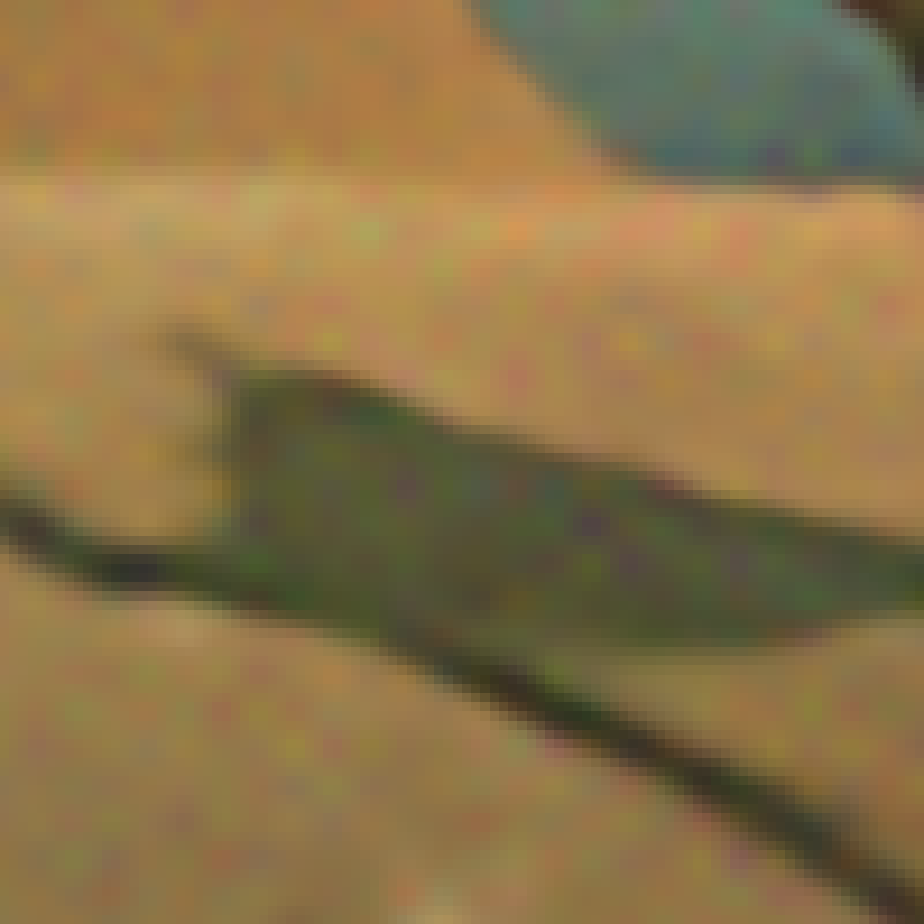}
		& \includegraphics[width=0.115\linewidth]{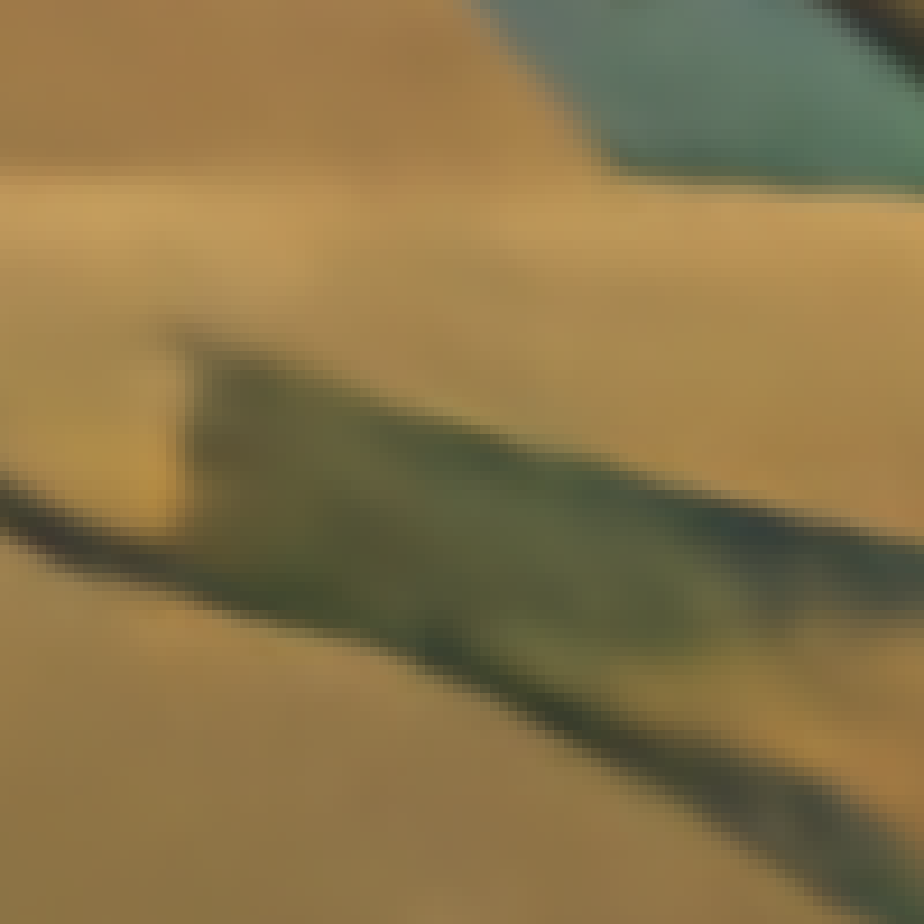}
		& \includegraphics[width=0.115\linewidth]{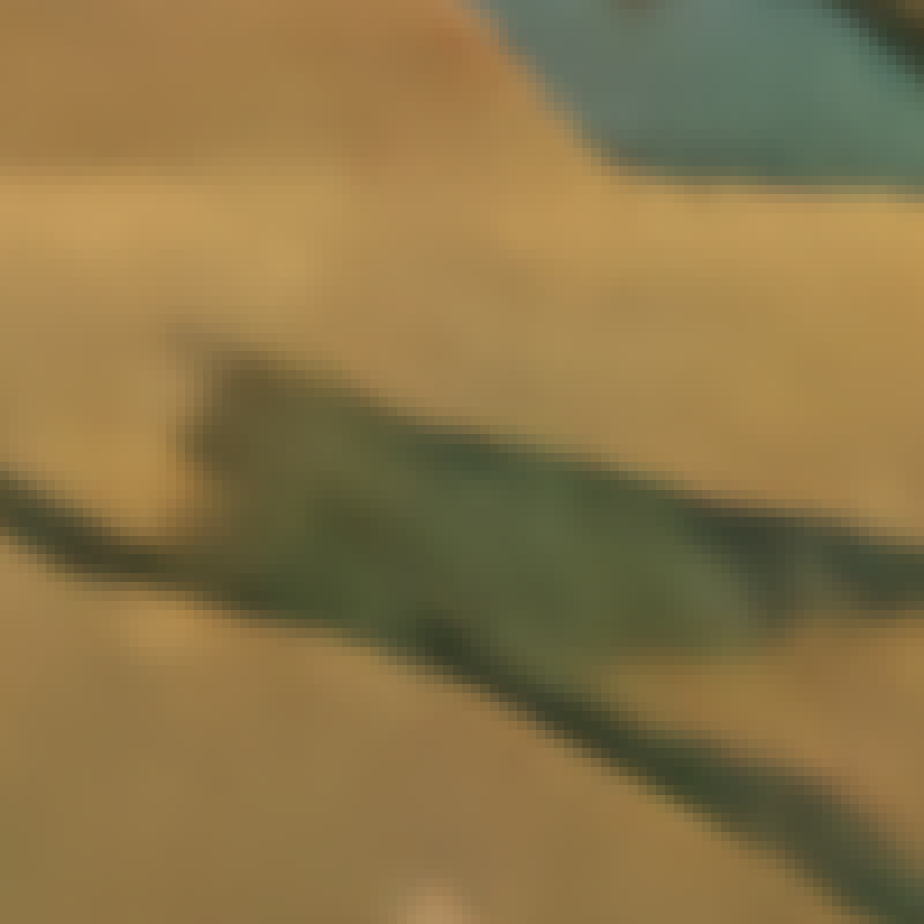}
		& \includegraphics[width=0.115\linewidth]{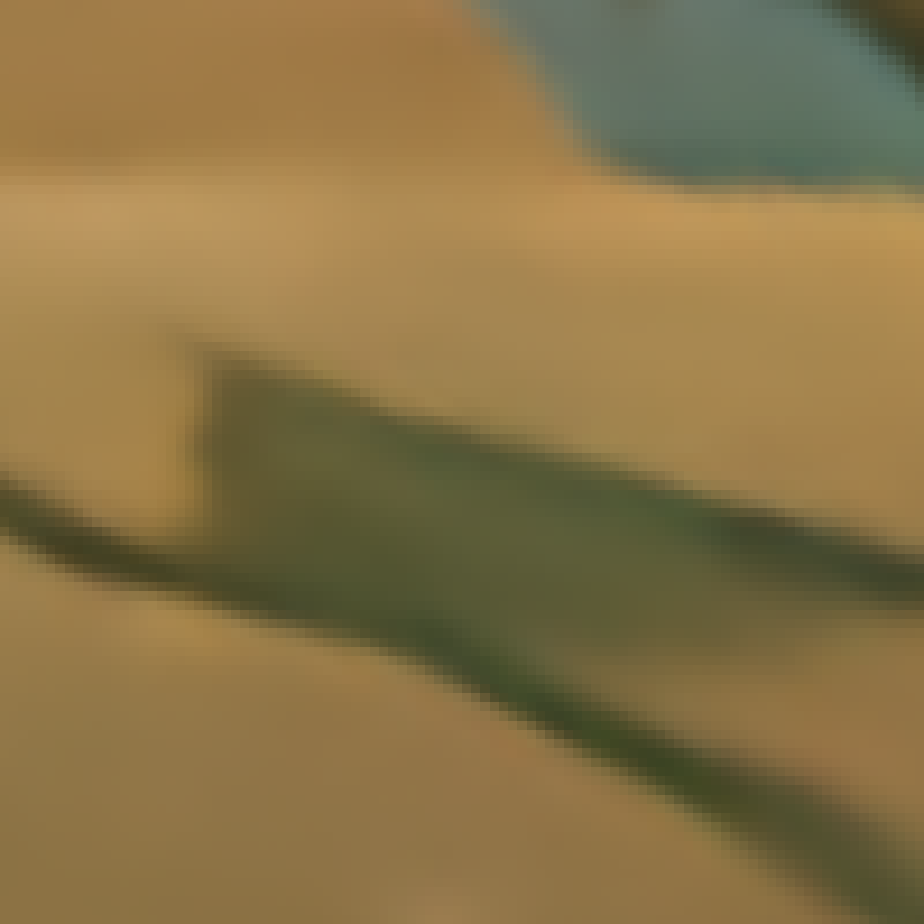}
		\\
		\multirow{-1.5}{*}{\small Reference Image} & LR Patch & +Alignment \cite{wang2024navigating} & +TFD (Ours) & +Alignment \cite{wang2024navigating} & +TFD (Ours)
	\end{tabular}
	\begin{tabular}{cccccccc}
		\multirow{-7.5}{*}{\includegraphics[width=0.37\linewidth]{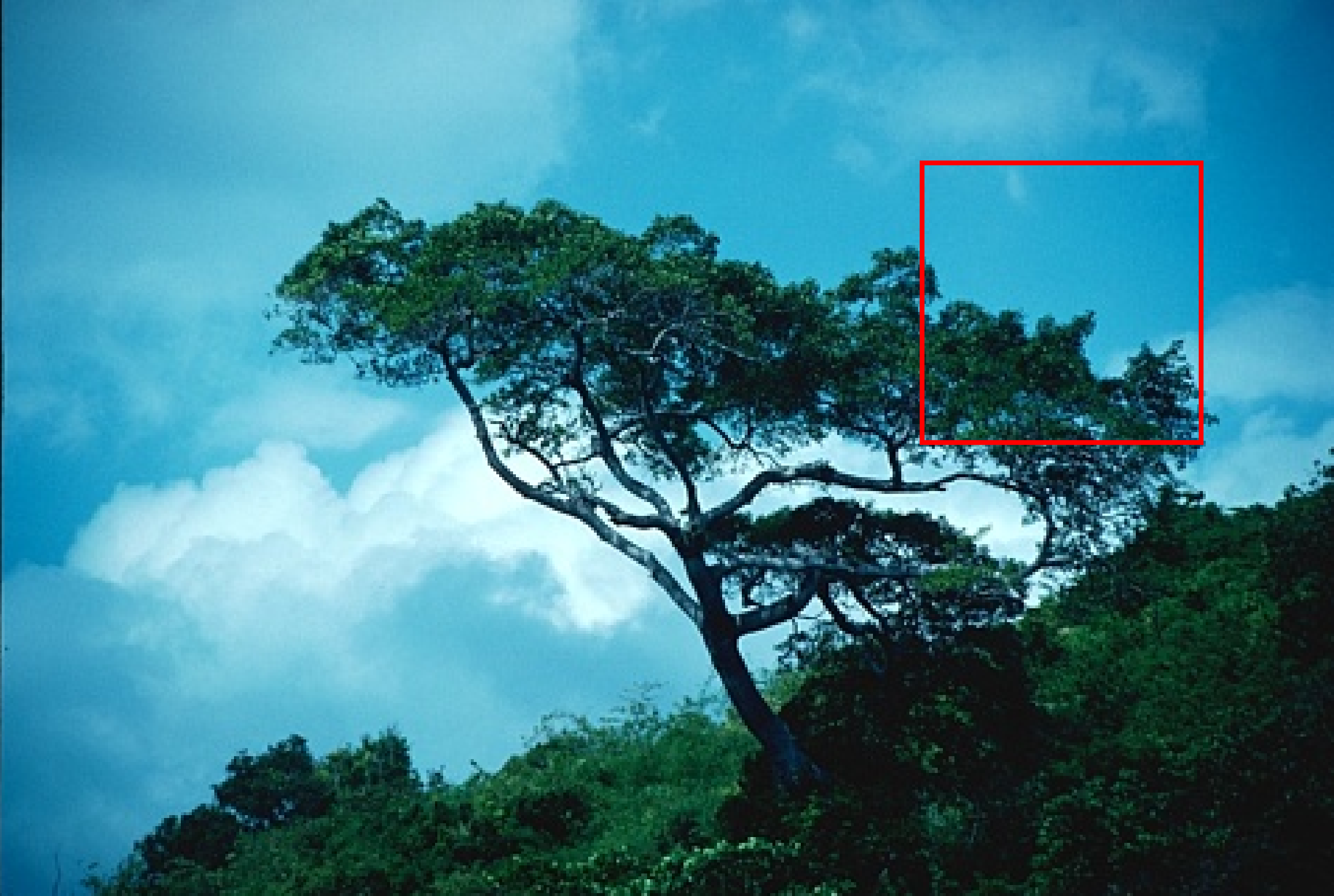}}
		& \includegraphics[width=0.115\linewidth]{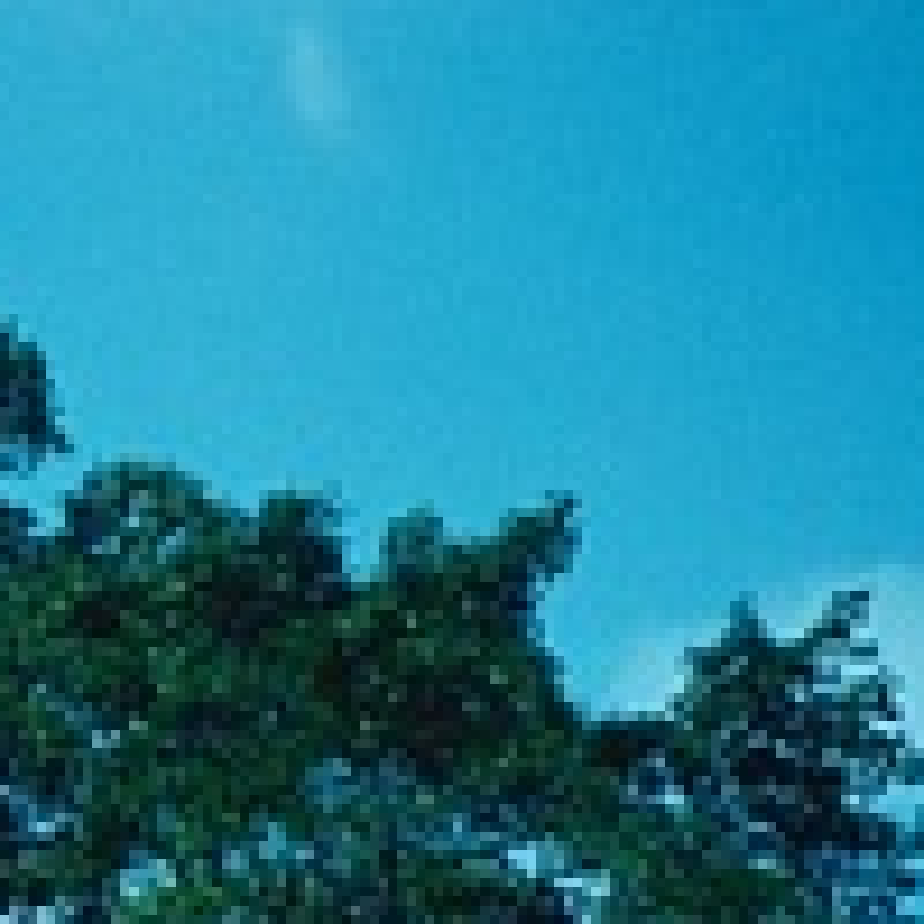}
		& \includegraphics[width=0.115\linewidth]{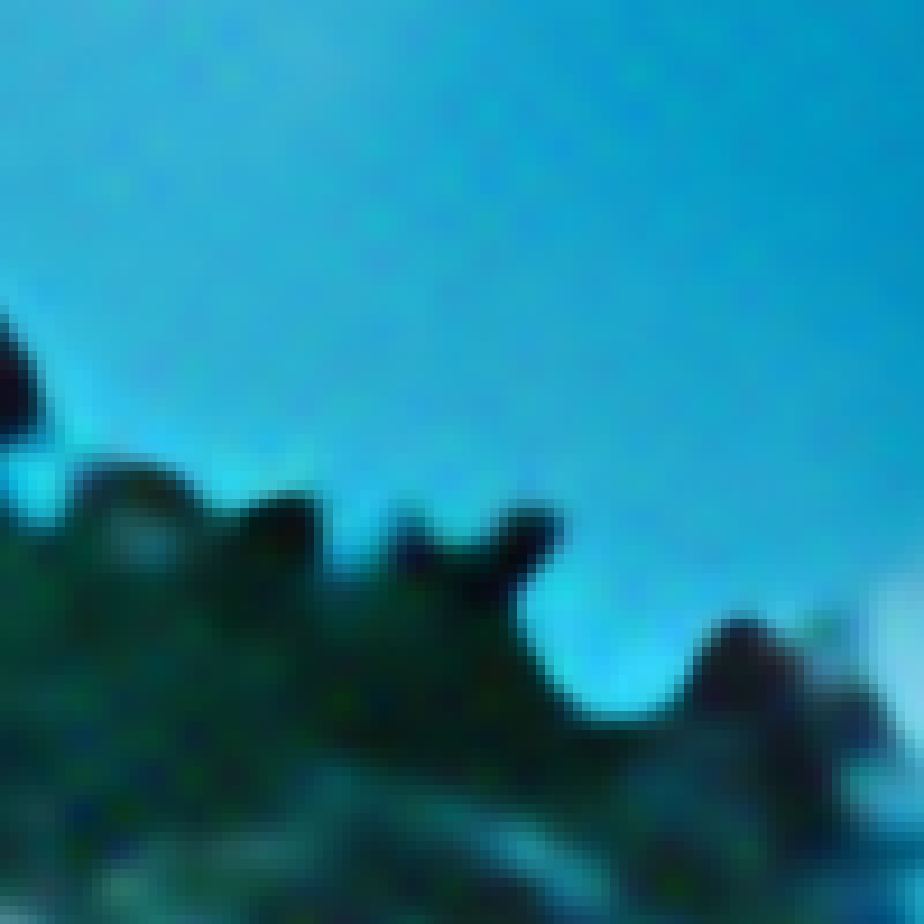}
		& \includegraphics[width=0.115\linewidth]{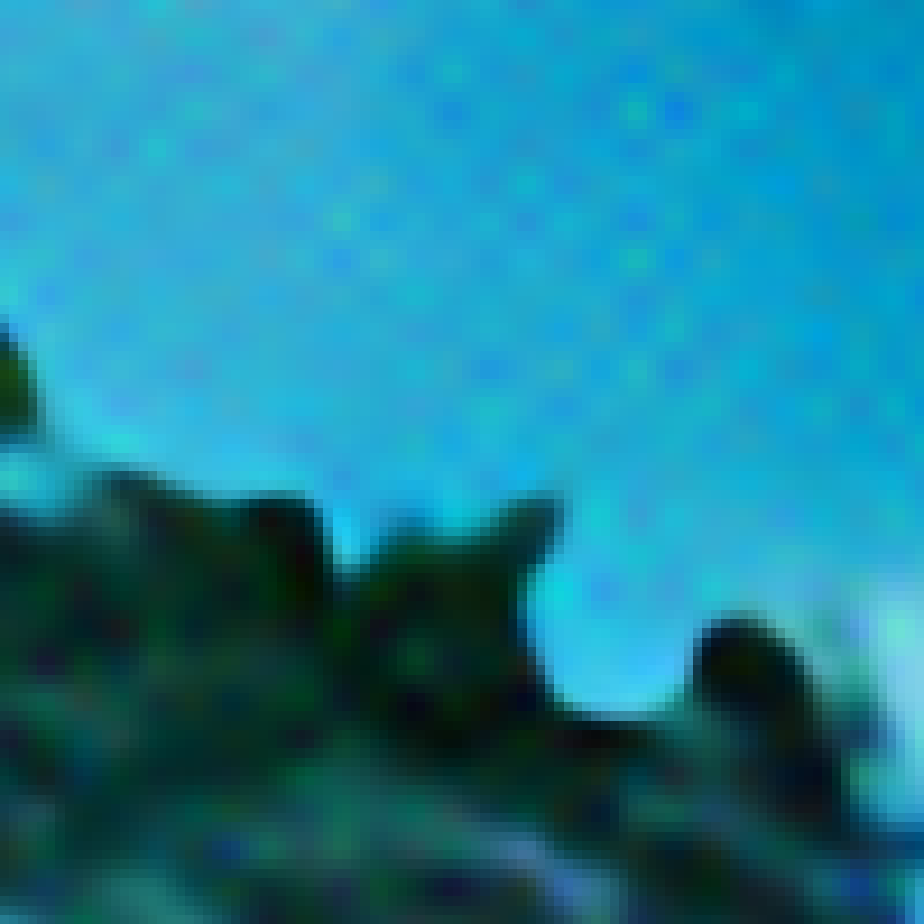}
		& \includegraphics[width=0.115\linewidth]{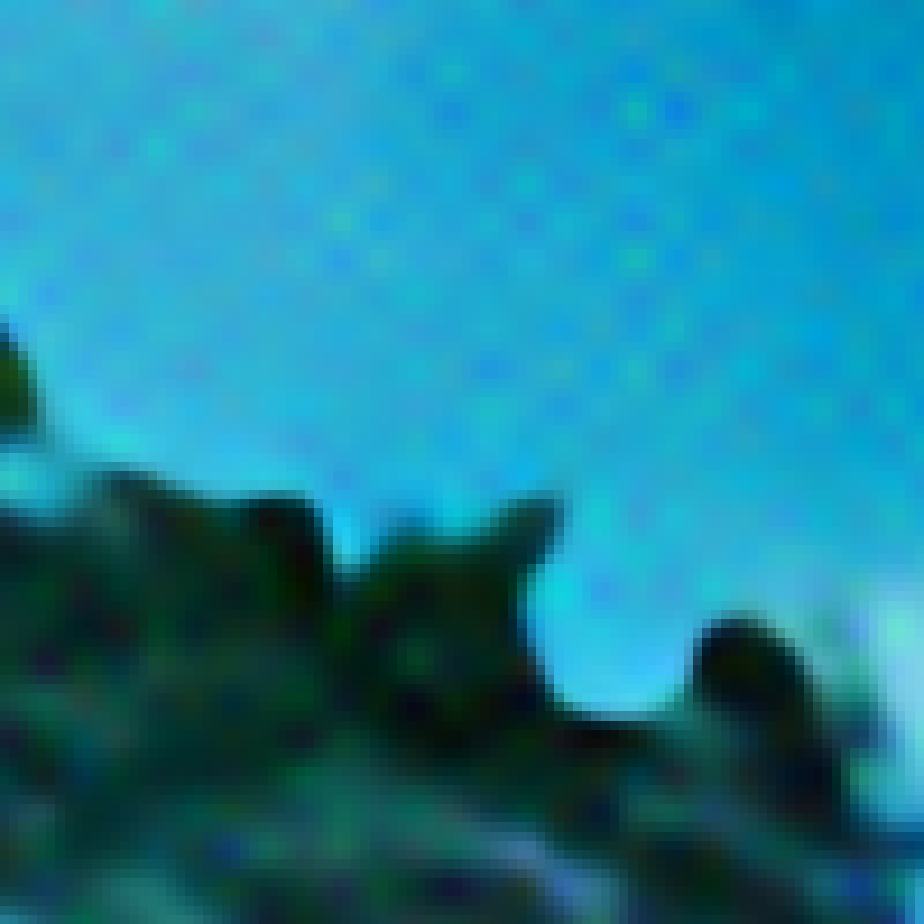}
		& \includegraphics[width=0.115\linewidth]{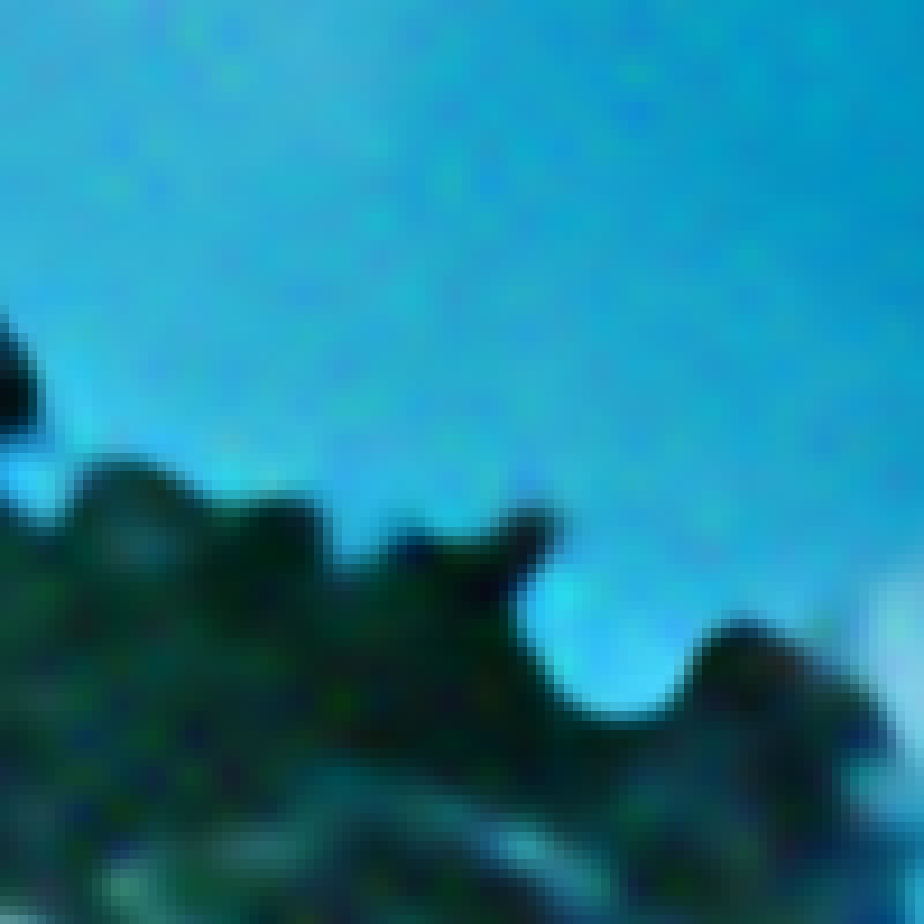}
		\\
		& GT Patch  & SRResNet & +Dropout \cite{kong2022reflash} & RRDBNet & +Dropout \cite{kong2022reflash}
		\\
		& \includegraphics[width=0.115\linewidth]{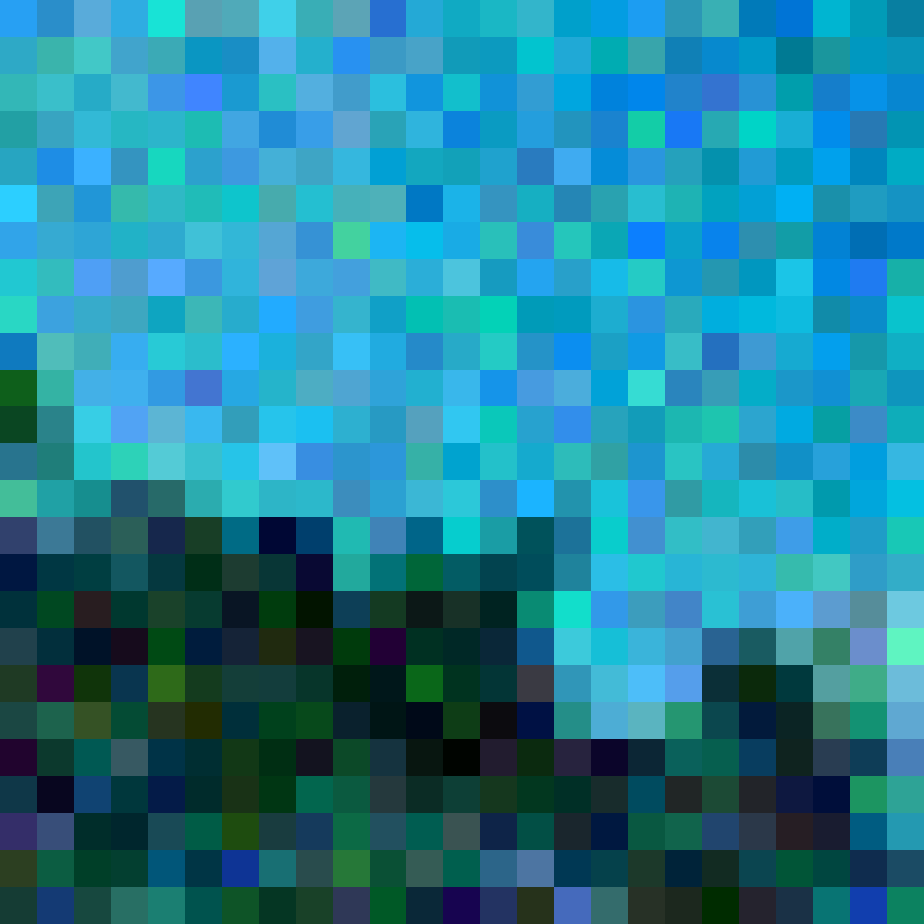}
		& \includegraphics[width=0.115\linewidth]{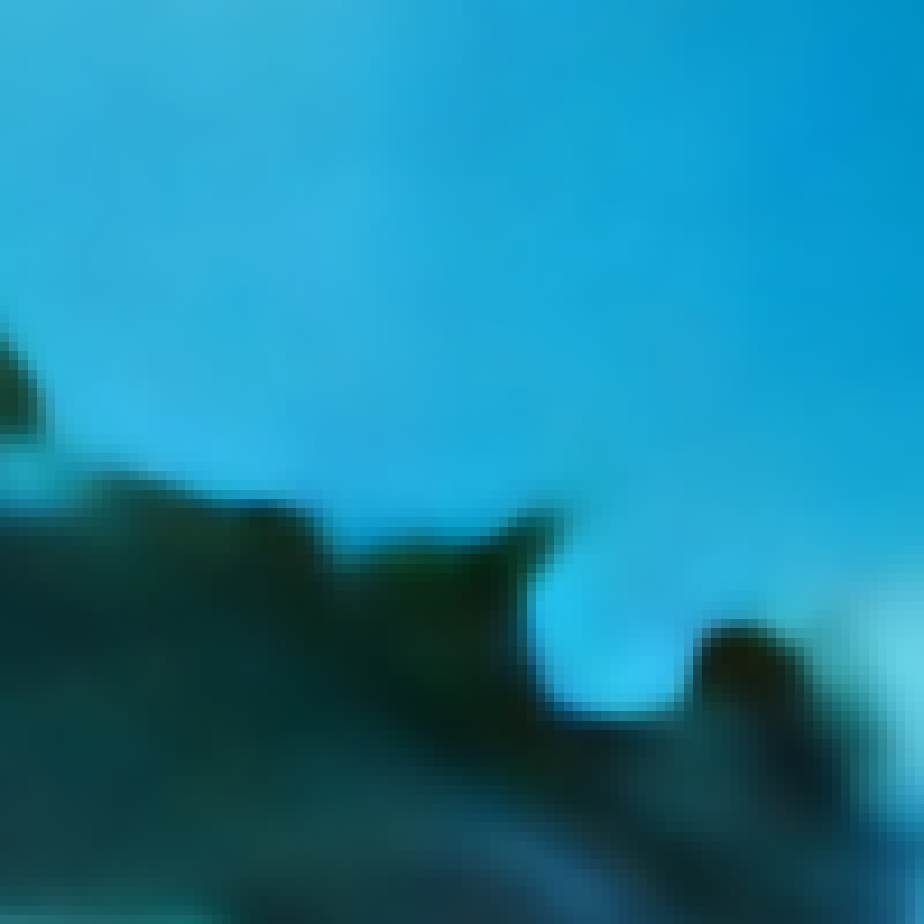}
		& \includegraphics[width=0.115\linewidth]{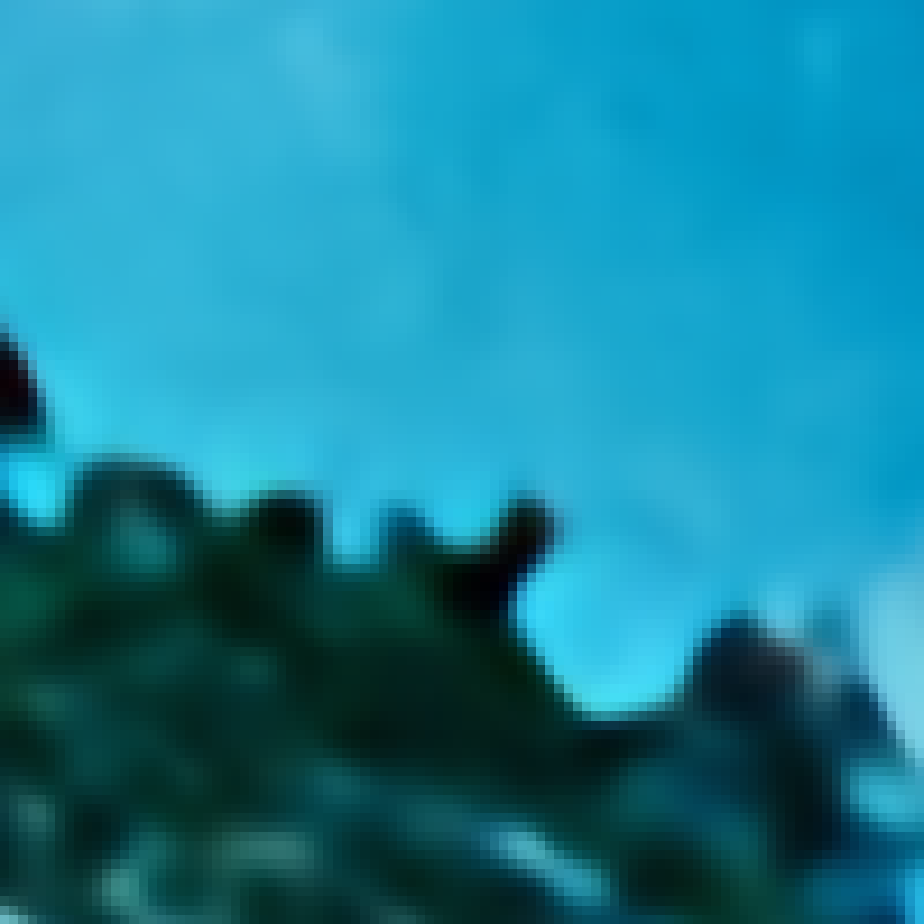}
		& \includegraphics[width=0.115\linewidth]{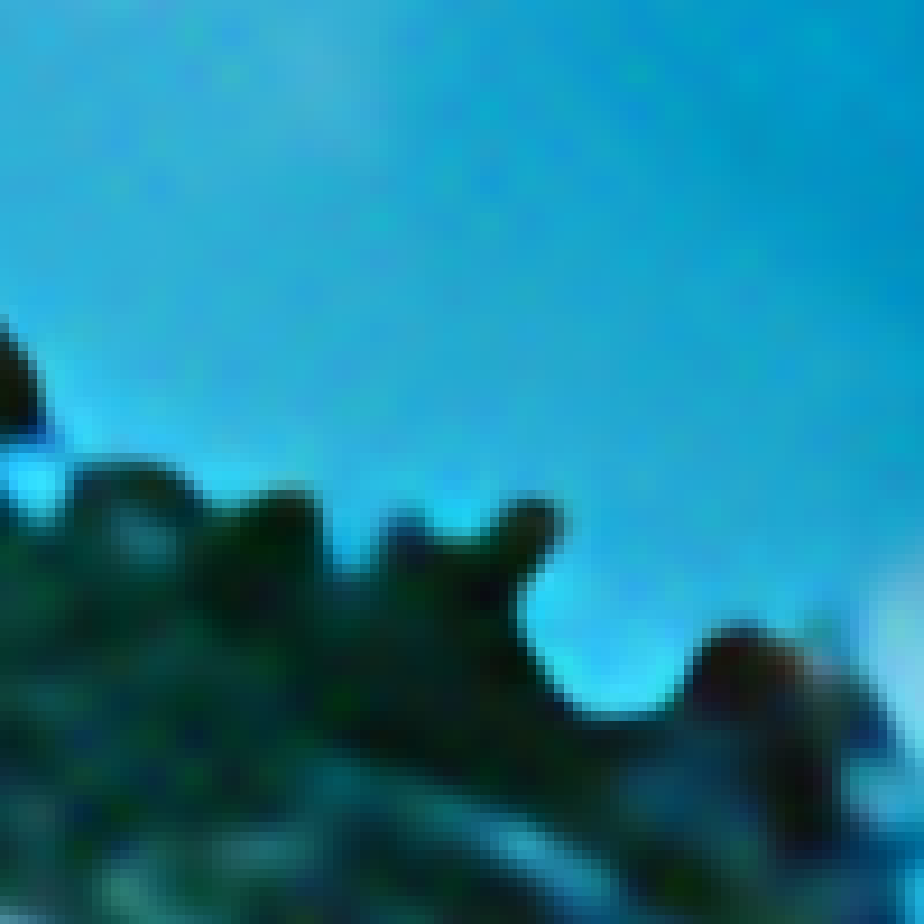}
		& \includegraphics[width=0.115\linewidth]{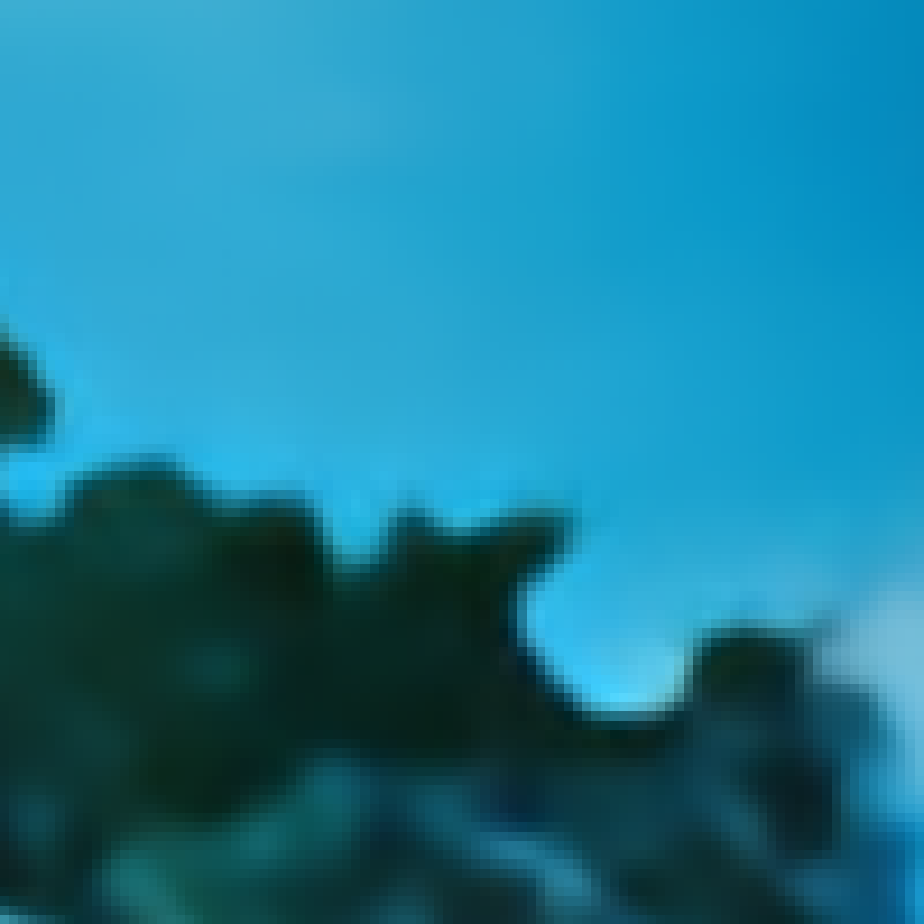}
		\\
		\multirow{-1.5}{*}{\small Reference Image} & LR Patch & +Alignment \cite{wang2024navigating} & +TFD (Ours) & +Alignment \cite{wang2024navigating} & +TFD (Ours)
	\end{tabular}
	\caption{\textbf{Visual comparison of different super-resolution methods on BSD100 dataset with bicubic\_noise20 degradation.}}
	\label{fig:vis}
\end{figure*}

\noindent\emph{Model Complexity vs. Performance.}
Our frequency analysis revealed noise's unique spectral characteristics, but implementing this insight requires efficient design. Table~\ref{tab:flops} examines this trade-off using SwinIR. The optimal configuration (Full model) adds just 51.60G MACs (+9.0\%) while delivering substantial quality improvements (+$0.41$dB on BSD100, +$0.52$dB on Urban100, +$0.64$dB on Manga109). Further increasing complexity (Heavy configuration) yields diminishing returns—a classic sign of overfitting to noise patterns rather than content features. This confirms that our dual-path denoising framework strikes an ideal balance, effectively targeting noise corruption with minimal computational overhead.

\noindent\emph{Hyperparameter Sensitivity.}
The effectiveness of our noise-content disentanglement depends on balancing classification accuracy and denoising strength. Table~\ref{tab:hyperparams} shows that $\lambda_{cls}=0.10$ and $\lambda_{denoise}=0.01$ consistently achieves optimal results across datasets and architectures, with improvements up to $1.86$dB (SRResNet) on Set5. This sweet spot reflects the inherent trade-off between noise suppression and content preservation identified in our theoretical analysis. Higher classification weights overemphasize degradation identification at the expense of reconstruction quality, while excessive denoising weights risk removing content features along with noise. The consistent optimal configuration across network paradigms demonstrates that our approach addresses a fundamental limitation in super-resolution rather than architecture-specific issues.

\begin{table}[t]
	\centering
	\fontsize{8.5pt}{9.5pt}\selectfont
	\tabcolsep=4pt
	\caption{\textbf{Analysis of loss weight hyperparameters on model performance  across benchmark datasets.}}
	\label{tab:hyperparams}
	\resizebox{1\linewidth}{!}
	{\begin{tabular}{l|cc|ccccc}
			\whline
			\multirow{2}{*}{Model} & \multicolumn{2}{c|}{Loss Weights} & \multicolumn{5}{c}{Test Sets (PSNR)} \\
			\cline{2-8}
			& $\lambda_{cls}$ & $\lambda_{denoise}$ & Set5 & Set14 & BSD100 & Urban100 & Manga109 \\
			\whline
			\multirow{5}{*}{SRResNet} 
			& 0.05 & 0.005 & 26.43 & 23.89 & 22.62 & 20.92 & 18.87 \\
			& 0.10 & 0.010 & \textbf{26.71} & \textbf{24.10} & \textbf{22.89} & \textbf{21.15} & \textbf{19.09} \\
			& 0.20 & 0.020 & 26.63 & 24.06 & 22.84 & 21.08 & 19.01 \\
			& 0.15 & 0.005 & 26.54 & 23.96 & 22.76 & 21.02 & 18.96 \\
			& 0.05 & 0.015 & 26.48 & 23.92 & 22.73 & 20.95 & 18.90 \\
			\hline
			\multirow{5}{*}{SwinIR} 
			& 0.05 & 0.005 & 26.71 & 24.78 & 23.08 & 21.22 & 19.12 \\
			& 0.10 & 0.010 & \textbf{26.92} & \textbf{24.96} & \textbf{23.28} & \textbf{21.44} & \textbf{19.34} \\
			& 0.20 & 0.020 & 26.84 & 24.90 & 23.20 & 21.39 & 19.28 \\
			& 0.15 & 0.005 & 26.77 & 24.85 & 23.15 & 21.36 & 19.23 \\
			& 0.05 & 0.015 & 26.74 & 24.81 & 23.11 & 21.32 & 19.19 \\
			\whline
	\end{tabular}}
\end{table}


\noindent\textbf{Qualitative Results.}
Figure~\ref{fig:vis} presents qualitative comparisons on challenging textures from BSD100 under the bicubic+noise20 setting. We highlight two representative examples: a pyramid with fine geometric patterns and a landscape with dense foliage—both exemplifying high-frequency details that are particularly vulnerable to noise corruption.
As shown, conventional super-resolution models (SRResNet, RRDBNet) visibly suffer from noise overfitting, leading to texture degradation and structural blurring—consistent with our hypothesis on corrupted feature representations. Existing regularization methods offer limited improvements: Dropout reduces noise at the expense of oversmoothing, while feature alignment preserves structures but retains noise residues. These observations align with our frequency analysis, confirming the spectral overlap between noise and high-frequency content as a core challenge.
In contrast, our proposed TFD framework achieves significantly cleaner and sharper reconstructions. The pyramid’s edges are crisp and well-defined, while the foliage retains natural texture without residual noise. This highlights the effectiveness of our dual-path design: the frequency branch isolates noise-corrupted components, while the spatial branch preserves semantic content, enabling perceptual quality that aligns closely with our quantitative gains.

\section{Conclusion}
This paper addresses generalizable image super-resolution by identifying noise as the primary source of feature corruption limiting model generalization. Through frequency analysis, we demonstrated noise's distinctive spectral characteristics compared to other degradations. We proposed a lightweight feature denoising framework comprising noise detection and dual-path denoising modules that selectively suppress noise-related features while preserving content details. Our model-agnostic framework integrates seamlessly with various SR architectures without structural modifications and demonstrates effectiveness on real-world images with complex degradations. 

\section*{Acknowledgment}
This work was partially supported by the Research Institute of Trustworthy Autonomous Systems at Southern University of Science and Technology, grants from the Jilin Provincial International Cooperation Key Laboratory for Super Smart City and the Jilin Provincial Key Laboratory of Intelligent Policing, the SPRING GX project of the University of Tokyo (grant number JPMJSP2108), JST-Mirai Program JPMJMI23G1, and JSPS KAKENHI Grant Numbers 24KK0209, 24K22318, and 22H00529.


{
	\small
	\bibliographystyle{ieeenat_fullname}
	\bibliography{egbib}
}


\clearpage
\setcounter{page}{1}
\setcounter{figure}{0}
\setcounter{section}{0}
\setcounter{table}{0} 
\maketitlesupplementary
\section{Experimental Details}

\noindent\textbf{Datasets and Training Setup}
For training, we utilize the DIV2K dataset~\cite{DIV2K}, which contains 800 high-quality images with diverse content. For evaluation, we employ five standard SR benchmark datasets: Set5~\cite{Set5}, Set14~\cite{Set14}, BSD100~\cite{BSD100}, Urban100~\cite{Urban100}, and Manga109~\cite{Manga109}. We also evaluate on real-world datasets including DIV2K validation tracks (Difficult, Wild, and Mild) and the real-world DSLR dataset~\cite{cai2019toward} with Canon and Nikon subsets to demonstrate generalization capabilities.
During training, we employ the L1 loss function in conjunction with the Adam optimizer ($\beta_1=0.9$, $\beta_2=0.999$). The batch size is set to 16, processing low-resolution (LR) images of size $32\times32$ pixels. We implement a cosine annealing learning rate strategy initialized at $2\times10^{-4}$ over 500,000 iterations. All experiments are conducted using the PyTorch framework on 4$\times$NVIDIA A800 GPUs.
For our targeted feature denoising framework, we adopt a multi-objective training scheme with loss weights set to $\lambda_{cls}=0.1$ and $\lambda_{feat}=0.01$. Following Theorem~\ref{thm}, training prioritizes reconstruction in early stages, deferring denoising until noise confidence surpasses 75\%. This scheduling ensures stable content preservation before handling high-frequency noise.

\noindent\textbf{Degradation Modeling Protocol}  
Following recent advances in blind image restoration~\cite{wang2021real, kong2022reflash, wang2024navigating}, we construct a comprehensive degradation pipeline to simulate diverse real-world distortions. Specifically, we adopt a \emph{second-order} degradation process~\cite{liang2021swinir}, which has become a standard benchmark for evaluating robustness and generalization. The following eight degradations are considered:

\begin{figure}
	\centering
	\includegraphics[width=1\linewidth]{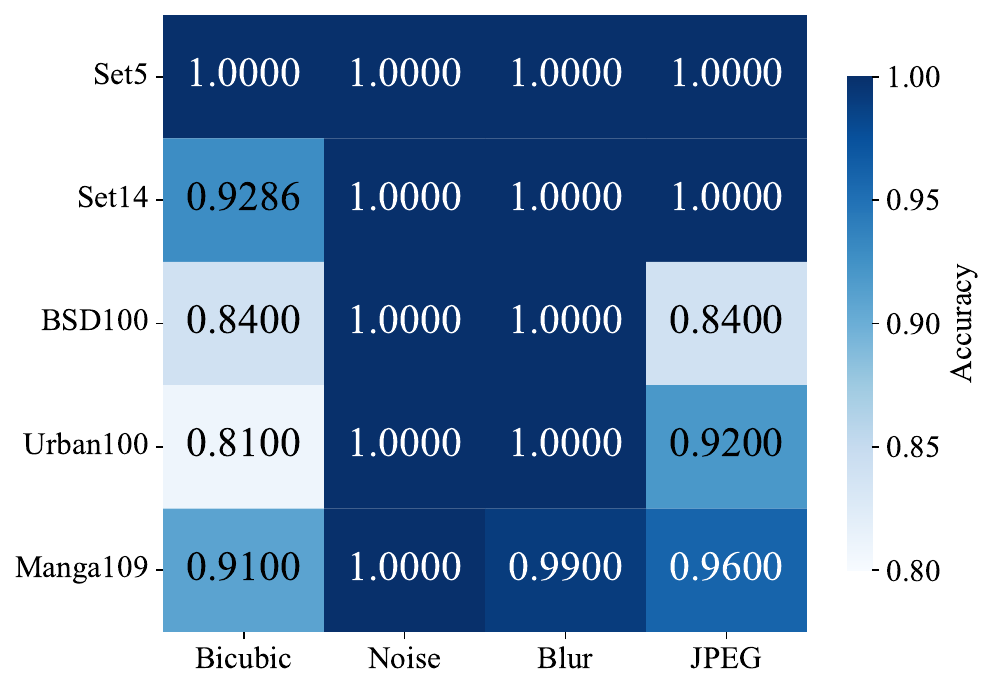}
	\caption{\textbf{Noise detection accuracy when facing unknown degradations during testing.}}
	\label{fig:noisedetectionaccuracyheatmap}
\end{figure}

\begin{figure*}[t]
	\centering
	\scriptsize
	\tabcolsep=2pt
	\begin{tabular}{cccccccc}
		\multirow{-7.5}{*}{\includegraphics[width=0.37\linewidth]{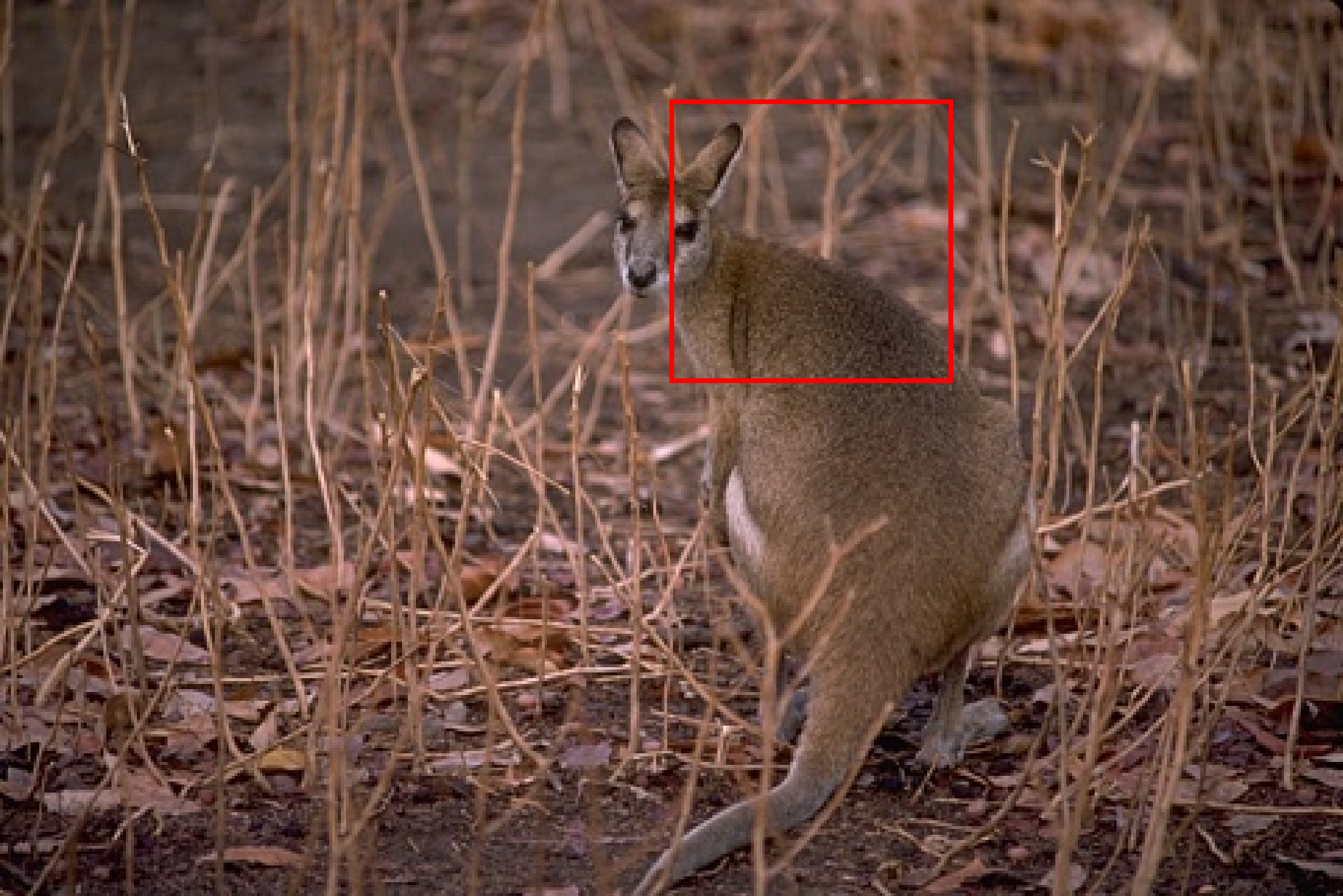}}
		& \includegraphics[width=0.115\linewidth]{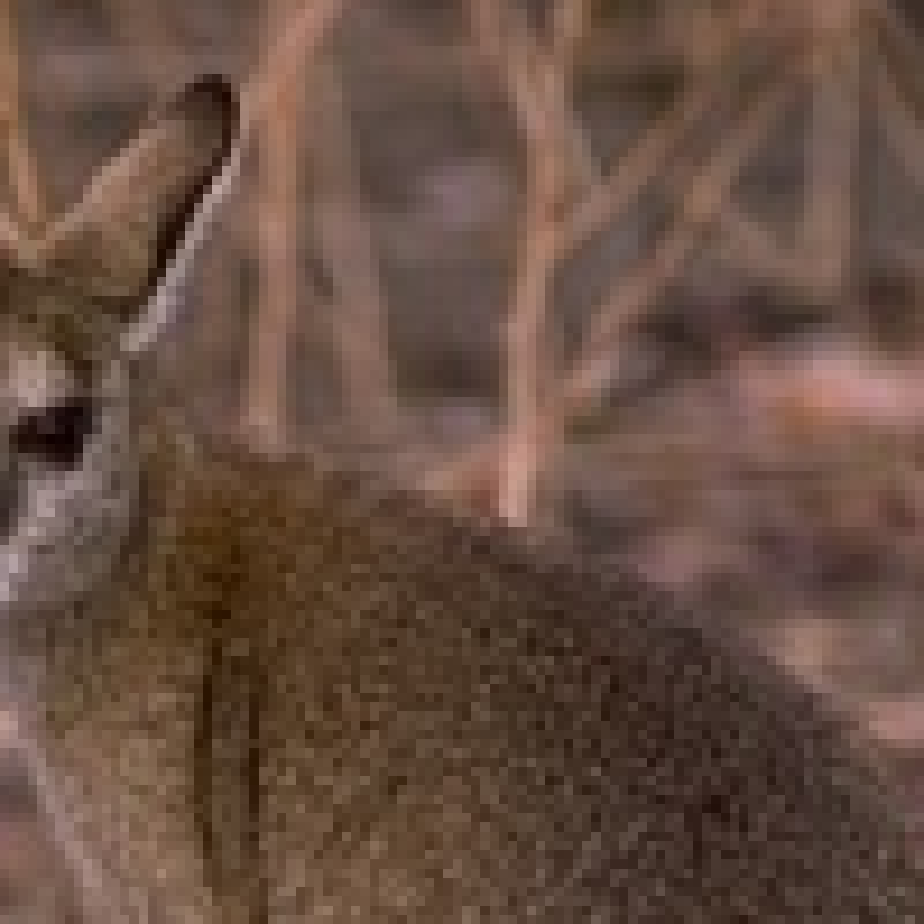}
		& \includegraphics[width=0.115\linewidth]{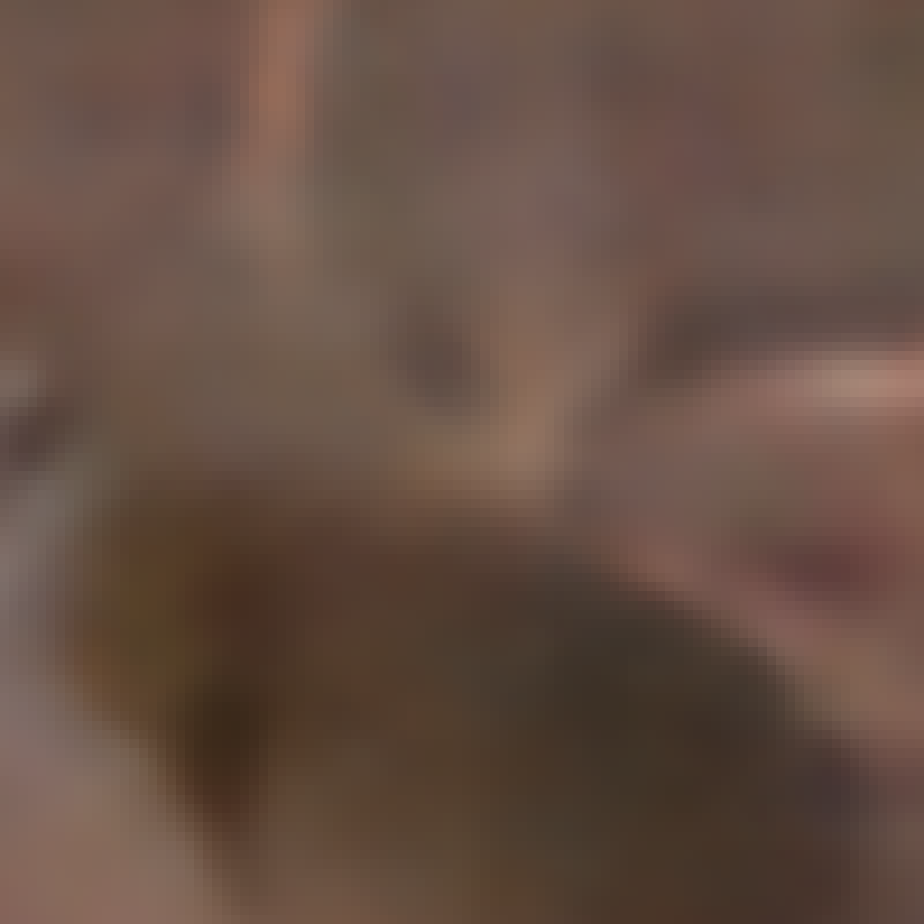}
		& \includegraphics[width=0.115\linewidth]{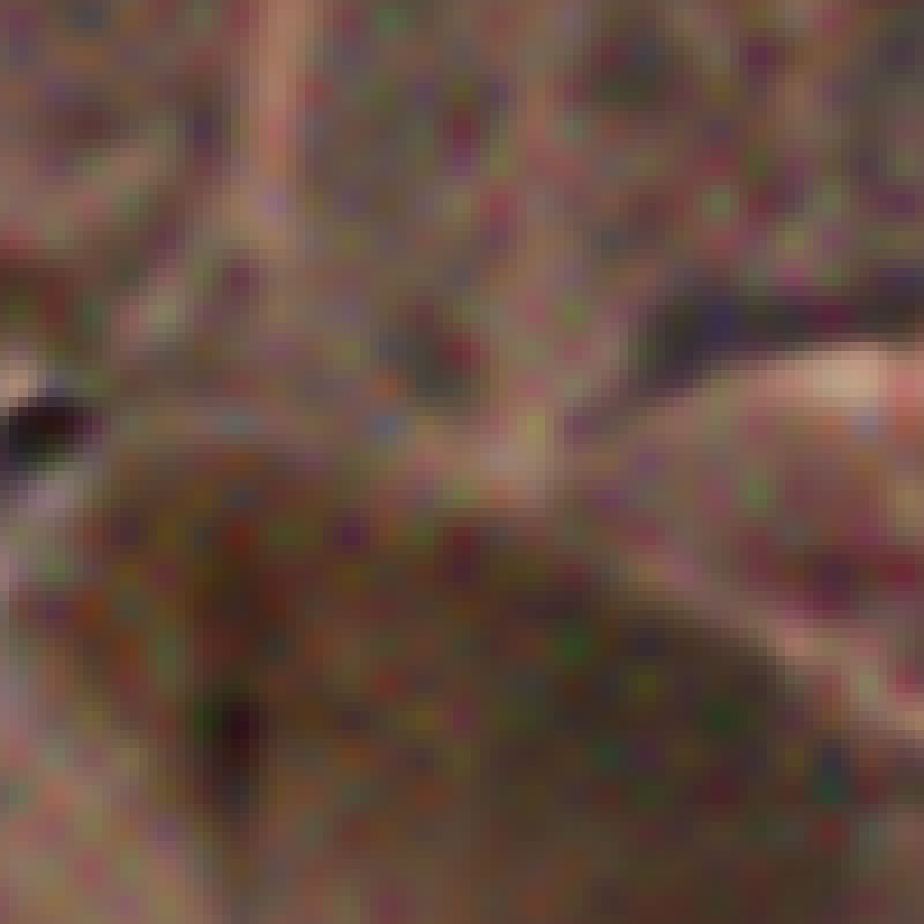}
		& \includegraphics[width=0.115\linewidth]{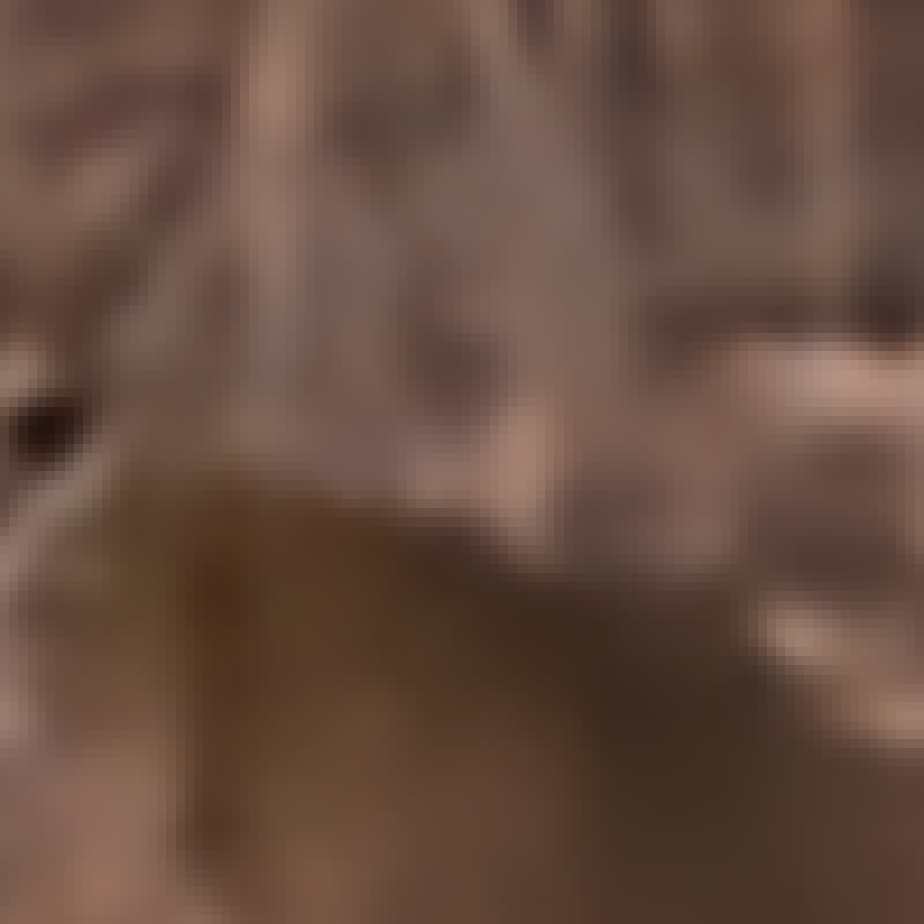}
		& \includegraphics[width=0.115\linewidth]{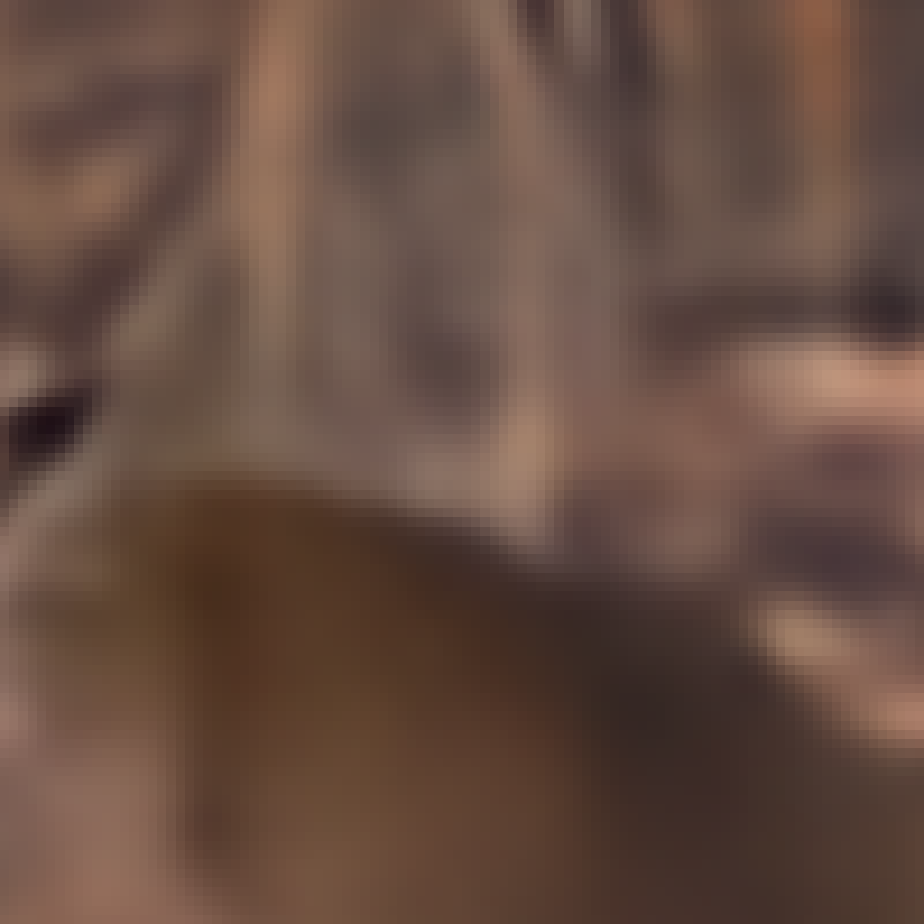}
		\\
		& GT Patch  & SWinIR & +Dropout \cite{kong2022reflash} & HAT & +Dropout \cite{kong2022reflash}
		\\
		& \includegraphics[width=0.115\linewidth]{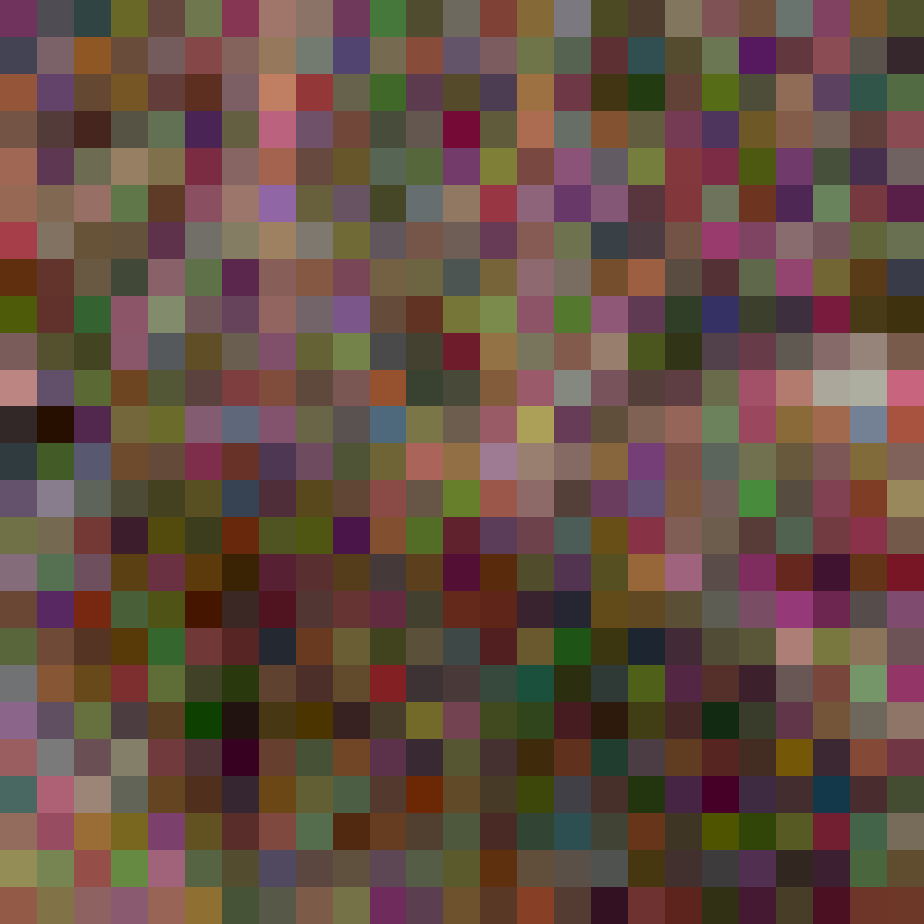}
		& \includegraphics[width=0.115\linewidth]{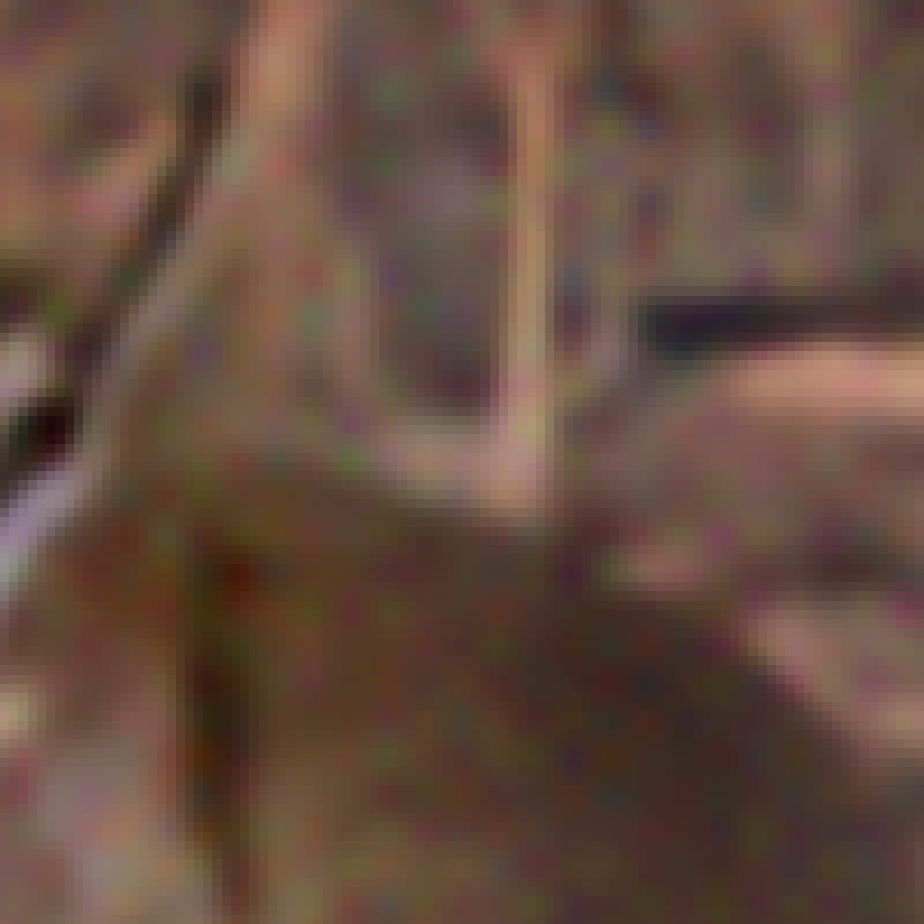}
		& \includegraphics[width=0.115\linewidth]{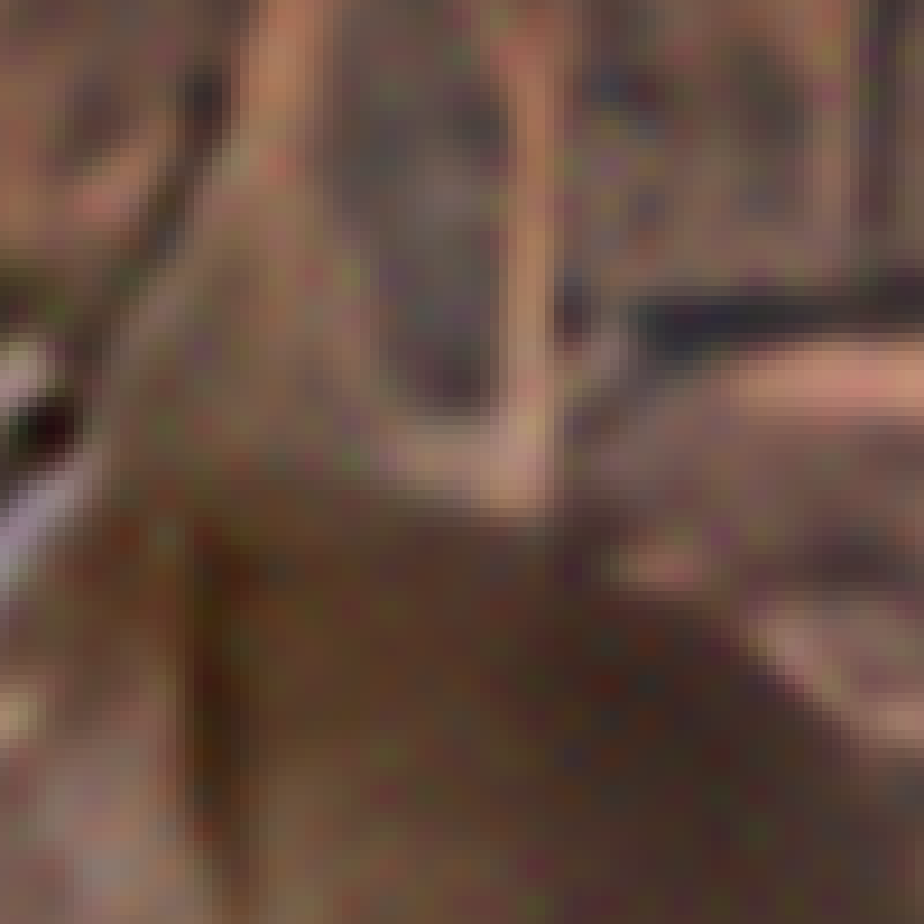}
		& \includegraphics[width=0.115\linewidth]{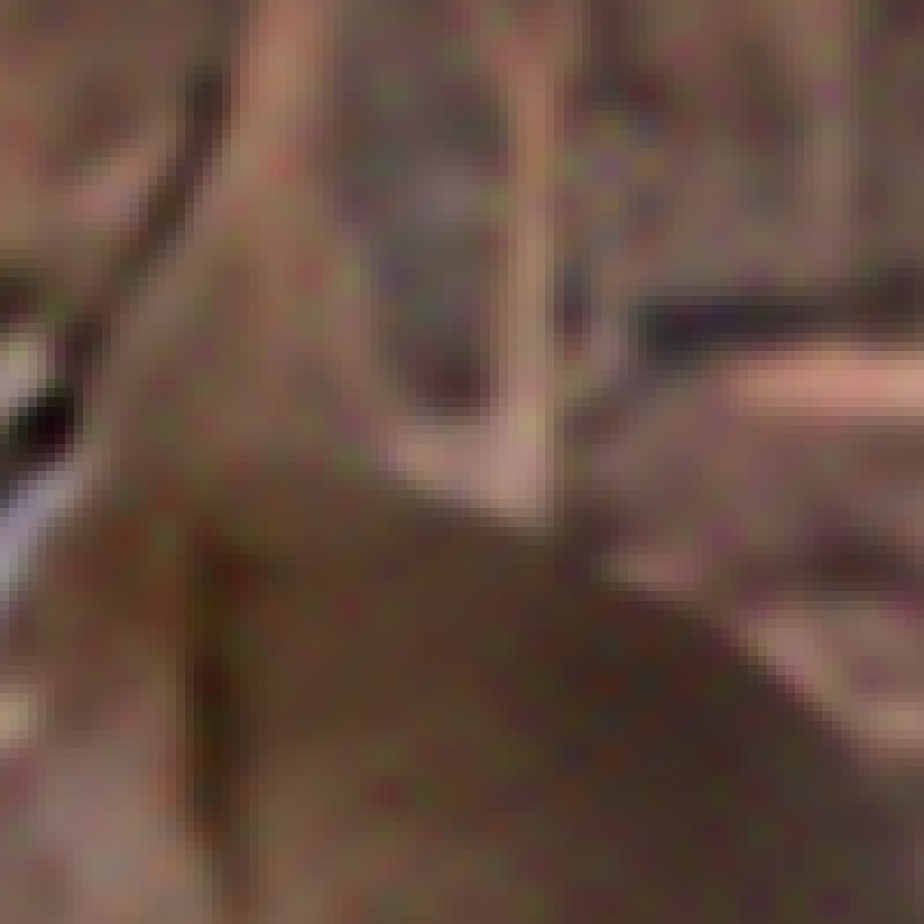}
		& \includegraphics[width=0.115\linewidth]{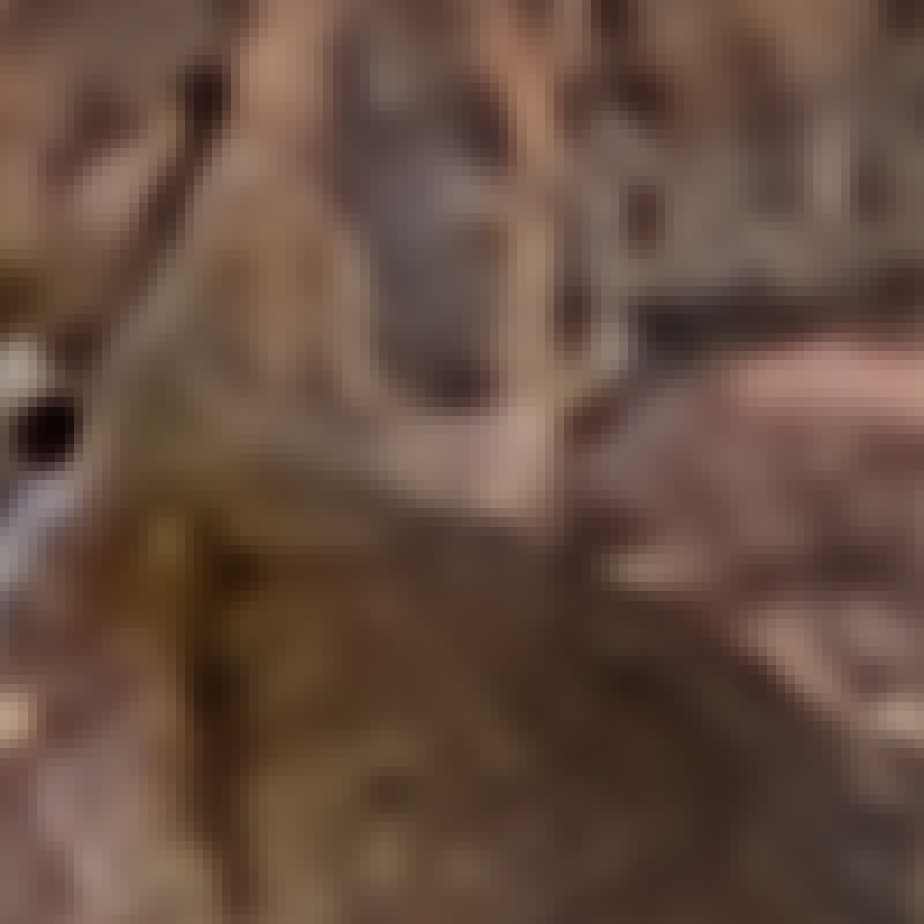}
		\\
		\multirow{-1.5}{*}{\small Reference Image} & LR Patch & +Alignment \cite{wang2024navigating} & +TDF  & +Alignment \cite{wang2024navigating} & +TDF (Ours)
	\end{tabular}
	\begin{tabular}{cccccccc}
		\multirow{-7.5}{*}{\includegraphics[width=0.37\linewidth]{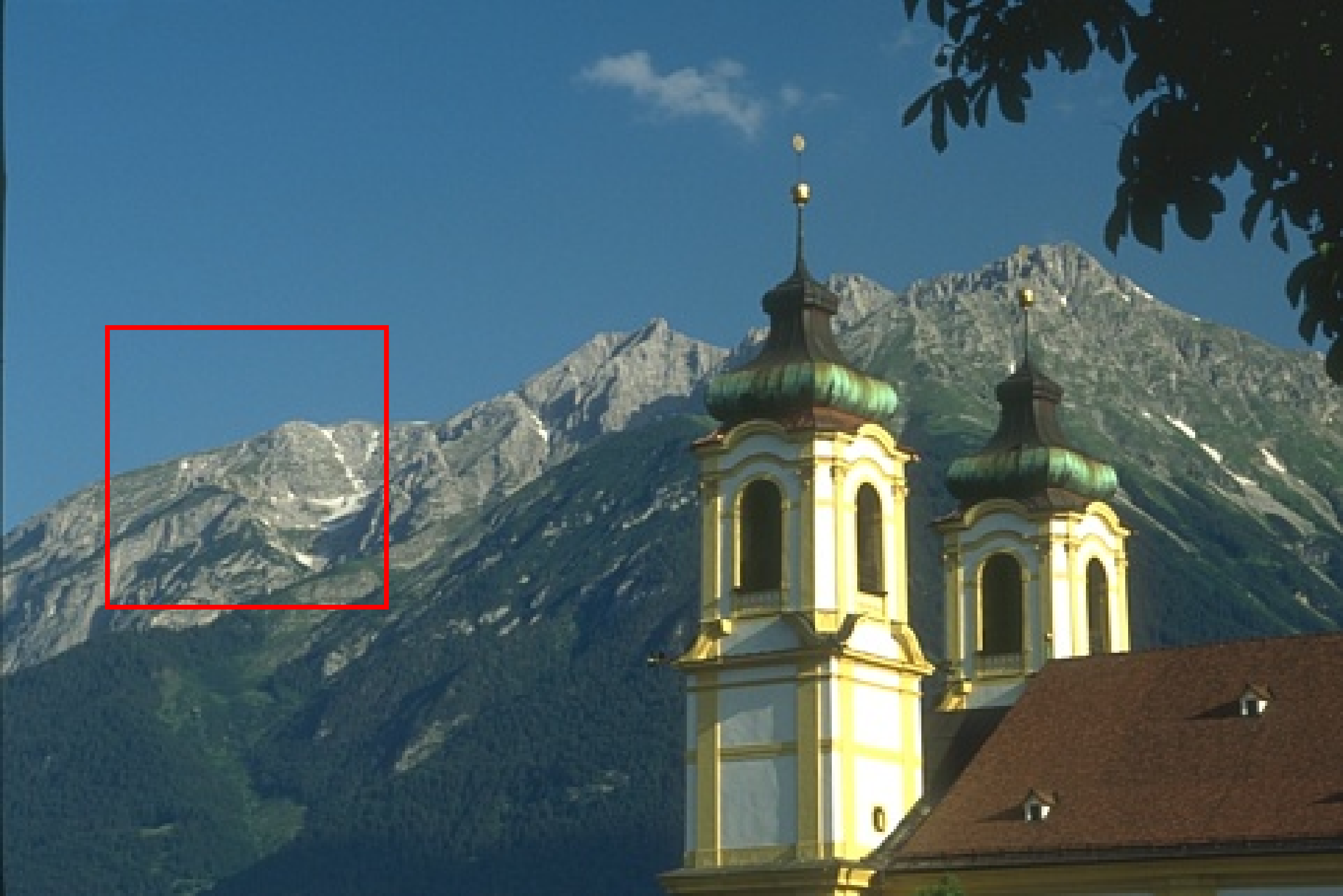}}
		& \includegraphics[width=0.115\linewidth]{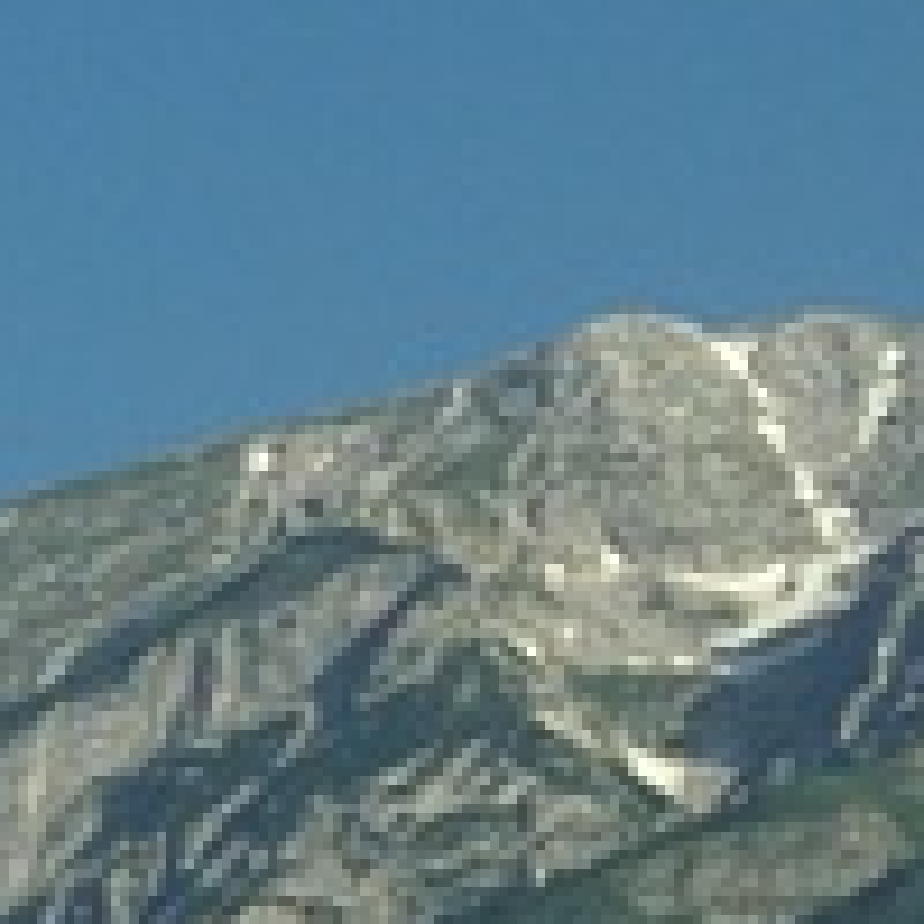}
		& \includegraphics[width=0.115\linewidth]{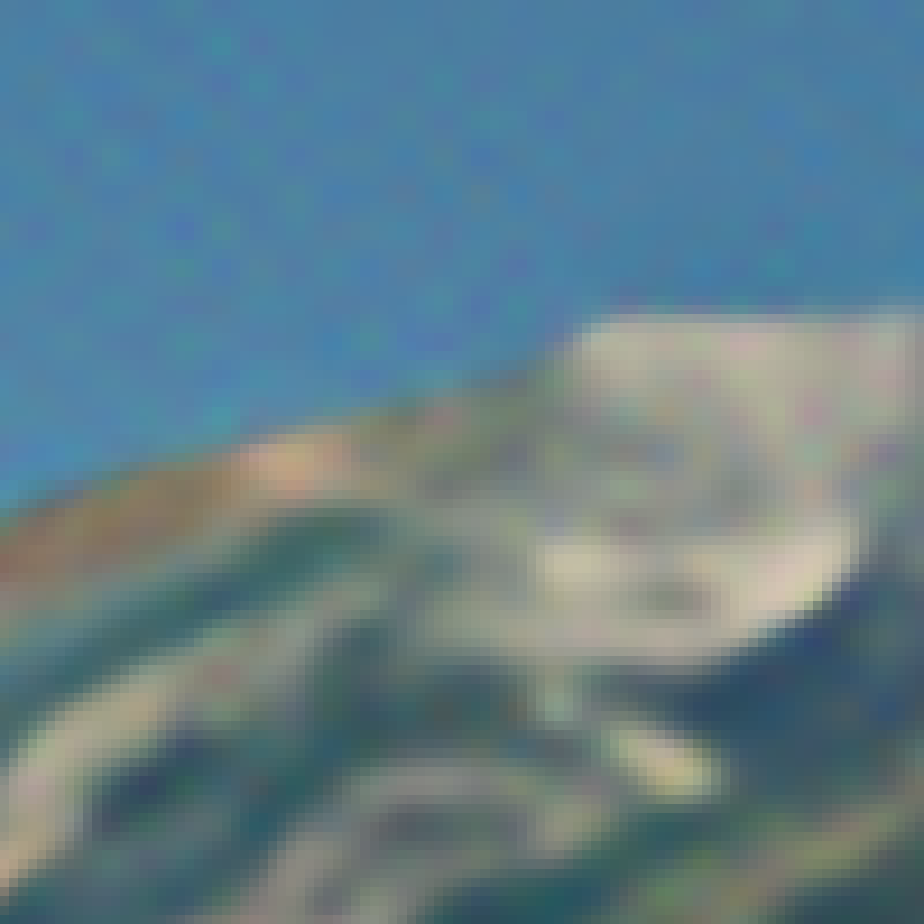}
		& \includegraphics[width=0.115\linewidth]{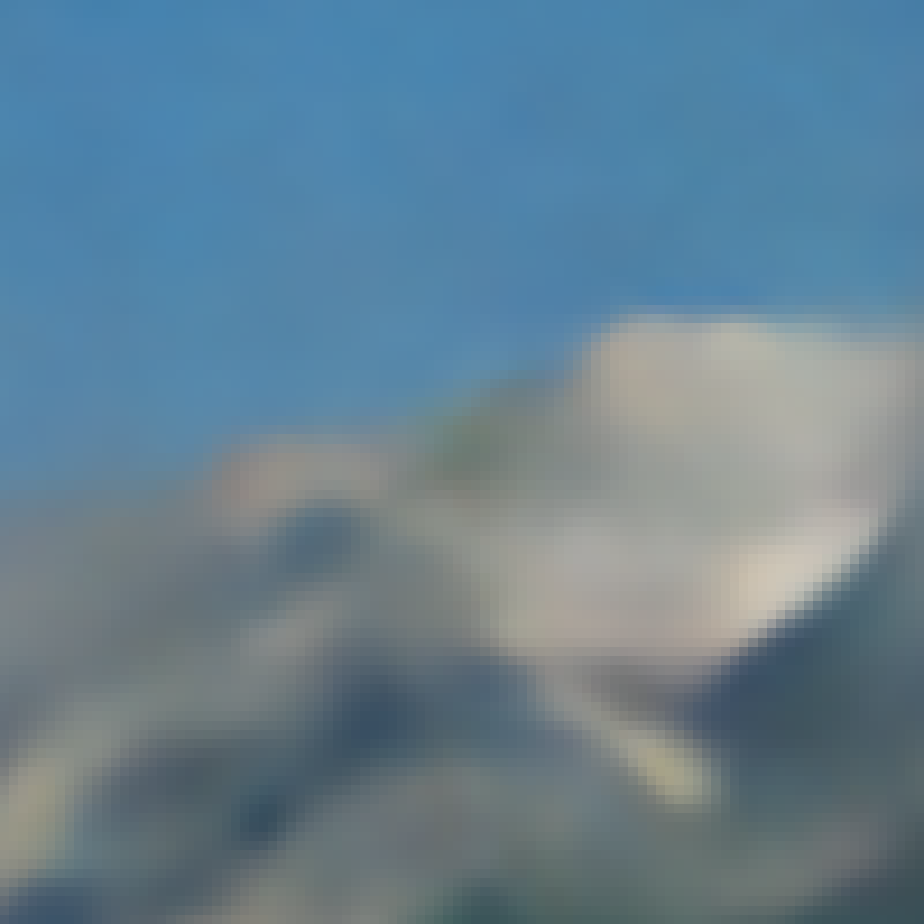}
		& \includegraphics[width=0.115\linewidth]{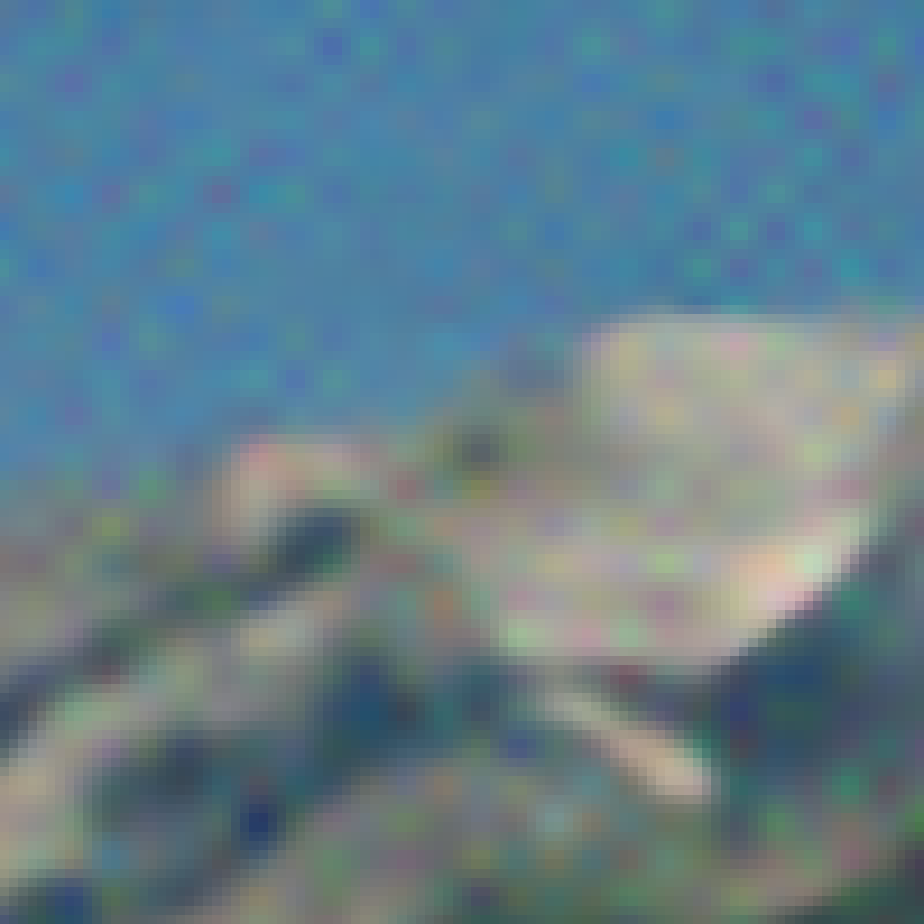}
		& \includegraphics[width=0.115\linewidth]{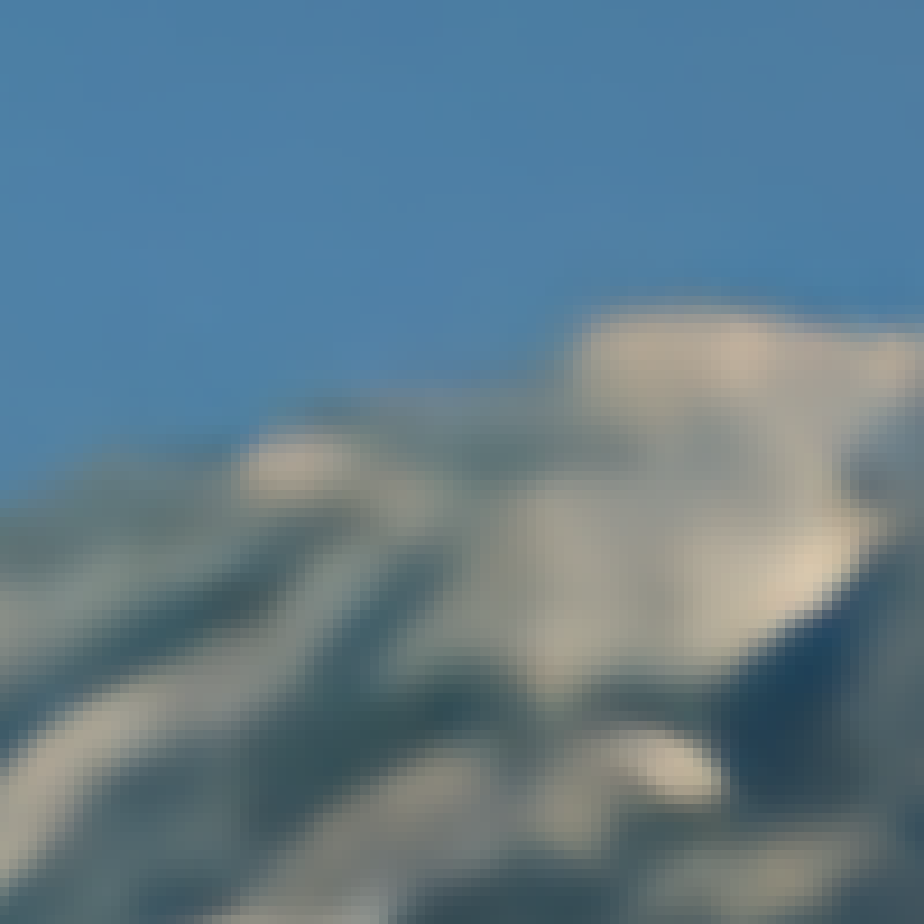}
		\\
		& GT Patch  & SWinIR & +Dropout \cite{kong2022reflash} & HAT & +Dropout \cite{kong2022reflash}
		\\
		& \includegraphics[width=0.115\linewidth]{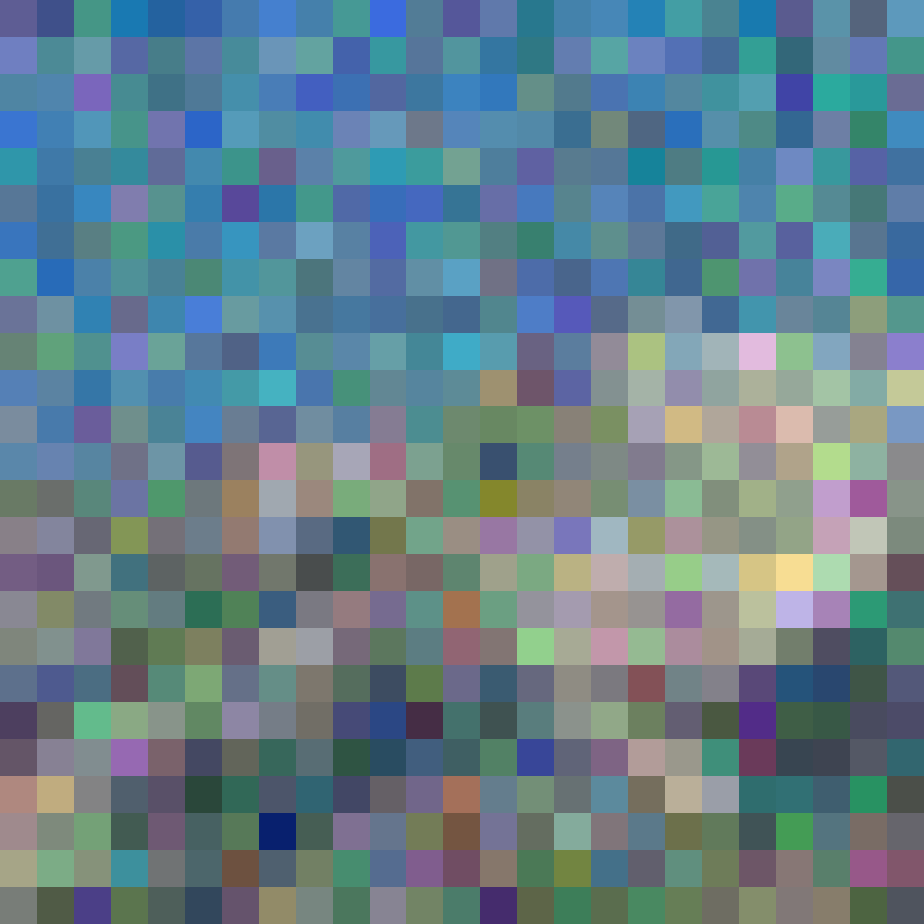}
		& \includegraphics[width=0.115\linewidth]{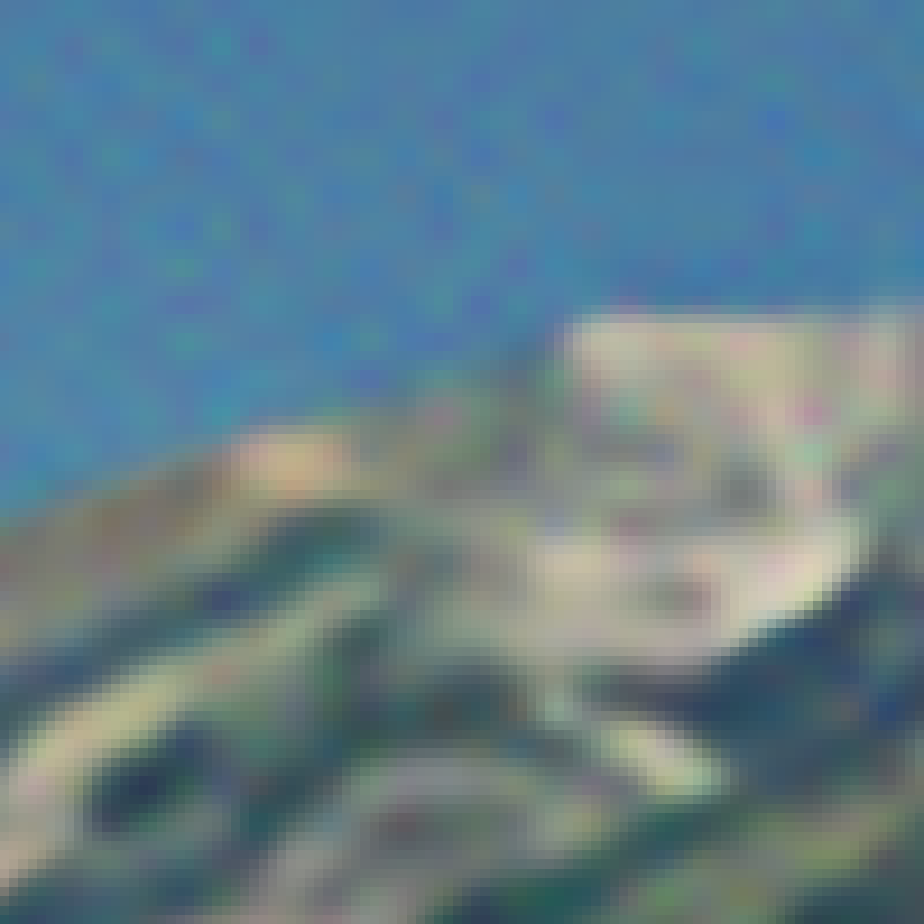}
		& \includegraphics[width=0.115\linewidth]{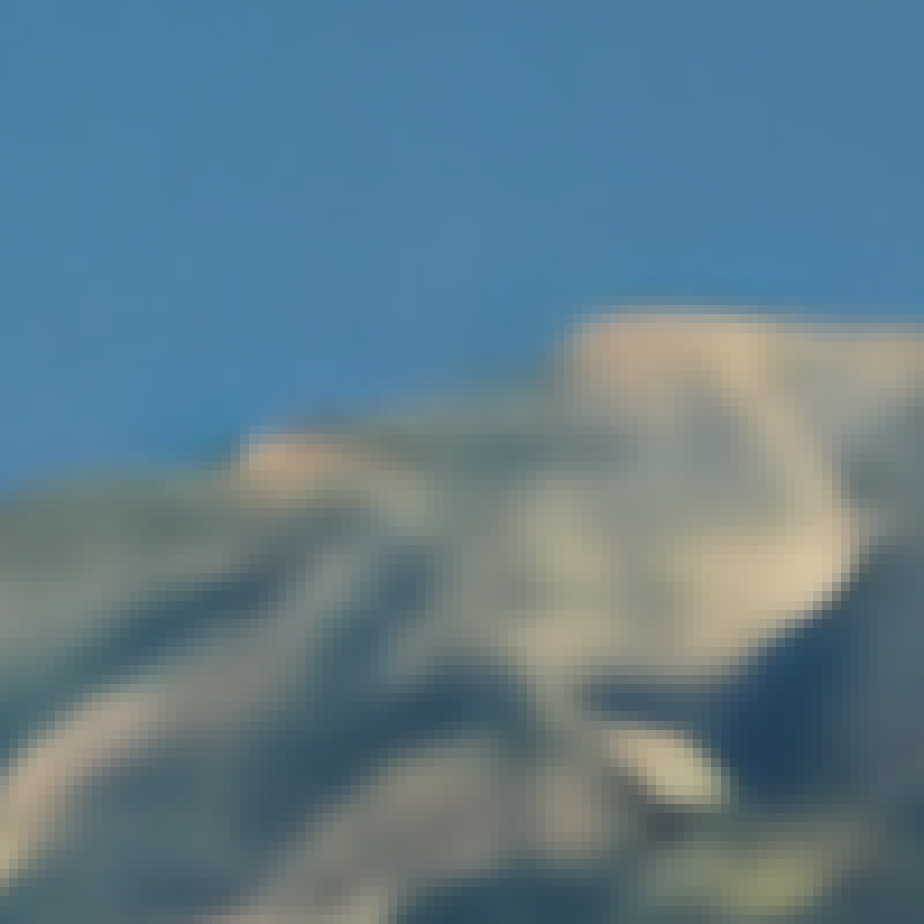}
		& \includegraphics[width=0.115\linewidth]{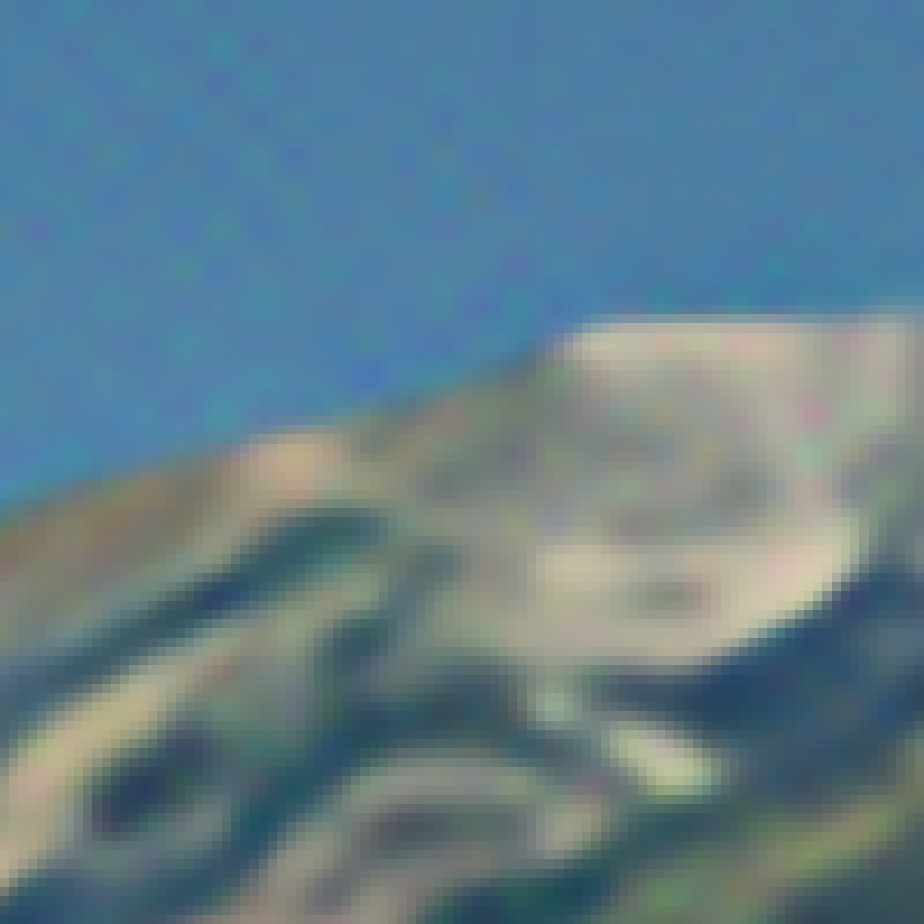}
		& \includegraphics[width=0.115\linewidth]{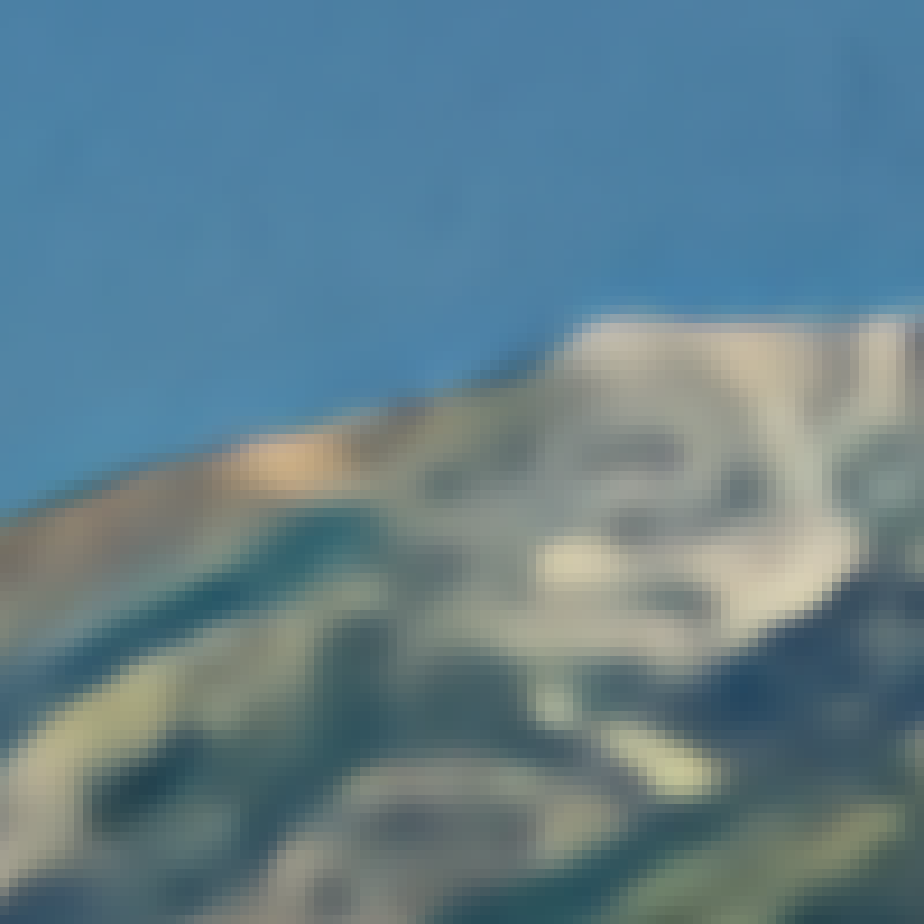}
		\\
		\multirow{-1.5}{*}{\small Reference Image} & LR Patch & +Alignment \cite{wang2024navigating} & +TDF (Ours) & +Alignment \cite{wang2024navigating} & +TDF (Ours)
	\end{tabular}
	\begin{tabular}{cccccccc}
		\multirow{-7.5}{*}{\includegraphics[width=0.37\linewidth]{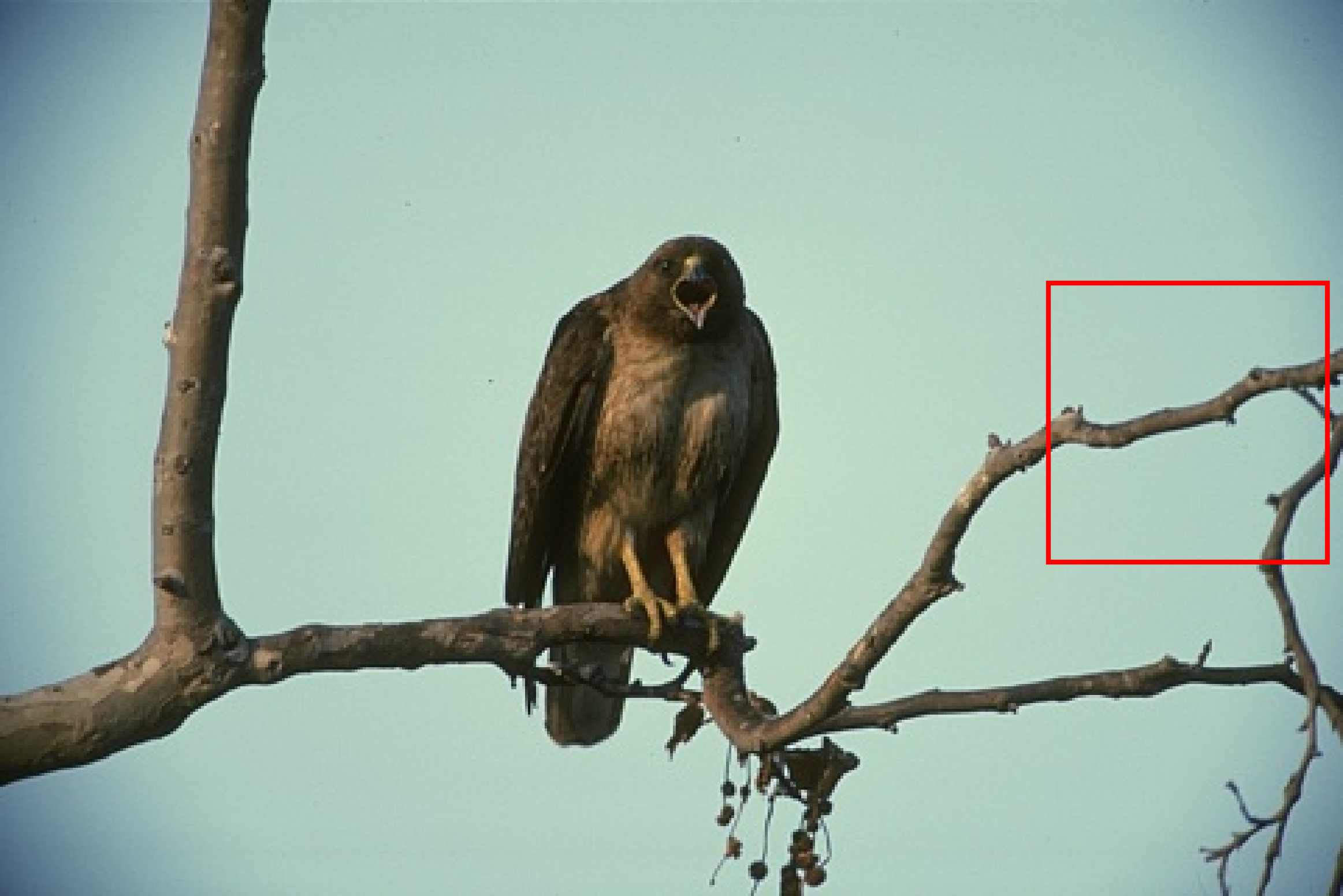}}
		& \includegraphics[width=0.115\linewidth]{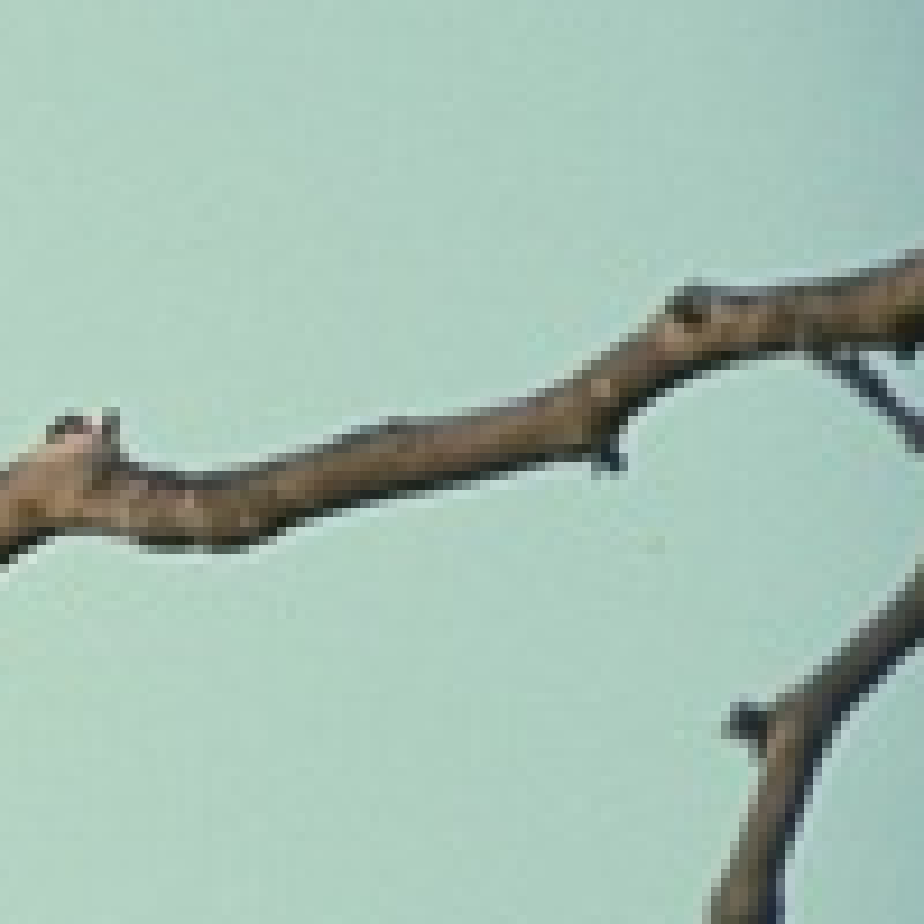}
		& \includegraphics[width=0.115\linewidth]{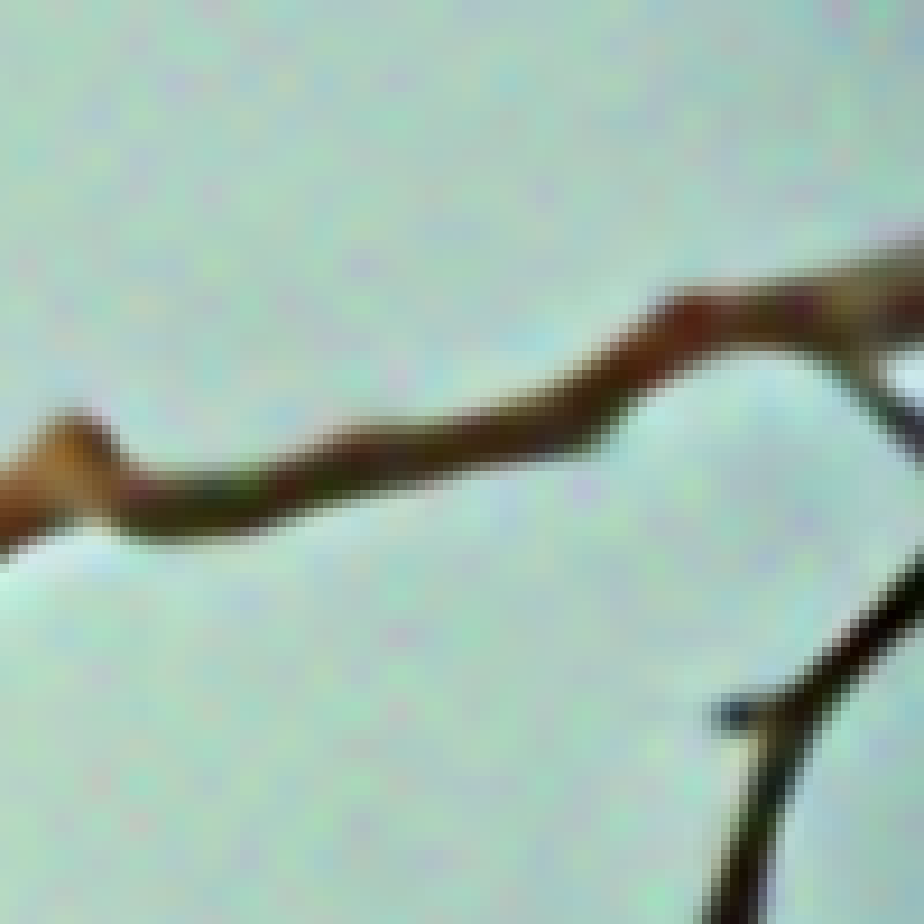}
		& \includegraphics[width=0.115\linewidth]{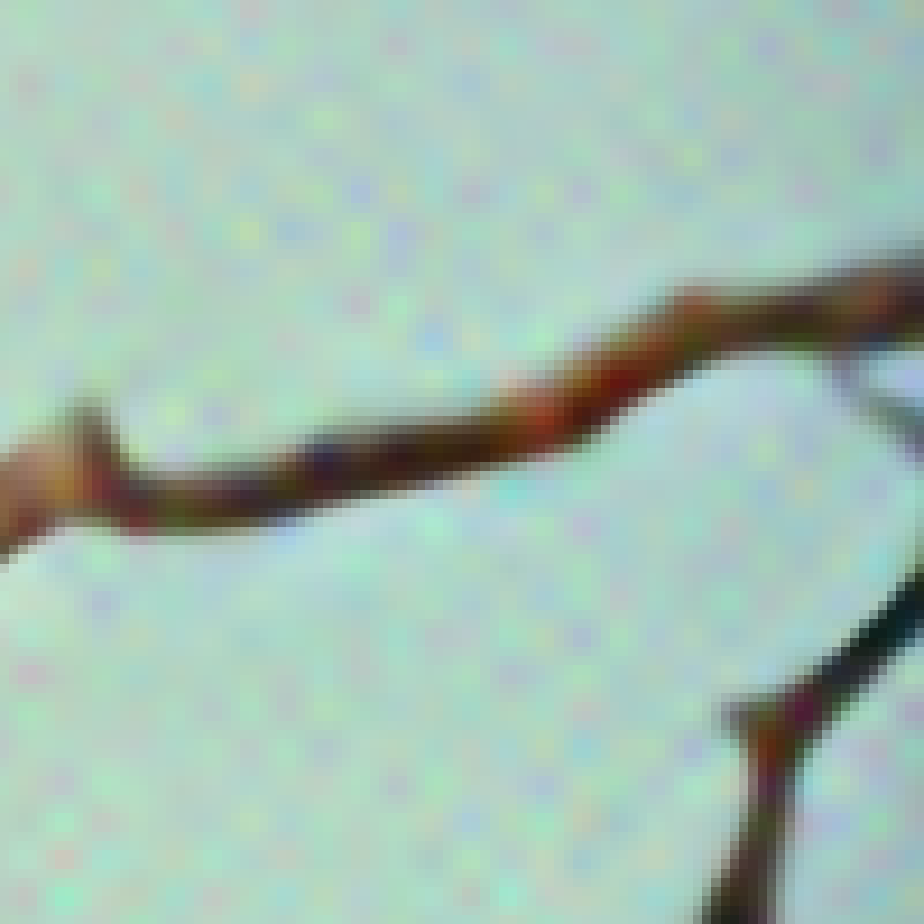}
		& \includegraphics[width=0.115\linewidth]{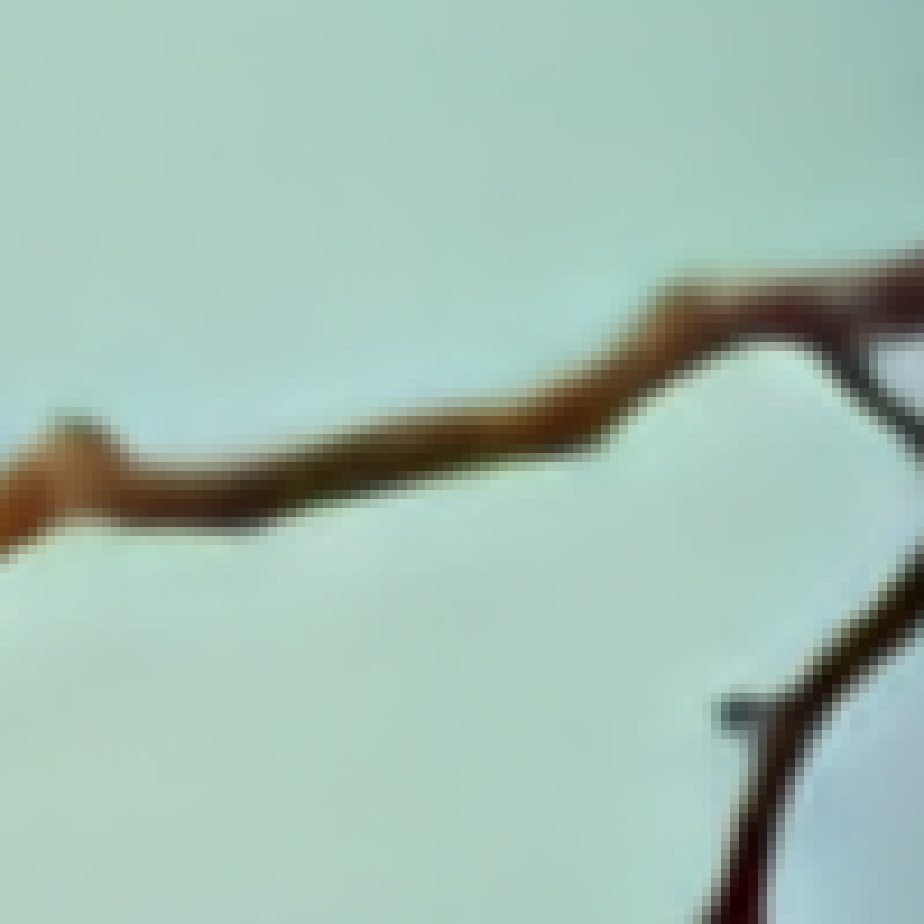}
		& \includegraphics[width=0.115\linewidth]{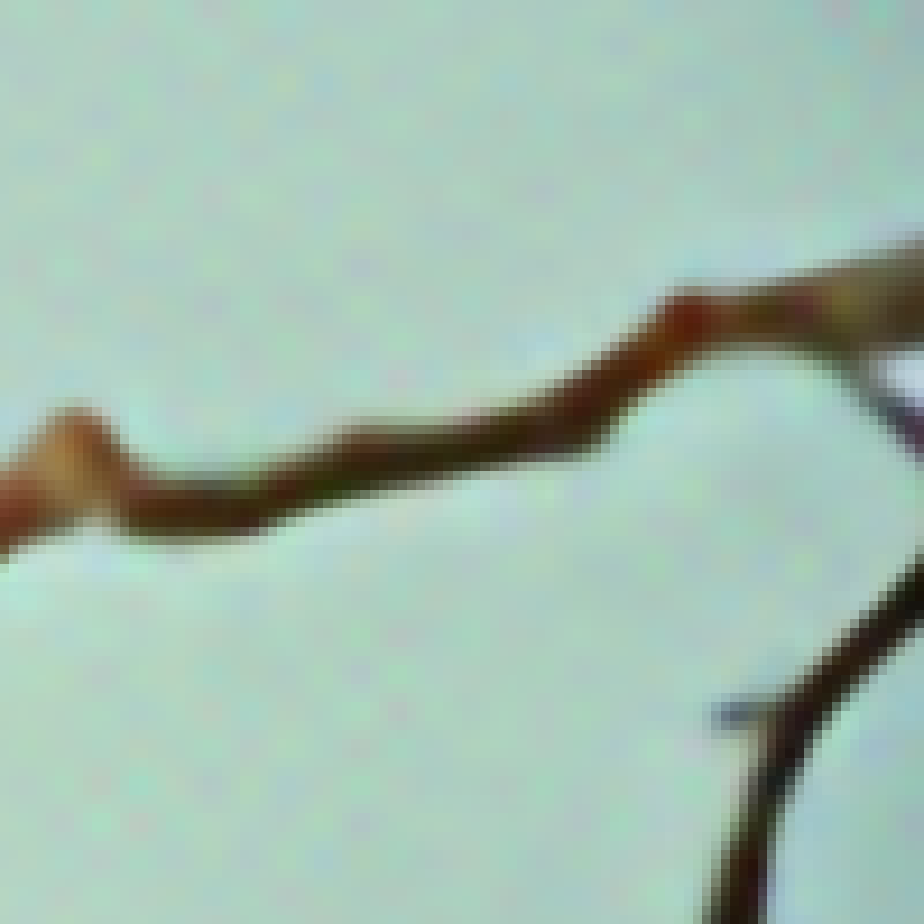}
		\\
		& GT Patch  & SWinIR & +Dropout \cite{kong2022reflash} & HAT & +Dropout \cite{kong2022reflash}
		\\
		& \includegraphics[width=0.115\linewidth]{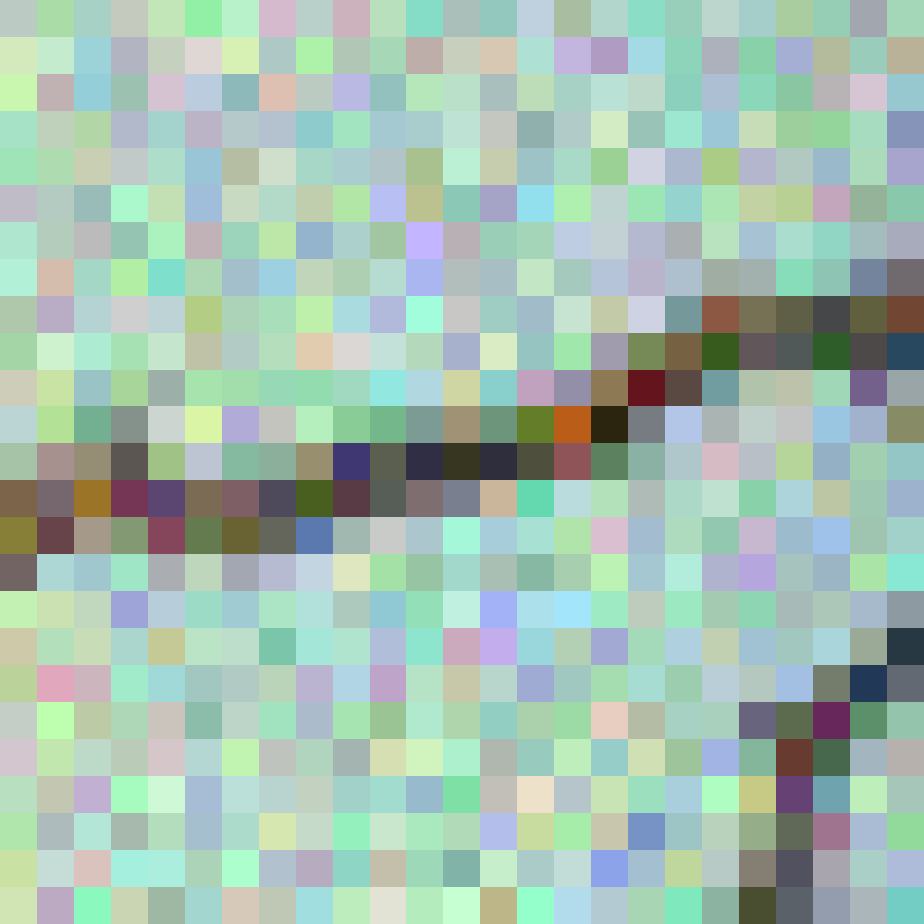}
		& \includegraphics[width=0.115\linewidth]{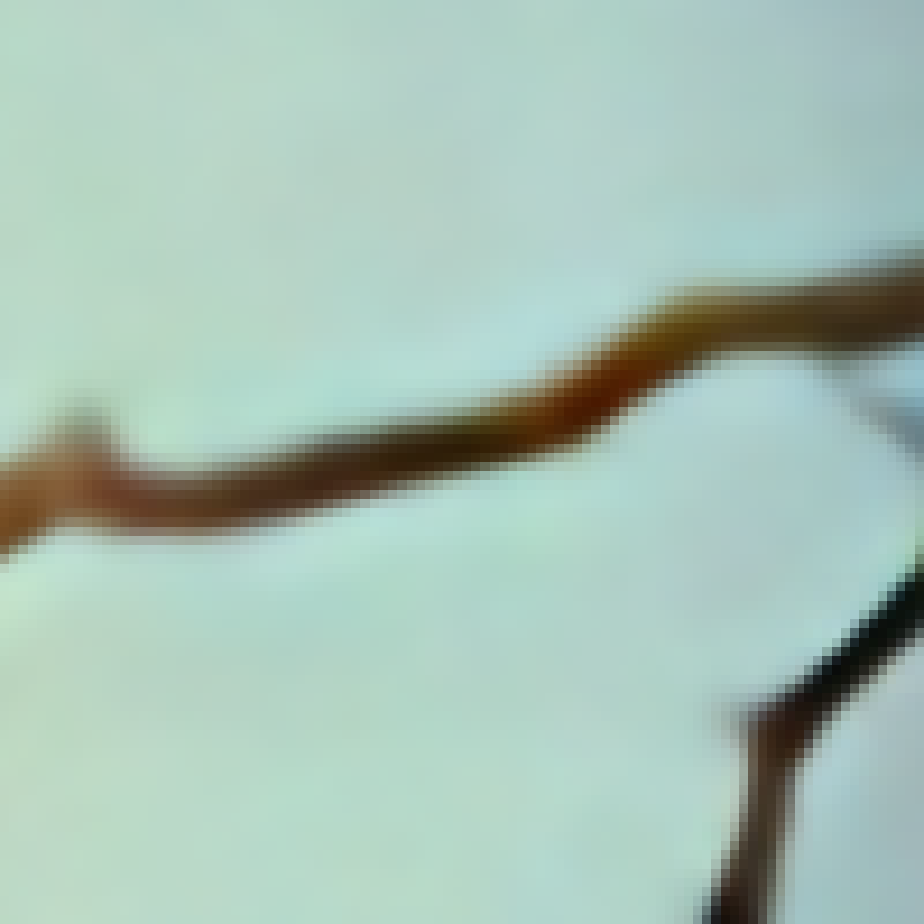}
		& \includegraphics[width=0.115\linewidth]{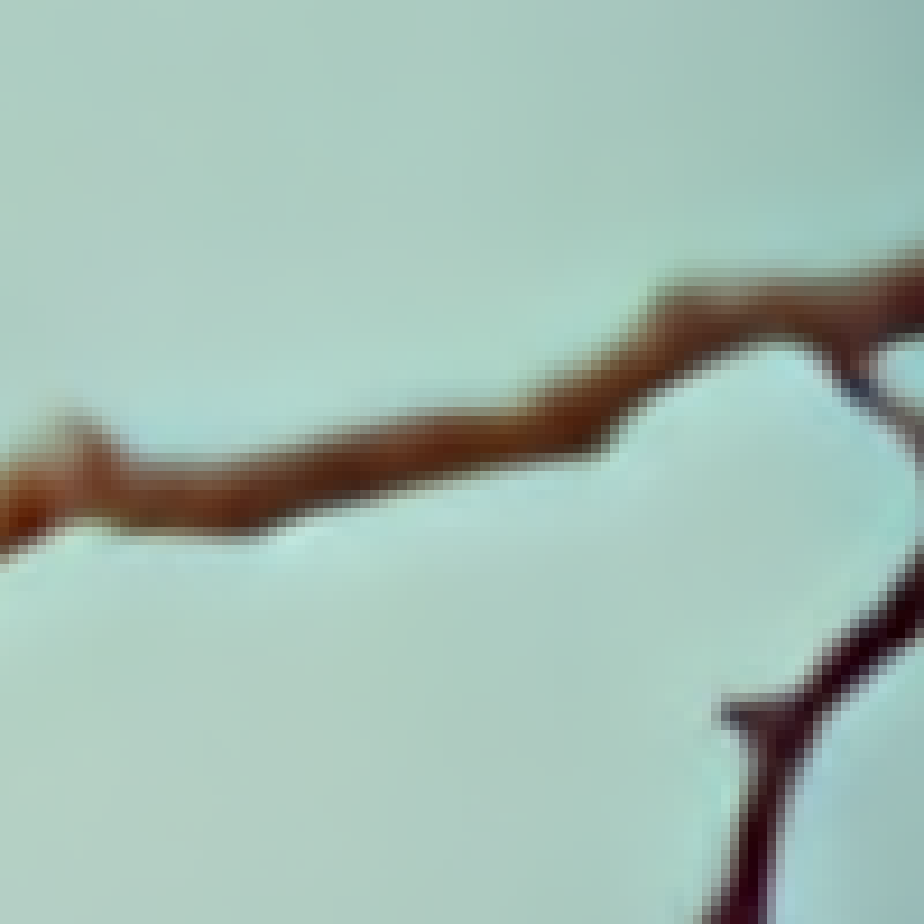}
		& \includegraphics[width=0.115\linewidth]{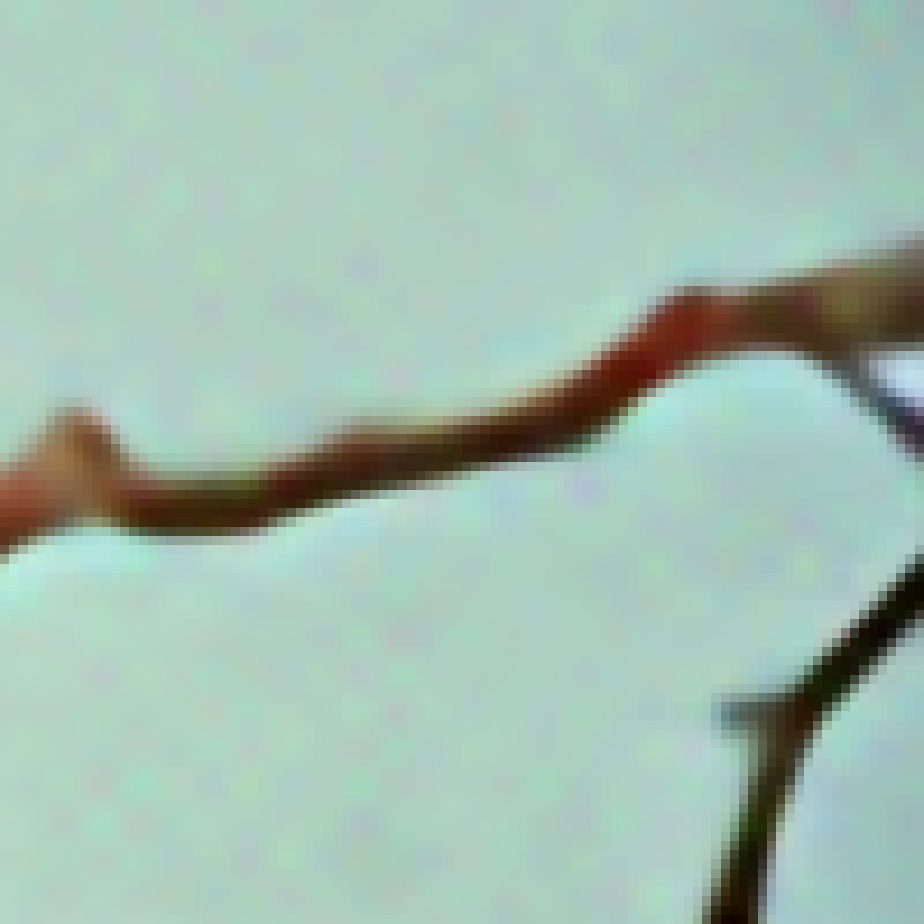}
		& \includegraphics[width=0.115\linewidth]{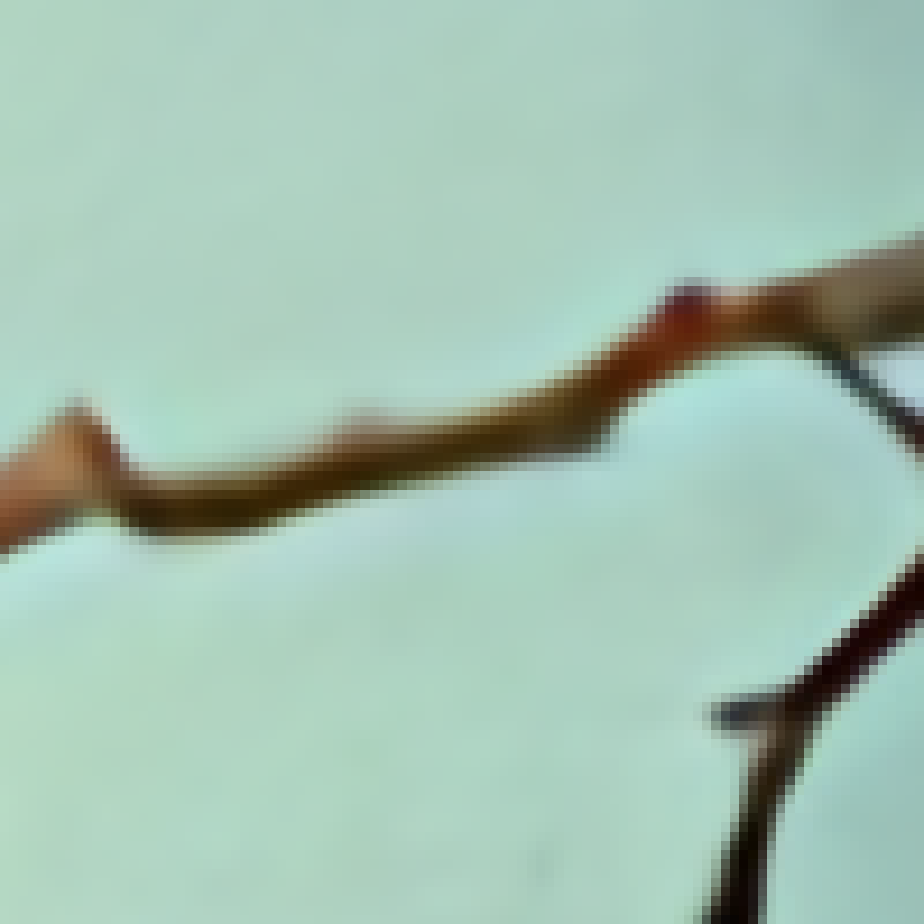}
		\\
		\multirow{-1.5}{*}{\small Reference Image} & LR Patch & +Alignment \cite{wang2024navigating} & +TDF (Ours) & +Alignment \cite{wang2024navigating} & +TDF (Ours)
	\end{tabular}
	\caption{\textbf{Visual comparison of different super-resolution methods on BSD100 dataset with bicubic\_noise20 degradation.}}
	\label{fig:visual_comparison}
\end{figure*}

\begin{enumerate}
	\item \textbf{Clean}: Bicubic downsampling only.
	\item \textbf{Blur}: Gaussian blur followed by bicubic downsampling.
	\item \textbf{Noise}: Additive Gaussian noise followed by bicubic downsampling.
	\item \textbf{JPEG}: JPEG compression followed by bicubic downsampling.
	\item \textbf{Blur+Noise}: Sequential application of Gaussian blur and additive Gaussian noise, followed by bicubic downsampling.
	\item \textbf{Blur+JPEG}: Sequential application of Gaussian blur and JPEG compression, followed by bicubic downsampling.
	\item \textbf{Noise+JPEG}: Sequential application of additive Gaussian noise and JPEG compression, followed by bicubic downsampling.
	\item \textbf{Blur+Noise+JPEG}: Combination of all three degradations.
\end{enumerate}
Formally, for a clean high-resolution image $\mathbf{x}$, each degraded low-resolution observation $\mathbf{y}$ is synthesized as:
\begin{equation}
	\mathbf{y} = \mathcal{D} \circ \mathcal{C} \circ \mathcal{N} \circ \mathcal{B}(\mathbf{x}),
\end{equation}
where $\mathcal{B}$ denotes blurring, $\mathcal{N}$ denotes noise injection, $\mathcal{C}$ denotes JPEG compression, and $\mathcal{D}$ denotes downsampling. Depending on the configuration, certain operators are replaced by the identity map.

\noindent\textbf{Noise Detection Accuracy in Unseen Degradations.} The noise detection module demonstrates robust discriminative capability across various benchmark datasets and degradation types, as illustrated in Figure \ref{fig:noisedetectionaccuracyheatmap}. Particularly noteworthy is the perfect detection accuracy (1.0) observed for noise degradation across all evaluated datasets, which validates our hypothesis regarding the distinctive spectral characteristics of noise-induced corruption. While the module maintains high accuracy for blur degradation ($\geq 0.99$), we observe marginally lower accuracy for JPEG artifacts in BSD100 (0.84) and bicubic degradation in Urban100 (0.81). This performance disparity aligns with our frequency-domain analysis, which revealed that noise exhibits uniform spectral distribution, making it more distinctively identifiable compared to other degradations that manifest primarily in specific frequency bands. The module's consistent performance across diverse datasets (Set5, Set14, BSD100, Urban100, Manga109) further substantiates the generalizability of our approach to real-world super-resolution scenarios involving complex degradation patterns.

\begin{table}[!t]
	\centering
	\caption{\textbf{Noise detection accuracy before and after denoising.} }
	\label{tab:noise_reversed}
	\renewcommand{\arraystretch}{1.2}
	\resizebox{1\linewidth}{!}{
		\begin{tabular}{lcccc}
			\hline
			\multirow{2}{*}{Dataset} & SRResNet & RRDB & HAT & SwinIR \\
			\cline{2-5}
			& Before $\rightarrow$ After & Before $\rightarrow$ After & Before $\rightarrow$ After & Before $\rightarrow$ After \\
			\hline
			Set5 & 1.00 $\rightarrow$ 0.00 & 1.00 $\rightarrow$ 0.01 & 0.99 $\rightarrow$ 0.00 & 0.99 $\rightarrow$ 0.00 \\
			\hline
			Set14 & 1.00 $\rightarrow$ 0.00 & 1.00 $\rightarrow$ 0.00 & 0.98 $\rightarrow$ 0.01 & 0.99 $\rightarrow$ 0.01 \\
			\hline
			BSD100 & 1.00 $\rightarrow$ 0.00 & 0.99 $\rightarrow$ 0.01 & 0.98 $\rightarrow$ 0.01 & 0.98 $\rightarrow$ 0.01 \\
			\hline
			Urban100 & 0.98 $\rightarrow$ 0.02 & 0.97 $\rightarrow$ 0.03 & 0.96 $\rightarrow$ 0.03 & 0.95 $\rightarrow$ 0.04 \\
			\hline
			Manga109 & 0.99 $\rightarrow$ 0.01 & 0.98 $\rightarrow$ 0.02 & 0.97 $\rightarrow$ 0.02 & 0.97 $\rightarrow$ 0.02 \\
			\hline
		\end{tabular}
	}
\end{table}

\noindent\textbf{Noise Detection Accuracy Before and After Denoising.} Table \ref{tab:noise_reversed} demonstrates the efficacy of our feature denoising framework by comparing noise detection rates before and after applying the denoising module across multiple super-resolution architectures and benchmark datasets. The results reveal a remarkable transition in detection rates, with pre-denoising values approaching perfect classification (0.95-1.00) across all model-dataset combinations, indicating consistent identification of noise-corrupted features. After applying our denoising module, detection rates plummet dramatically to near-zero values (0.00-0.04), providing compelling evidence that our approach effectively eliminates noise characteristics from the feature representations. This pronounced before-after contrast is particularly evident in the SRResNet architecture, where the Set5, Set14, and BSD100 datasets exhibit a complete reversal from 1.00 to 0.00 detection rates. The consistently low post-denoising detection rates across architectures (SRResNet, RRDB, HAT, and SwinIR) and datasets substantiate the architecture-agnostic nature of our method. The marginally higher post-denoising rates observed in Urban100 (0.02-0.04) likely reflect the dataset's complex structural patterns, which present greater challenges for discriminating between residual noise and high-frequency content details. These quantitative results corroborate our qualitative observations and theoretical analysis, confirming that our approach effectively addresses the noise overfitting phenomenon by selectively suppressing noise-related features while preserving content-relevant information.

\begin{figure}
	\centering
	\subfloat[Input]{\includegraphics[width=0.33\linewidth]{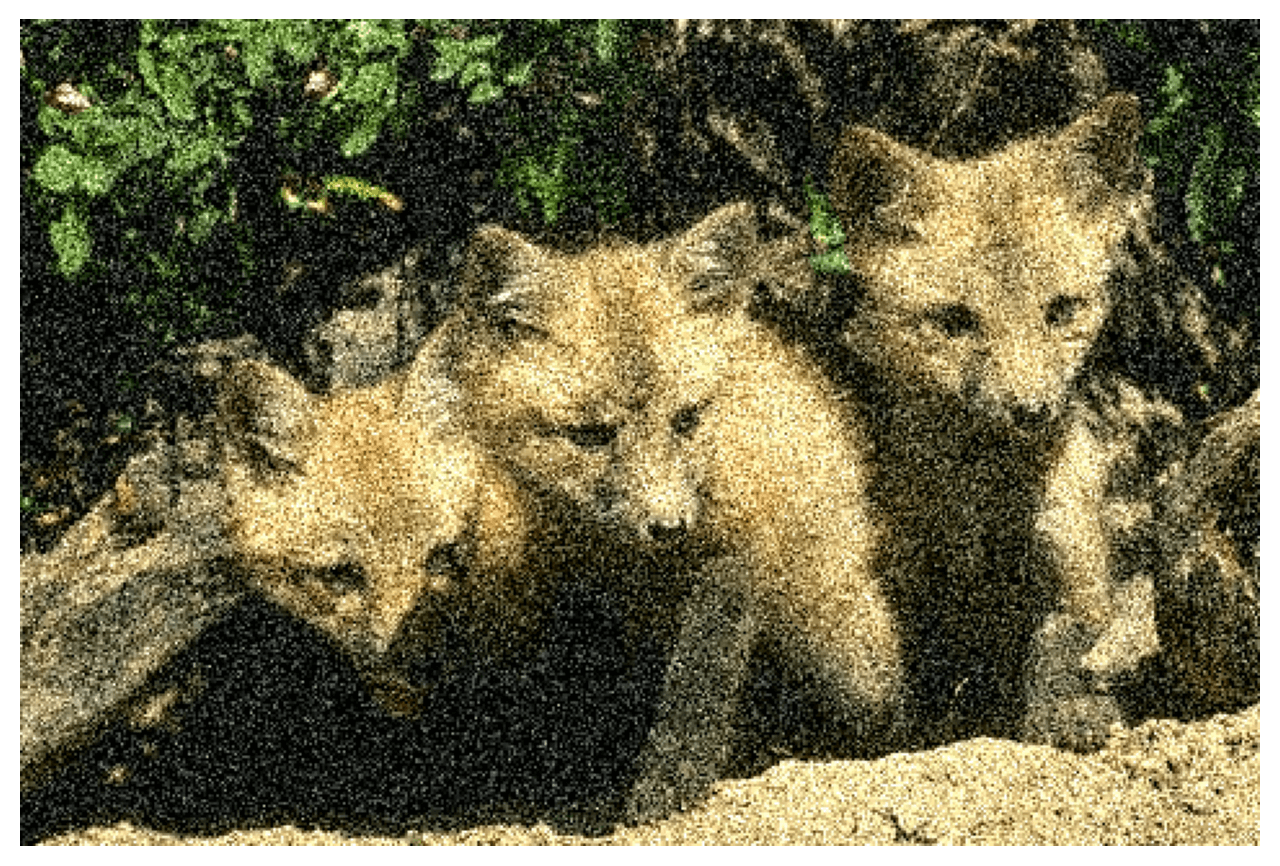}}
	\hfill
	\subfloat[SRresNet]{\includegraphics[width=0.33\linewidth]{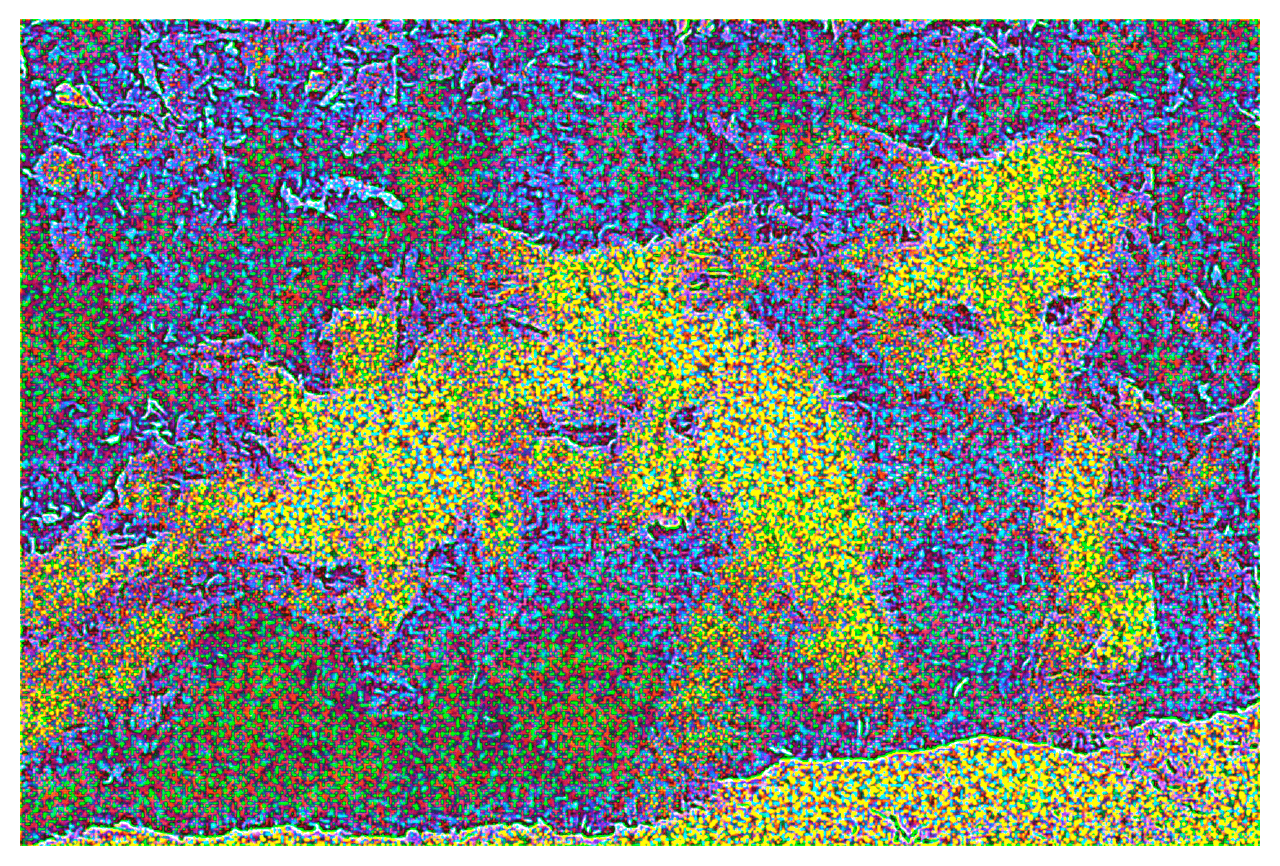}}
	\hfill
	\subfloat[+TFD]{\includegraphics[width=0.33\linewidth]{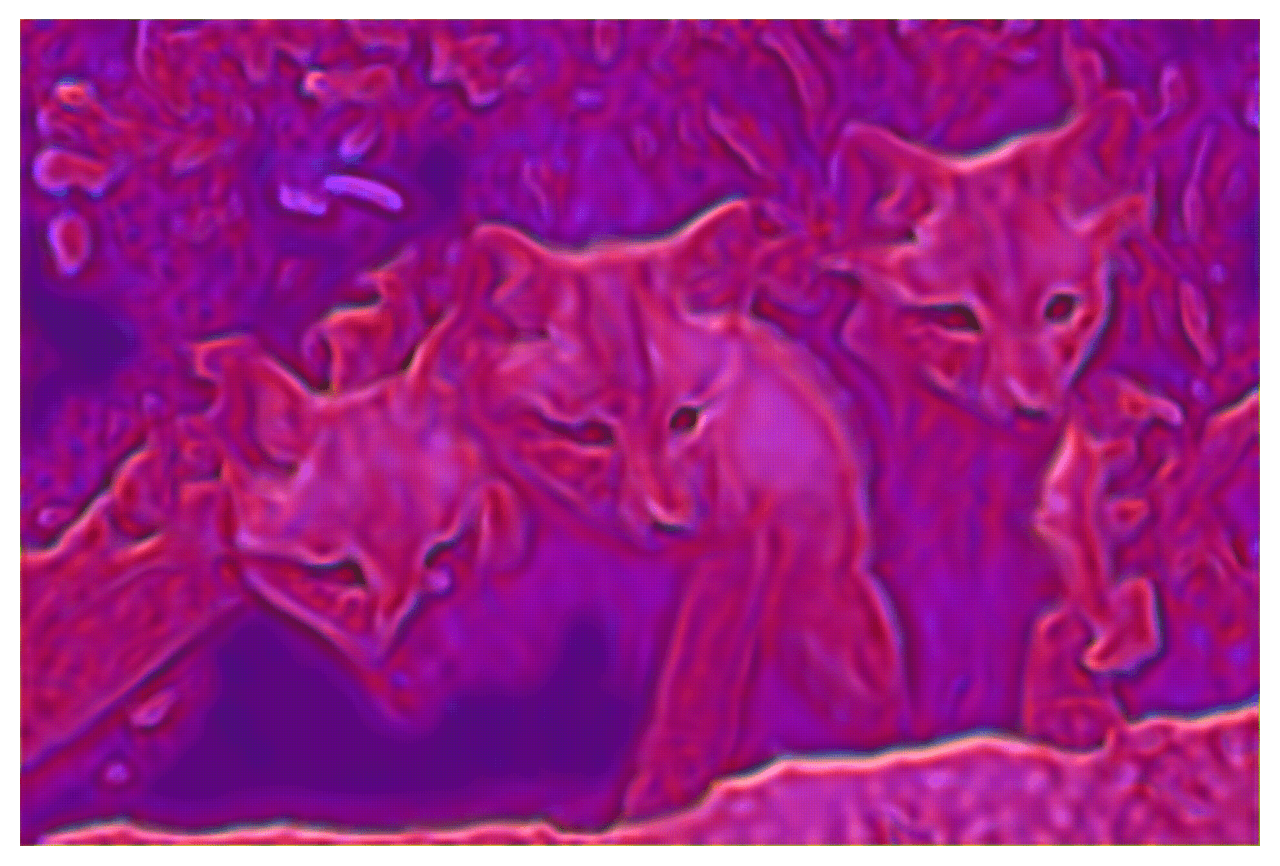}}
	
	\subfloat[Input]{\includegraphics[width=0.33\linewidth]{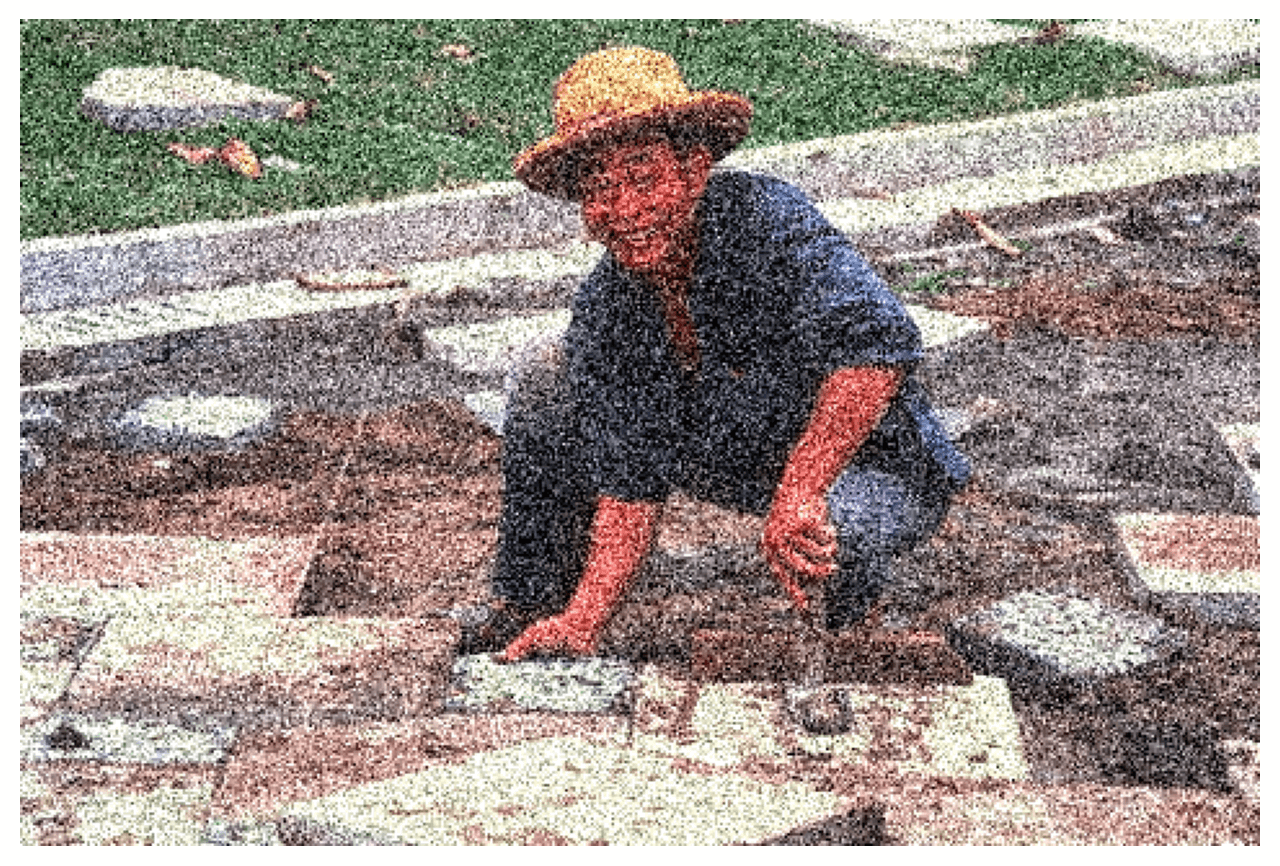}}
	\hfill
	\subfloat[SRresNet]{\includegraphics[width=0.33\linewidth]{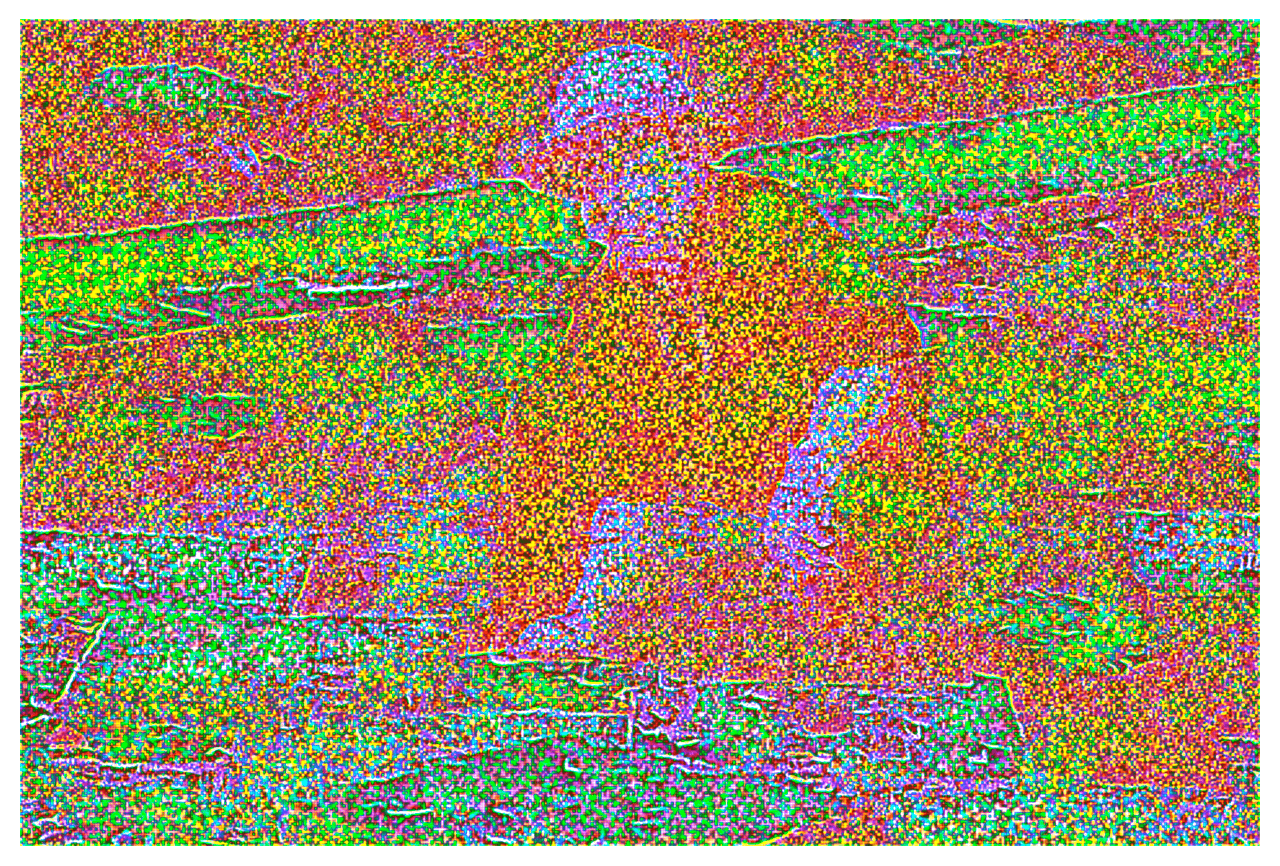}}
	\hfill
	\subfloat[+TFD]{\includegraphics[width=0.33\linewidth]{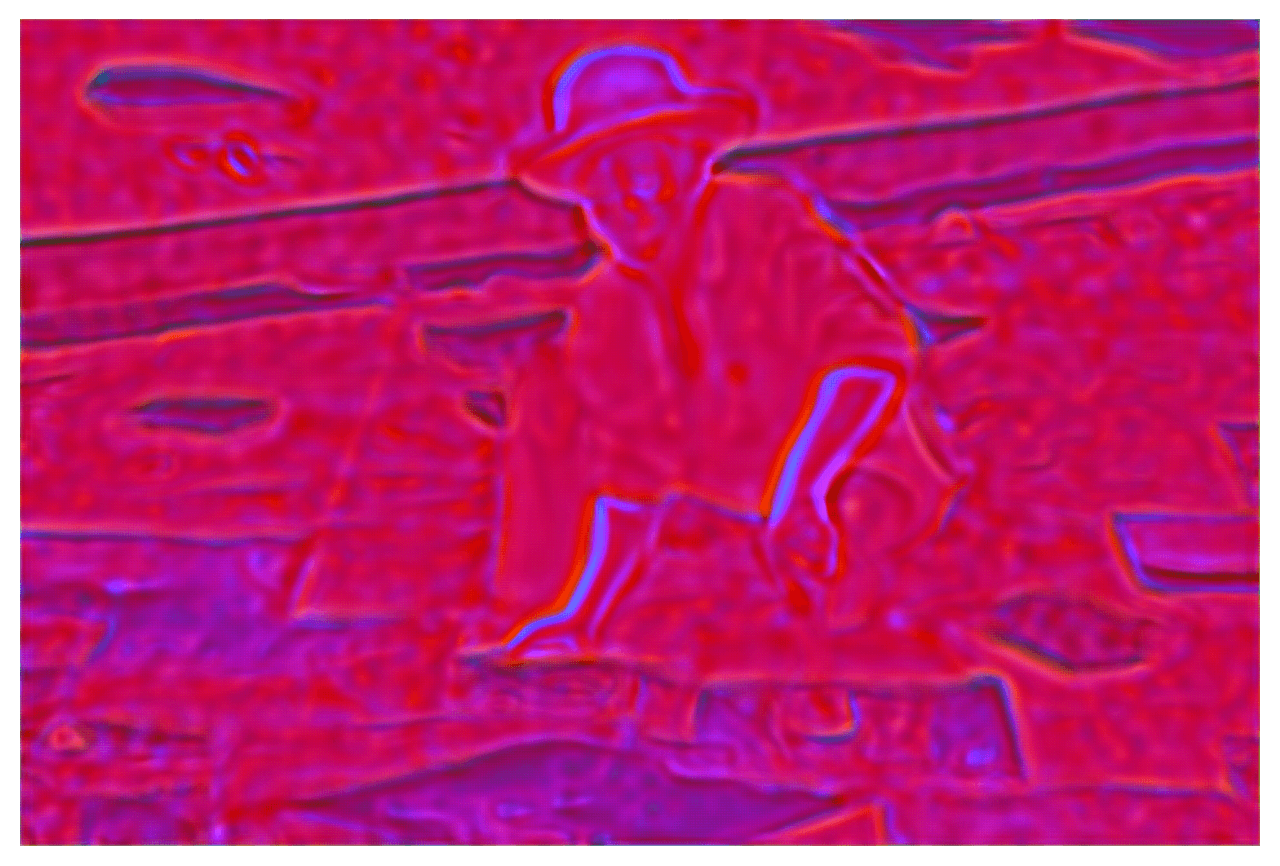}}
	
	\subfloat[Input]{\includegraphics[width=0.33\linewidth]{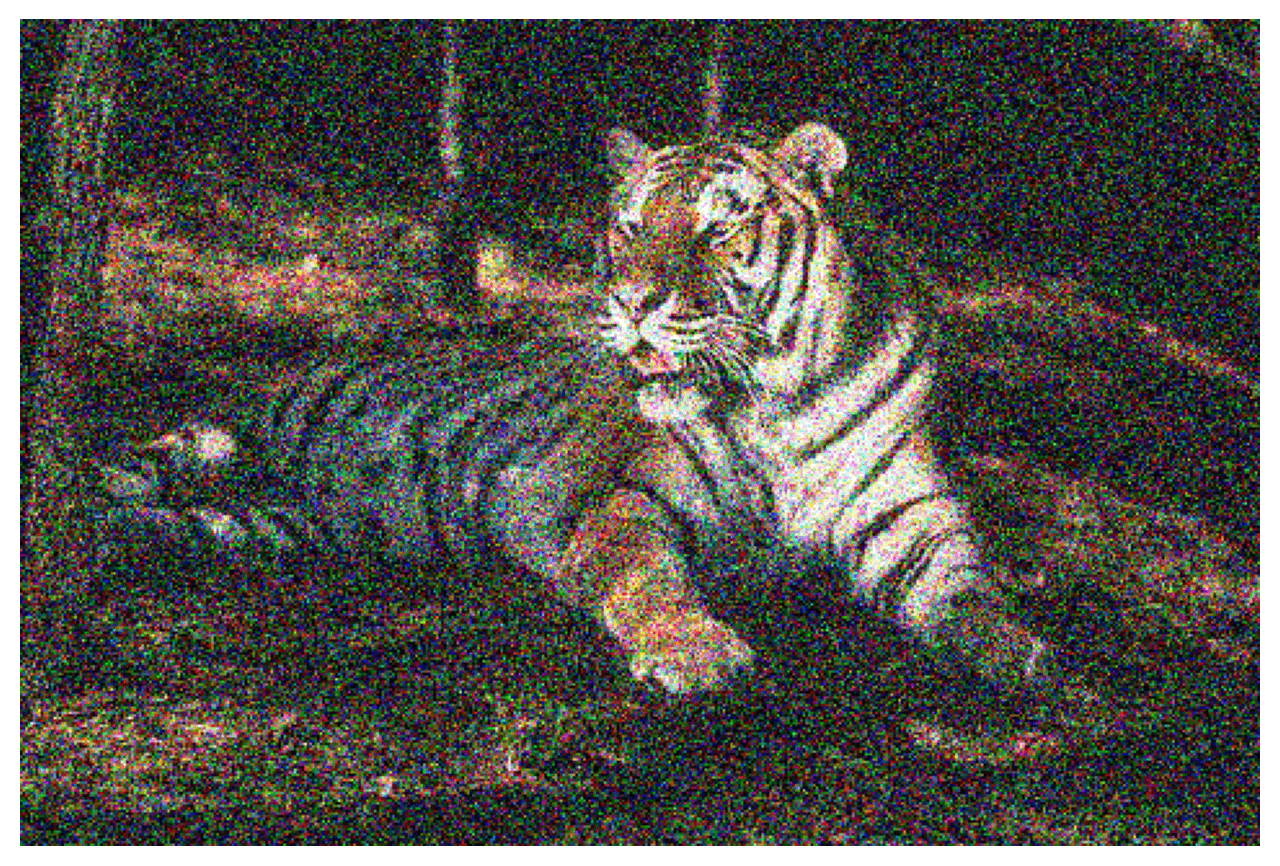}}
	\hfill
	\subfloat[SRresNet]{\includegraphics[width=0.33\linewidth]{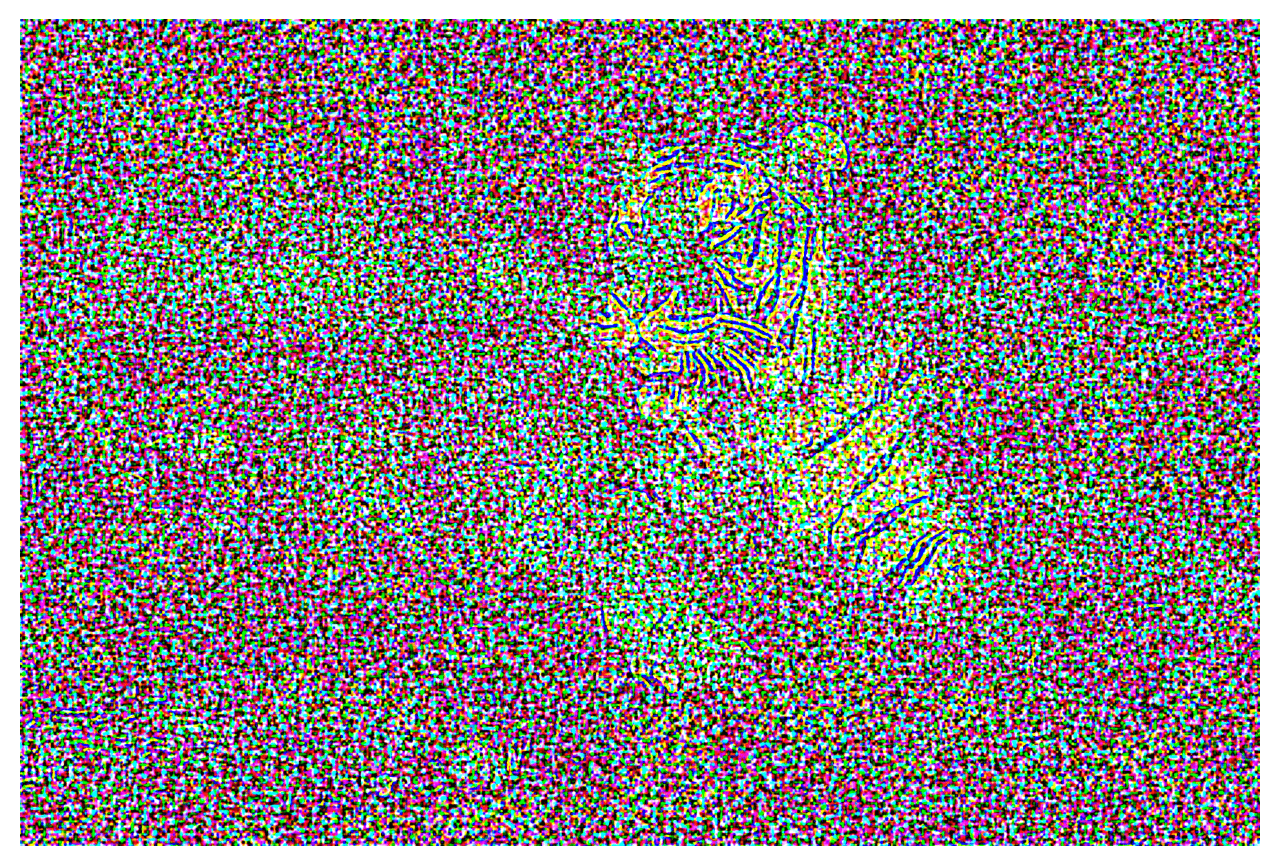}}
	\hfill
	\subfloat[+TFD]{\includegraphics[width=0.33\linewidth]{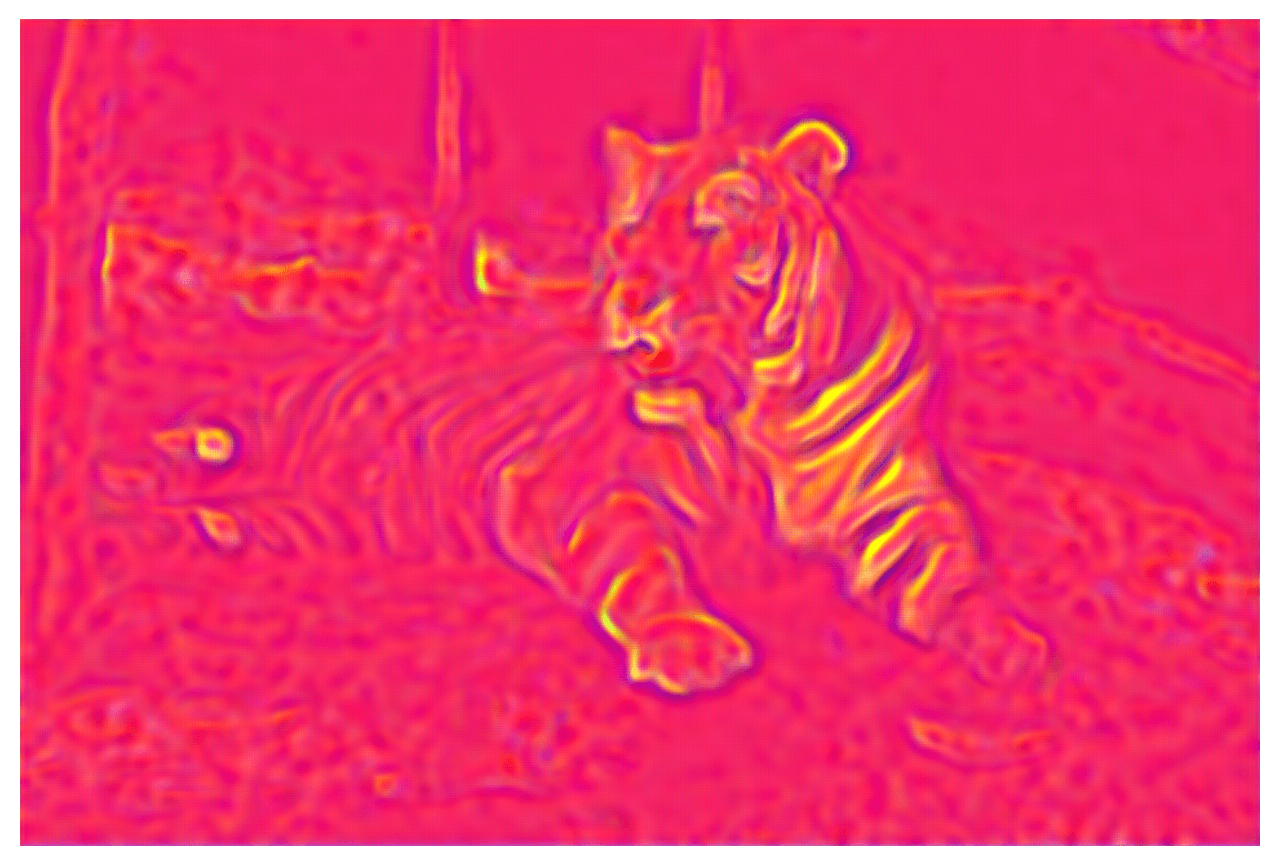}}
	
	\subfloat[Input]{\includegraphics[width=0.33\linewidth]{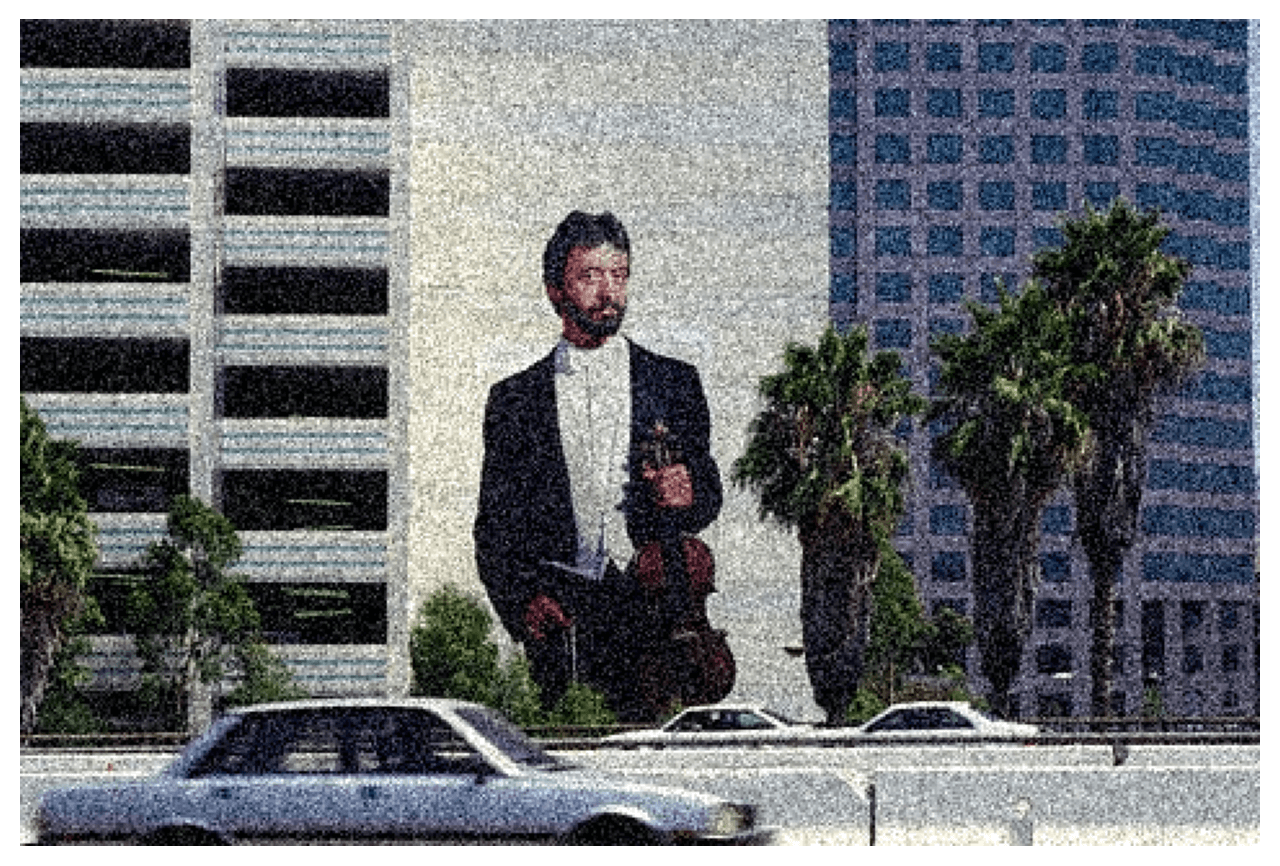}}
	\hfill
	\subfloat[SRresNet]{\includegraphics[width=0.33\linewidth]{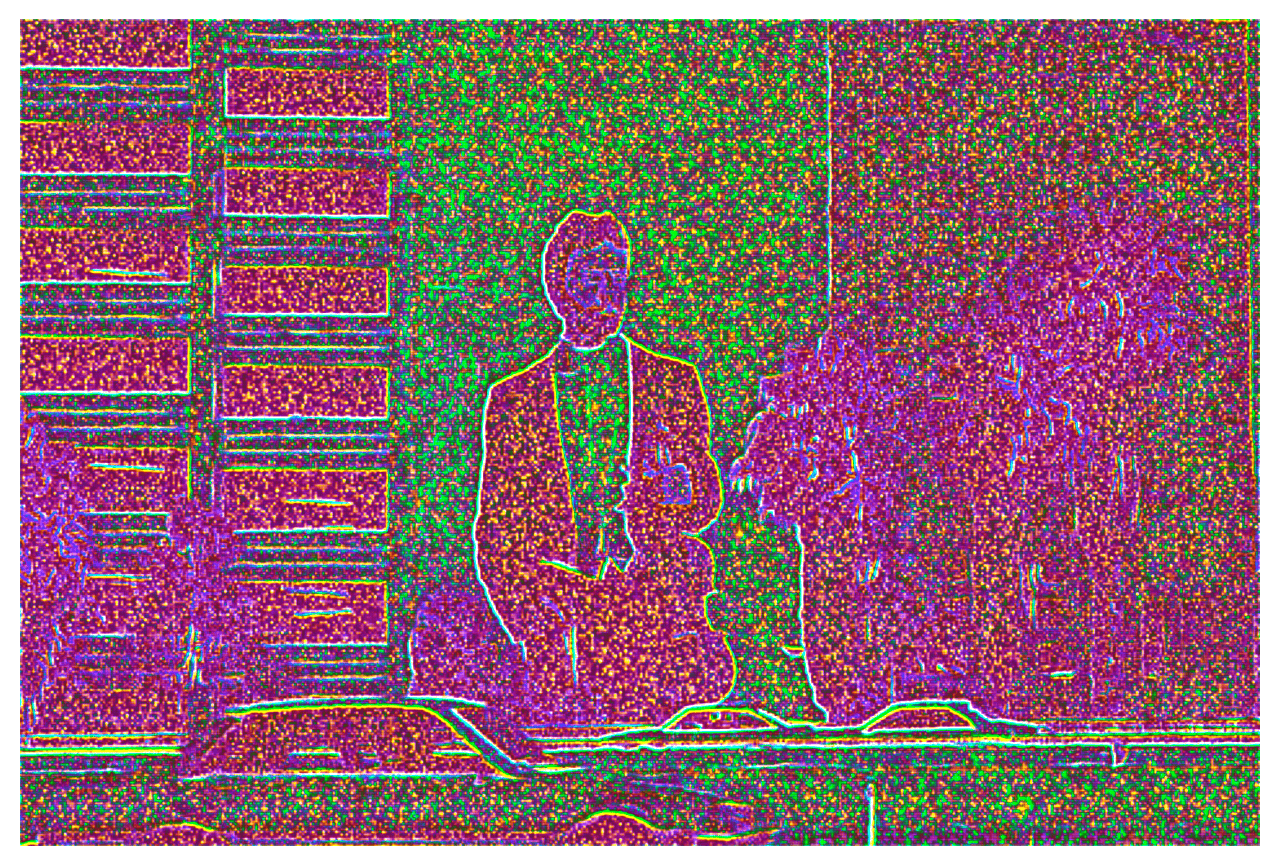}}
	\hfill
	\subfloat[+TFD]{\includegraphics[width=0.33\linewidth]{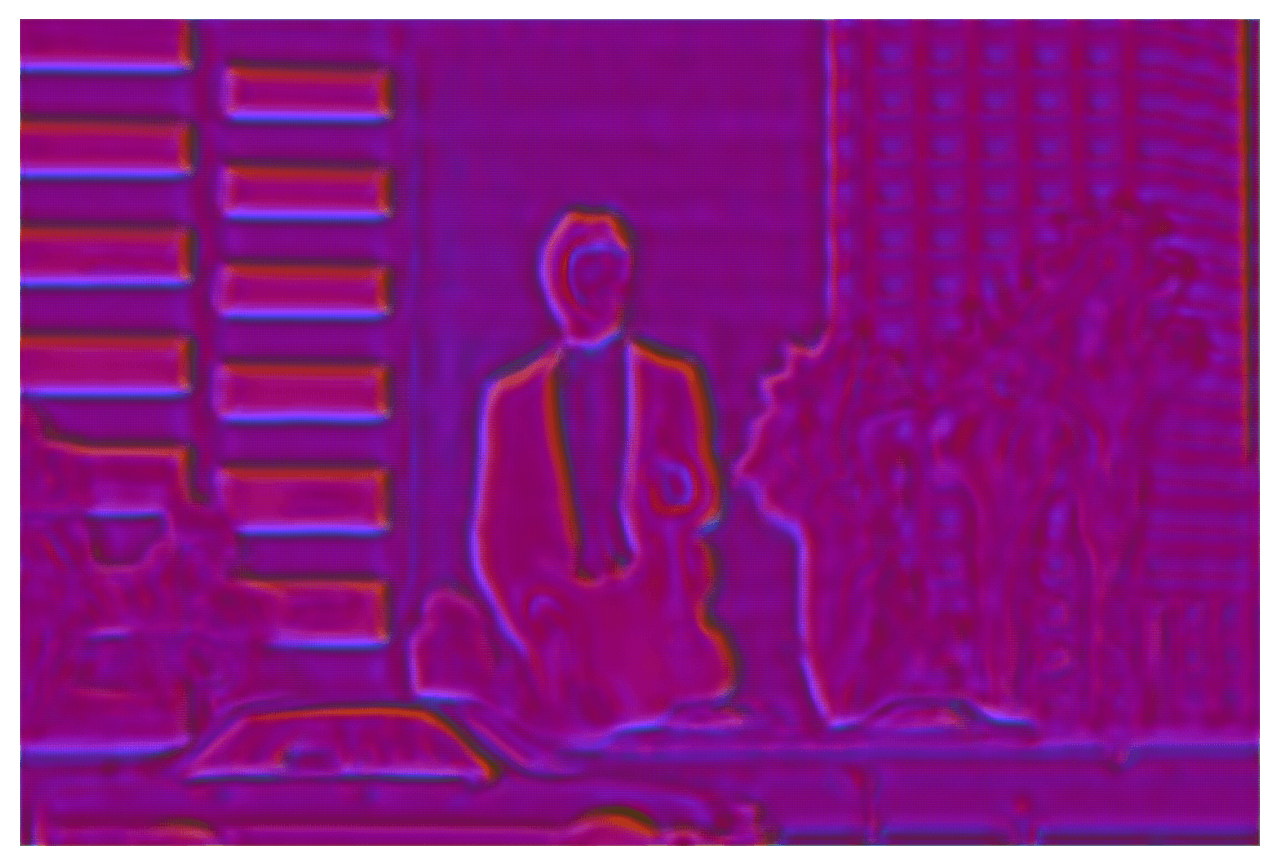}}
	
	\caption{\textbf{Visualization of Feature Denoising Efficacy Across Diverse Visual Domains.}}\label{fig:denoise} 
\end{figure}

\begin{table}[t]
	\centering
	\fontsize{8.5pt}{9.5pt}\selectfont
	\tabcolsep=3.8pt
	\caption{\textbf{Ablation study on different feature fusion strategies.}}
	\label{table:fusion_methods}
	\resizebox{1\linewidth}{!}{\begin{tabular}{l|c|ccccc}
			\whline
			\multirow{2}{*}{Model} & \multirow{2}{*}{Fusion Method} & \multicolumn{5}{c}{Test Sets (PSNR)} \\
			\cline{3-7}
			&  & Set5 & Set14 & BSD100 & Urban100 & Manga109 \\
			\whline
			\multirow{4}{*}{SRResNet} 
			& Addition & 26.27 & 23.76 & 23.45 & 21.89 & 18.86 \\
			& Concatenation & 26.44 & 23.92 & 23.62 & 22.05 & 19.01 \\
			& Multiplication & \textbf{26.71} & \textbf{24.10} & \textbf{23.87} & \textbf{22.28} & \textbf{19.22} \\
			& Baseline & 24.85 & 23.25 & 23.06 & 21.24 & 18.42 \\
			\hline
			\multirow{4}{*}{SwinIR} 
			& Addition & 26.53 & 24.62 & 23.83 & 22.28 & 19.12 \\
			& Concatenation & 26.68 & 24.78 & 23.97 & 22.46 & 19.25 \\
			& Multiplication & \textbf{26.92} & \textbf{24.96} & \textbf{24.14} & \textbf{22.63} & \textbf{19.34} \\
			& Baseline  & 26.25 & 24.53 & 23.91 & 22.18 & 19.10 \\
			\whline
	\end{tabular}}
\end{table}

\noindent\textbf{Feature Visualization Analysis.}
\figref{fig:denoise} compares feature maps across degraded inputs, SRResNet, and our TFD-enhanced outputs, revealing how TFD reshapes feature representation under noise. In a wildlife scene with fox cubs, SRResNet’s features are dominated by chaotic color noise, masking structural details, while TFD recovers clear object boundaries and preserves the cubs’ morphology. For a human subject outdoors, SRResNet features are corrupted by irregular activations, weakening the semantic consistency of facial and body contours, whereas TFD suppresses noise and enhances structural clarity. In a challenging case of a tiger in natural habitat, SRResNet’s features dissolve into noise, making the striped pattern almost unrecognizable, while TFD restores both texture and shape with remarkable fidelity. In an urban scene, SRResNet struggles to maintain geometric regularity, fragmenting building edges and human outlines, while TFD reconstructs rectilinear structures and preserves human silhouettes. These consistent improvements across diverse cases demonstrate that TFD selectively suppresses noise-induced distortions while safeguarding content-relevant details, offering a robust and generalizable solution to feature corruption in degraded image super-resolution.

\noindent\textbf{Comparison with Existing Regulation Strategies.} The quantitative results presented in Tables \ref{table:degradations2}  provide a systematic evaluation of TFD against existing regularization techniques across five benchmark datasets. For CNN-based architectures (SRResNet, RRDB), TFD consistently outperforms both Dropout and Feature Alignment methods across all degradation types, with particularly substantial gains on noise-corrupted images. Specifically, on SRResNet, TFD achieves average PSNR improvements of 0.78dB over the baseline and 0.42dB over the next best method (Alignment) on Set5. This performance advantage extends to transformer-based architectures (HAT, SwinIR), where TFD maintains its superiority despite their inherently stronger baseline performance. The improvement pattern is consistent across datasets of varying complexity - from the simpler Set5 to the challenging Urban100 and content-specialized Manga109. Notably, TFD's efficacy becomes more pronounced under complex degradation scenarios (e.g., Blur+Noise+JPEG), suggesting its robust generalization capability. These results empirically validate our hypothesis that targeted noise suppression, rather than uniform regularization, is crucial for enhancing cross-degradation generalization in image super-resolution.

\noindent\textbf{Comparison with Different Fusion Strategies.} The results in Table \ref{table:fusion_methods} demonstrate the effectiveness of different cross-domain feature integration strategies within our framework. When comparing various integration methods, we observe that  Multiplication consistently outperforms alternative strategies across all benchmark datasets. For SRResNet, adaptive modulation delivers significant improvements over element-wise addition  and channel concatenation. This pattern holds for SwinIR as well, though with smaller margins due to its stronger baseline performance. The superiority of Multiplication can be attributed to its dynamic nature—the frequency-derived attention mask selectively modulates spatial features based on noise concentration, effectively preserving structural details while suppressing noise artifacts. In contrast, element-wise addition treats all features equally, while concatenation merely combines rather than filters information. 
\begin{table*}[t]
	\centering
	\caption{\textbf{Average PSNR of different methods in $\times$4 blind SR on five benchmarks with eight types of degradations.}}
	\label{table:degradations2}
	\renewcommand{\arraystretch}{0.95}
	\resizebox{1\linewidth}{!}{
		\begin{tabular}{ccccccccccc}
			\hline
			Data & Method & Clean & Blur & Noise & JPEG & Blur+Noise & Blur+JPEG & Noise+JPEG & Blur+Noise+JPEG & Average \\ \hline
			\multirow{16}{*}{\rotatebox{90}{Set5}}
			& SRResNet \cite{SRResNet} & 24.85 & 24.73 & 23.69 & 23.69 & 23.25 & 23.41 & 23.10 & 22.68 & 23.68 \\
			& +Dropout ($p$ = 0.7) & 25.63 & 25.23 & 23.82 & 24.05 & 23.47 & 23.64 & 23.46 & 23.01 & 24.04 \\
			& +Alignment & 25.93 & 25.62 & 24.15 & 24.38 & 23.79 & 23.86 & 23.71 & 23.19 & 24.33 \\
			& +TFD & \bf 26.71 & \bf 26.20 & \bf 24.31 & \bf 24.49 & \bf 23.39 & \bf 23.92 & \bf 23.67 & \bf 22.99 & \bf 24.46 \\ \cline{2-11}
			
			& RRDB \cite{ESRGAN} & 25.18 & 25.12 & 22.92 & 23.82 & 23.44 & 23.45 & 23.32 & 22.81 & 23.76 \\
			& +Dropout ($p$ = 0.5) & 26.02 & 26.07 & 23.23 & 24.15 & 23.73 & 23.88 & 23.68 & 23.18 & 24.24 \\
			& +Alignment & 26.78 & 26.55 & 24.02 & 24.70 & 24.12 & 24.14 & 23.93 & 23.26 & 24.69 \\
			& +TFD & \bf 26.83 & \bf 26.59 & \bf 24.96 & \bf 24.56 & \bf 24.07 & \bf 24.04 & \bf 23.81 & \bf 23.14 & \bf 24.75 \\ \cline{2-11}
			
			
			& SwinIR \cite{liang2021swinir} & 26.25 & 26.03 & 24.15 & 24.37 & 23.80 & 23.84 & 23.67 & 22.99 & 24.39 \\
			& +Dropout ($p$ = 0.5) & 26.32 & 26.08 & 24.21 & 24.41 & 24.00 & 23.93 & 23.65 & 23.09 & 24.46 \\
			& +Alignment & 26.49 & 26.23 & 24.61 & 24.68 & 24.13 & 24.17 & 23.89 & 23.09 & 24.66 \\
			& +TFD & \bf 26.92 & \bf 26.43 & \bf 24.76 & \bf 24.56 & \bf 23.90 & \bf 24.03 & \bf 23.77 & \bf 23.09 & \bf 24.68 \\
			\midrule
			
			\multirow{16}{*}{\rotatebox{90}{Set14}}
			& SRResNet \cite{SRResNet} & 23.25 & 23.05 & 22.50 & 22.36 & 22.23 & 22.10 & 22.06 & 21.77 & 22.41 \\
			& +Dropout ($p$ = 0.7) & 23.73 & 23.45 & 22.53 & 22.62 & 22.28 & 22.39 & 22.28 & 21.98 & 22.66 \\
			& +Alignment & 24.12 & 23.80 & 22.68 & 22.99 & 22.65 & 22.63 & 22.55 & 22.16 & 22.95 \\
			& +TFD & \bf 24.54 & \bf 24.15 & \bf 23.02 & \bf 23.11 & \bf 22.43 & \bf 22.79 & \bf 22.60 & \bf 22.13 & \bf 23.10 \\ \cline{2-11}
			
			& RRDB \cite{ESRGAN} & 23.74 & 23.36 & 22.33 & 22.59 & 22.47 & 22.17 & 22.29 & 21.95 & 22.61 \\
			& +Dropout ($p$ = 0.5) & 24.02 & 23.87 & 22.54 & 22.83 & 22.58 & 22.59 & 22.45 & 22.10 & 22.87 \\
			& +Alignment & 24.70 & 24.35 & 22.91 & 23.21 & 22.80 & 22.76 & 22.71 & 22.21 & 23.21 \\
			& +TFD & \bf 24.72 & \bf 24.37 & \bf 23.49 & \bf 23.27 & \bf 22.85 & \bf 22.89 & \bf 22.75 & \bf 22.25 & \bf 23.32 \\ \cline{2-11}
			
			
			& SwinIR \cite{liang2021swinir} & 24.53 & 24.25 & 23.46 & 23.14 & 22.53 & 22.73 & 22.59 & 22.20 & 23.18 \\
			& +Dropout ($p$ = 0.5) & 24.57 & 24.19 & 23.53 & 23.18 & 22.73 & 22.71 & 22.65 & 22.22 & 23.22 \\
			& +Alignment & 24.65 & 24.28 & 23.53 & 23.29 & 22.87 & 22.79 & 22.81 & 22.28 & 23.31 \\
			& +TFD & \bf 24.96 & \bf 24.60 & \bf 23.56 & \bf 23.23 & \bf 22.70 & \bf 22.83 & \bf 22.69 & \bf 22.30 & \bf 23.36 \\ 
			
			\midrule 
			\multirow{16}{*}{\rotatebox{90}{BSD100}}
			& SRResNet \cite{SRResNet} & 23.06 & 22.99 & 22.45 & 22.48 & 22.26 & 22.34 & 22.22 & 22.05 & 22.48 \\
			& +Dropout ($p$ = 0.7) & 23.31 & 23.26 & 22.50 & 22.69 & 22.25 & 22.50 & 22.41 & 22.16 & 22.64 \\
			& +Alignment & 23.83 & 23.64 & 22.77 & 23.04 & 22.53 & 22.79 & 22.62 & 22.32 & 22.94 \\
			& +TFD & \bf 23.87 & \bf 23.71 & \bf 22.67 & \bf 22.96 & \bf 22.36 & \bf 22.78 & \bf 22.52 & \bf 22.29 & \bf 22.89 \\ \cline{2-11}
			
			& RRDB \cite{ESRGAN} & 23.38 & 23.32 & 22.09 & 22.73 & 22.39 & 22.47 & 22.42 & 22.15 & 22.62 \\
			& +Dropout ($p$ = 0.5) & 23.59 & 23.66 & 22.68 & 22.86 & 22.53 & 22.71 & 22.52 & 22.28 & 22.85 \\
			& +Alignment & 24.59 & 24.54 & 23.47 & 23.67 & 22.85 & 23.21 & 22.97 & 22.54 & 23.48 \\
			& +TFD & \bf 24.11 & \bf 24.05 & \bf 23.13 & \bf 23.19 & \bf 22.74 & \bf 22.95 & \bf 22.69 & \bf 22.41 & \bf 23.15 \\ \cline{2-11}
			
			
			& SwinIR \cite{liang2021swinir} & 23.91 & 23.83 & 23.27 & 23.04 & 22.61 & 22.82 & 22.61 & 22.34 & 23.05 \\
			& +Dropout ($p$ = 0.5) & 23.90 & 23.87 & 23.30 & 23.08 & 22.68 & 22.80 & 22.64 & 22.33 & 23.08 \\
			& +Alignment & 24.04 & 23.96 & 23.40 & 23.15 & 22.77 & 22.98 & 22.76 & 22.40 & 23.18 \\
			& +TFD & \bf 24.14 & \bf 24.06 & \bf 23.50 & \bf 23.27 & \bf 22.84 & \bf 23.05 & \bf 22.84 & \bf 22.57 & \bf 23.28 \\ 
			\hline
			\multirow{16}{*}{\rotatebox{90}{Urban100}}
			& SRResNet \cite{SRResNet} & 21.24 & 21.06 & 20.82 & 20.60 & 20.46 & 20.30 & 20.43 & 20.10 & 20.63 \\
			& +Dropout ($p$ = 0.7) & 21.57 & 21.25 & 20.85 & 20.90 & 20.48 & 20.49 & 20.66 & 20.22 & 20.80 \\
			& +Alignment & 21.94 & 21.65 & 21.19 & 21.20 & 20.73 & 20.72 & 20.91 & 20.37 & 21.09 \\
			& +TFD & \bf 22.28 & \bf 21.89 & \bf 21.20 & \bf 21.30 & \bf 20.52 & \bf 20.84 & \bf 20.87 & \bf 20.33 & \bf 21.15 \\ \cline{2-11}
			
			& RRDB \cite{ESRGAN} & 21.57 & 21.18 & 19.61 & 20.93 & 20.57 & 20.40 & 20.74 & 20.24 & 20.66 \\
			& +Dropout ($p$ = 0.5) & 21.89 & 21.75 & 19.92 & 21.12 & 20.53 & 20.70 & 20.84 & 20.33 & 20.89 \\
			& +Alignment & 22.29 & 21.95 & 20.21 & 21.40 & 20.76 & 20.85 & 21.03 & 20.38 & 21.11 \\
			& +TFD & \bf 22.44 & \bf 22.13 & \bf 21.66 & \bf 21.45 & \bf 20.99 & \bf 20.93 & \bf 21.09 & \bf 20.53 & \bf 21.40 \\ \cline{2-11}
			
			
			& SwinIR \cite{liang2021swinir} & 22.18 & 21.90 & 20.56 & 21.32 & 20.89 & 20.79 & 20.98 & 20.45 & 21.13 \\
			& +Dropout ($p$ = 0.5) & 22.27 & 21.99 & 20.67 & 21.38 & 20.92 & 20.91 & 20.96 & 20.55 & 21.21 \\
			& +Alignment & 22.34 & 22.07 & 20.69 & 21.48 & 21.02 & 20.98 & 21.12 & 20.53 & 21.28 \\
			& +TFD & \bf 22.63 & \bf 22.31 & \bf 21.61 & \bf 21.47 & \bf 20.95 & \bf 20.93 & \bf 21.08 & \bf 20.55 & \bf 21.44 \\
			\midrule
			
			\multirow{16}{*}{\rotatebox{90}{Manga109}}
			& SRResNet \cite{SRResNet} & 18.42 & 18.75 & 18.32 & 18.30 & 18.60 & 18.53 & 18.25 & 18.43 & 18.45 \\
			& +Dropout ($p$ = 0.7) & 18.98 & 19.12 & 18.52 & 18.66 & 18.94 & 18.85 & 18.66 & 18.72 & 18.81 \\
			& +Alignment & 19.18 & 19.46 & 19.90 & 19.02 & 19.27 & 19.17 & 18.98 & 19.01 & 19.25 \\
			& +TFD & \bf 19.22 & \bf 19.52 & \bf 18.98 & \bf 18.96 & \bf 19.14 & \bf 19.11 & \bf 18.83 & \bf 18.92 & \bf 19.09 \\ \cline{2-11}
			
			& RRDB \cite{ESRGAN} & 18.59 & 18.64 & 18.30 & 18.41 & 18.83 & 18.43 & 18.38 & 18.41 & 18.50 \\
			& +Dropout ($p$ = 0.5) & 18.73 & 19.03 & 18.72 & 18.60 & 19.15 & 18.81 & 18.59 & 18.71 & 18.79 \\
			& +Alignment & 19.40 & 19.61 & 18.96 & 19.24 & 19.43 & 19.31 & 19.12 & 19.15 & 19.28 \\
			& +TFD & \bf 19.28 & \bf 18.64 & \bf 19.09 & \bf 19.05 & \bf 19.21 & \bf 19.09 & \bf 18.84 & \bf 18.91 & \bf 19.01 \\ \cline{2-11}
			
			
			& SwinIR \cite{liang2021swinir} & 19.10 & 19.27 & 18.71 & 18.95 & 19.07 & 19.02 & 18.79 & 18.80 & 18.96 \\
			& +Dropout ($p$ = 0.5) & 19.15 & 19.30 & 18.83 & 19.03 & 19.12 & 18.98 & 18.75 & 18.84 & 19.00 \\
			& +Alignment & 19.24 & 19.45 & 18.98 & 19.28 & 19.37 & 19.35 & 19.15 & 19.12 & 19.24 \\
			& +TFD & \bf 19.20 & \bf 19.37 & \bf 19.34 & \bf 18.91 & \bf 19.17 & \bf 19.12 & \bf 18.89 & \bf 18.90 & \bf 19.34 \\ 
			\bottomrule
		\end{tabular}
	}
\end{table*}

\end{document}

%% file: preamble.tex
\usepackage[dvipsnames]{xcolor}
